\documentclass[fleqn,10pt]{wlscirep}
\usepackage{geometry}
\usepackage{amsmath, amssymb, amsfonts}
\usepackage{graphicx}
\usepackage{booktabs}
\usepackage{hyperref}
\usepackage{authblk}
\usepackage{float}
\usepackage{placeins}
\usepackage{rotating}
\usepackage{tabularx}
\usepackage{pdflscape}
\usepackage{longtable}
\usepackage{caption}
\usepackage{booktabs}
\usepackage{xcolor}
\usepackage{xurl}
  \usepackage{listings}
  \usepackage[most]{tcolorbox}
  \usepackage{etoolbox}
  \newfloat{listing}{tbp}{lol}
  \floatname{listing}{Listing}

  \definecolor{pyGreen}{HTML}{007020}
  \definecolor{pyBlue}{HTML}{4070A0}
  \definecolor{pyModule}{HTML}{0E84B5}
  \definecolor{pyComment}{HTML}{60A0B0}

  \lstnewenvironment{minted}[2][]{%
    \lstset{
      language=Python,
      basicstyle=\small\ttfamily,
      keywordstyle=\color{pyGreen}\bfseries,
      morekeywords=[3]{glyph},
      keywordstyle=[3]\color{pyModule},
      stringstyle=\color{pyBlue},
      commentstyle=\color{pyComment}\itshape,
      columns=fixed,
      basewidth=0.525em,
      keepspaces=true,
      showstringspaces=false,
      upquote=true,
      tabsize=4,
      breaklines=true,
      breakautoindent=true,
      aboveskip=0pt,
      belowskip=0pt
    }%
    \ifstrequal{#2}{text}{%
      \lstset{language={},deletekeywords=[3]{glyph}}%
    }{}%
  }{}

\definecolor{codebg}{HTML}{F6F6F7}
\definecolor{modImaging}{HTML}{4682B4}
\definecolor{modReport}{HTML}{FE9638}
\definecolor{modLabs}{HTML}{4A9748}

\usepackage{fontspec}
\newfontfamily\hiero{NotoSansEgyptianHieroglyphs-Regular.ttf}[Path=./]

\newcolumntype{L}{>{\raggedright\arraybackslash}X} 
\usepackage{fancyvrb}
\usepackage{fvextra}   
\usepackage{etoc}

\title{\textbf{Glyph: An Interpretability Engine for AI-Driven Scientific Discovery from Clinical Data}}
\author[1,2]{Robbie Holland\thanks{Correspondence: \texttt{robbie.holland@stanford.edu}}}
\author[1,2]{Ashwin Kumar}
\author[2]{Mateus A. Esmeraldo}
\author[3]{David Le}
\author[1,2]{Andrea Fisher}
\author[1,2,4]{Jiazhen Pan}
\author[1,2]{Louisa Fay}
\author[1,2]{Magdalini Paschali}
\author[1,2,4]{Vasiliki Sideri-Lampretsa}
\author[1,2]{Ates Fettahoglu}
\author[1,2]{Stefania Moroianu}
\author[1,2]{Philipp Wesp}
\author[1,2]{Ivan Lopez}
\author[4]{Tom Nuno Wolf}
\author[1,2]{McKenzie S. White}
\author[1,2]{Eduardo Pontes Reis}
\author[5]{Fatima Rodriguez}
\author[1,2]{Curtis P. Langlotz}
\author[6]{Vasilis Syrgkanis}
\author[3]{Imon Banerjee}
\author[1,2,7,8,9]{Akshay S. Chaudhari\thanks{Co-senior authors.}}
\author[1,2]{Sergios Gatidis\protect\footnotemark[2]}

\affil[1]{Stanford Center for Artificial Intelligence in Medicine and Imaging, Stanford University, CA, USA}
\affil[2]{Department of Radiology, Stanford University, Stanford, CA, USA}
\affil[3]{Department of Radiology, Mayo Clinic, Phoenix, Arizona, USA}
\affil[4]{Technical University of Munich, Munich, Germany}
\affil[5]{Division of Cardiology, Stanford University School of Medicine, Stanford, CA, USA}
\affil[6]{Stanford University, Stanford, CA, USA}
\affil[7]{Department of Biomedical Data Science, Stanford University, Stanford, CA, USA}
\affil[8]{Stanford Cardiovascular Institute, Stanford, CA, USA}
\affil[9]{Weill Cancer Hub West, Stanford, CA, USA}

\begin{abstract}
Automated scientific discovery could advance understanding of disease mechanisms and accelerate identification of early disease markers \cite{wang2023scientific}.
Manual annotation of marker candidates is infeasible at population scale, motivating methods that automatically extract predefined phenotypes from clinical data \cite{denny2013systematic,bai2020population}.
These predefined features limit which associations can be discovered, including by agentic AI systems that rely on such tools \cite{gottweis2025towards,swanson2025virtual}.
In contrast, foundation models trained on large clinical datasets can perform tasks beyond their training objectives \cite{blankemeier2026merlin}.
This suggests their embeddings may encode unexplored clinical features and associations that could offer new insights into disease risk, but isolating and interpreting them within dense embeddings remains challenging.
To address this, we introduce \textit{Glyph}, an automated framework that uses sparse autoencoders to extract features from foundation model embeddings of multimodal clinical data, interpret them as coded clinical concepts, and evaluate their value for scientific discovery.
Applied to a large multimodal clinical dataset comprising abdominal CT volumes, radiology reports and laboratory measurements from 41,628 patients, Glyph identified 397 predictive concepts supported by 4,463 Bonferroni-significant associations.
These recovered established prognostic relationships across hepatic, renal, cardiovascular and oncologic disease, with most associations reproducing in held-out cohorts across three sites.
Concept-level confounder adjustment further characterized these concepts by weakening associations known to reflect general illness, surfacing several disease-specific signals and systemic risk markers.
Moreover, in a comparison with literature-derived markers drawn from 342,647 PubMed articles, Glyph concepts showed more frequent replication and significantly larger median odds ratios in multisite data.
These findings reinforce that Glyph provides a platform for exploring clinical hypotheses at scale.
More broadly, they suggest that foundation model features revealed through interpretability methods could expand clinical understanding of disease.
We release Glyph at \url{https://github.com/StanfordMIMI/Glyph} to facilitate hypothesis generation and dissemination of findings across domains.
\end{abstract}

\providecommand{\mainpagetotal}{?}
\begin{document}

\maketitle

\section{Introduction}
Identifying early disease markers has repeatedly transformed clinical practice and patient outcomes \cite{xiao2025viral}.
Clinical discovery is inherently challenging as evidence of potentially novel markers is captured across diverse, often unstructured and high-dimensional modalities, including 2D and 3D imaging, laboratory results, and textual reports \cite{sedlakova2023challenges}.
Moreover, the growing volume of clinical data, including approximately 300 million CT scans acquired worldwide each year \cite{schockel2020developments}, creates opportunities to identify new markers and uncover unknown associations with a broader range of clinical outcomes.
Yet this growth also highlights a gap in our ability to process these data at scale, given the time and expertise required to manually define, annotate, and validate candidate disease markers.

Artificial intelligence (AI) is increasingly proposed as a means of accelerating scientific discovery \cite{wang2023scientific}.
In protein folding and materials science, targeted models have sharply reduced the time and cost of identifying promising candidates \cite{jumper2021highly,varadi2022alphafold,merchant2023scaling}.
Attempts to automate clinical science have searched large databases of manually coded phenotypes through phenome-wide association studies (PheWAS) \cite{denny2013systematic,jensen2014temporal}.
Efforts to accelerate this approach use deep learning or agents to automate the annotation of phenotypes directly from images, but still restricts investigation to a preselected set of chosen markers \cite{bai2020population,gottweis2025towards,gottweis2026accelerating}.
As a result, clinical discovery still lacks an efficient framework for identifying open-ended, reproducible, and independent disease markers within multimodal data.

One potential solution lies in foundation model (FM) embeddings.
Trained on large volumes of multimodal medical data, FMs produce embeddings that characterize patient states \cite{blankemeier2026merlin}.
Such embeddings have been observed to perform well on tasks they were not explicitly trained for \cite{moor2023foundation,zhou2023foundation,ding2025multimodal,blankemeier2026merlin}, including disease prognosis and risk stratification, suggesting that they contain clinically meaningful associations beyond their original training objectives.
However, these markers remain entangled within high-dimensional vectors, making them difficult to isolate and evaluate as candidate markers.
Recently, mechanistic interpretability \cite{bricken2023monosemanticity,gao2024scaling,templeton2024scaling} has emerged as a self-supervised approach to uncover interpretable concepts encoded within FM embeddings, without the need for prospective annotation.
Yet its broader potential for transforming FMs into engines of clinical discovery remains largely unexplored.

To address this gap, we introduce \textit{Glyph}, a framework that turns foundation model (FM) embeddings into concepts for clinical discovery.
The name derives from ``glyph,'' or a symbol that conveys meaning, reflecting our goal of giving hidden information a form that can be studied.
Drawing on mechanistic interpretability, Glyph discovers coded \textit{concepts} directly in FM embeddings and detects them across patients without requiring candidate markers to be defined and annotated in advance.
These extracted concepts are measurable, interpretable patterns in patient data that together preserve the FM's predictive capacity.
Glyph integrates them with established clinical variables and tests their associations with disease-related outcomes across patient populations.
This enables efficient, closed- and open-ended discovery that is infeasible with embeddings alone or with agentic systems operating on a limited set of pre-defined concepts.
Our contributions are as follows:

\begin{figure}[htbp]
    \centering
    \includegraphics[width=0.9\textwidth]{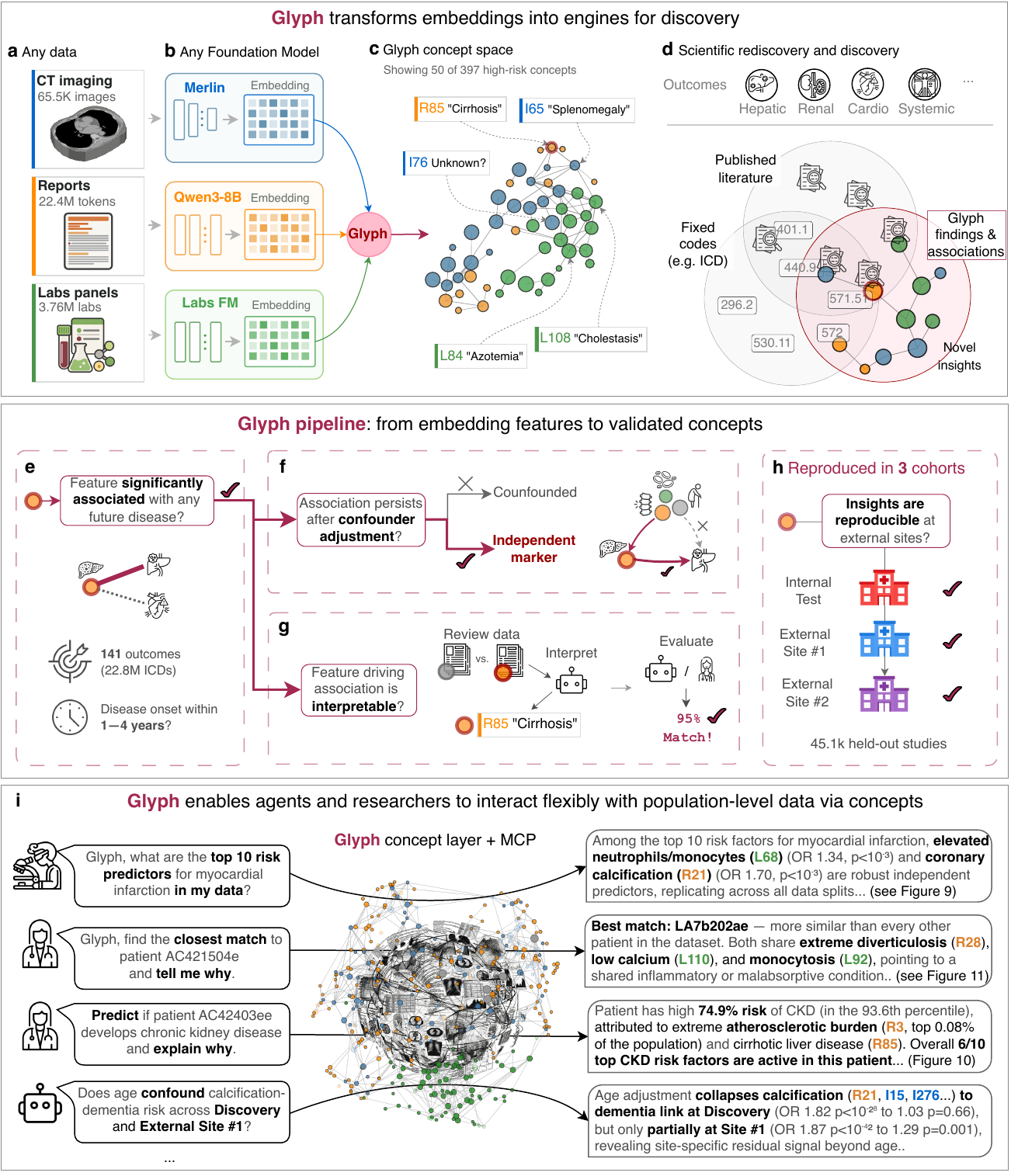}
    \caption{\textbf{Transforming foundation model embeddings into engines of clinical discovery.}
    \textbf{a} Glyph can be applied to any data expected to contain useful markers for discovery.
    \textbf{b} 
    Foundation models compress multimodal clinical data into high-dimensional embeddings, typically used for predefined prediction tasks. 
    The structural contents of embeddings driving these predictions is typically unexplored.
    \textbf{c} 
    Glyph instead transforms FM embeddings into engines for discovery, mapping a space of concepts hidden within these embeddings.
    Glyph goes beyond embeddings by decomposing them into sparse concepts that can be studied individually and in selected combinations (Figure~\ref{figure:1}c).
    \textbf{d} 
    Glyph concepts recover associations established in clinical research and ICD-based PheWAS analyses, while revealing complementary associations not yet discovered by either approach.
    \textbf{e--f} Glyph evaluates concepts for \textbf{e} associations with future disease after correction for multiple testing, \textbf{f} associations that persist after adjustment for measured confounders, and \textbf{g} clinical interpretability. \textbf{h} All findings are tested for reproducibility across multiple cohorts.
    \textbf{i}
    Validated concepts provide an interface for asking complex questions of FM embeddings and grounding the answers in the underlying data.}
    \label{figure:1}
\end{figure}

\begin{enumerate}
\item \textbf{Establishing the concept space.}
To make FM embeddings useful for discovery, we develop the Glyph framework to first represent such embeddings in sets of sparse features, and then filter and interpret these as clinically meaningful concepts. 
Specifically, we evaluated features for significant prognostic associations, reproducibility across three sites, persistence after adjustment for measured confounders, and overall interpretability.
Glyph identified 397 predictive concepts bearing 4,463 significant associations to 141 future disease outcomes meeting a Bonferroni threshold of $p \leq 8.28\times10^{-8}$.
Identifying concepts with at least one significant association yielded a space of high-risk concepts, capturing disease risk across modalities.
Concept interpretations and prognostic associations also reproduced in external cohorts.

\item \textbf{Rediscovery and discovery.}
The validated concept space recovered established prognostic relationships across several disease axes, including hepatic, renal, cardiovascular, and oncologic disease. 
When tested in held-out cohorts, these associations reproduced across institutions.
Glyph then extended these findings by linking established risk markers to additional outcomes and identifying reproducible prognostic associations that remain poorly characterized in the literature.
Beyond disease-specific risk profiles, Glyph identified systemic risk markers across modalities, with one imaging and three laboratory concepts retaining broad associations across disease groups after adjustment for potential confounders.
Glyph revealed that nearly half of the high-risk imaging concepts were not consistently documented in the accompanying radiology reports.

\item \textbf{Comparison with literature}
We investigated whether and in which domains Glyph brought additional prognostic value over the clinical literature.
To this end, a co-scientist agent processed 342,647 clinical papers covering each discovery axis, and reproduced 411 published markers in our data.
We found using just ten Glyph concepts per axis led to significantly stronger models in external data than literature markers alone (3.34\% AUC gain, $p=2.66\times10^{-28}$), with predictive improvements of up to 11.7\% in specific axes.
Comparing the top ten Glyph concepts with the top ten literature markers per outcome, selected at the Discovery Site, Glyph associations reproduced more frequently in the pooled held-out cohort (64.8\% vs. 41.2\%, $p=2.54\times10^{-6}$) at higher median confounder-adjusted odds ratios (1.31 vs. 1.16, $p=5.0\times10^{-5}$).

\item \textbf{Comparison with pre-coded phenotypes}
Similarly, we also asked whether pre-coded phenotypes could support the same discoveries, comparing Glyph with an automated PheWAS-style analysis of ICD-derived Phecodes.
Across the same outcomes, associations from Glyph reproduced more often than those from Phecodes at Internal Test (69.3\% vs. 49.1\%, $p=9.77\times10^{-4}$) and External Site \#1 (74.3\% vs. 63.9\%, $p=2.93\times10^{-3}$), and had larger median effect sizes.
Overall, Glyph showed a complementary benefit to existing disease coding protocols and clinical literature.

\item \textbf{Open-source and use.}
Glyph extracts additional value from existing pretrained FMs, without retraining.
To enable researchers to find our concept space in their cohorts, or build new concept spaces from their data, we open-source Glyph at \href{https://github.com/StanfordMIMI/Glyph}{\texttt{github.com/StanfordMIMI/Glyph}}, installable via \texttt{pip install glyph-discovery}.
A validated concept space also supports other applications including explainable patient-level predictions and retrieval.
We also release the Glyph MCP, enabling clinicians, engineers, and AI agents to investigate patient data and test open-ended hypotheses and applications through concepts.
\end{enumerate}

\begin{figure}[htbp]
    \centering
    \includegraphics[width=0.95\textwidth]{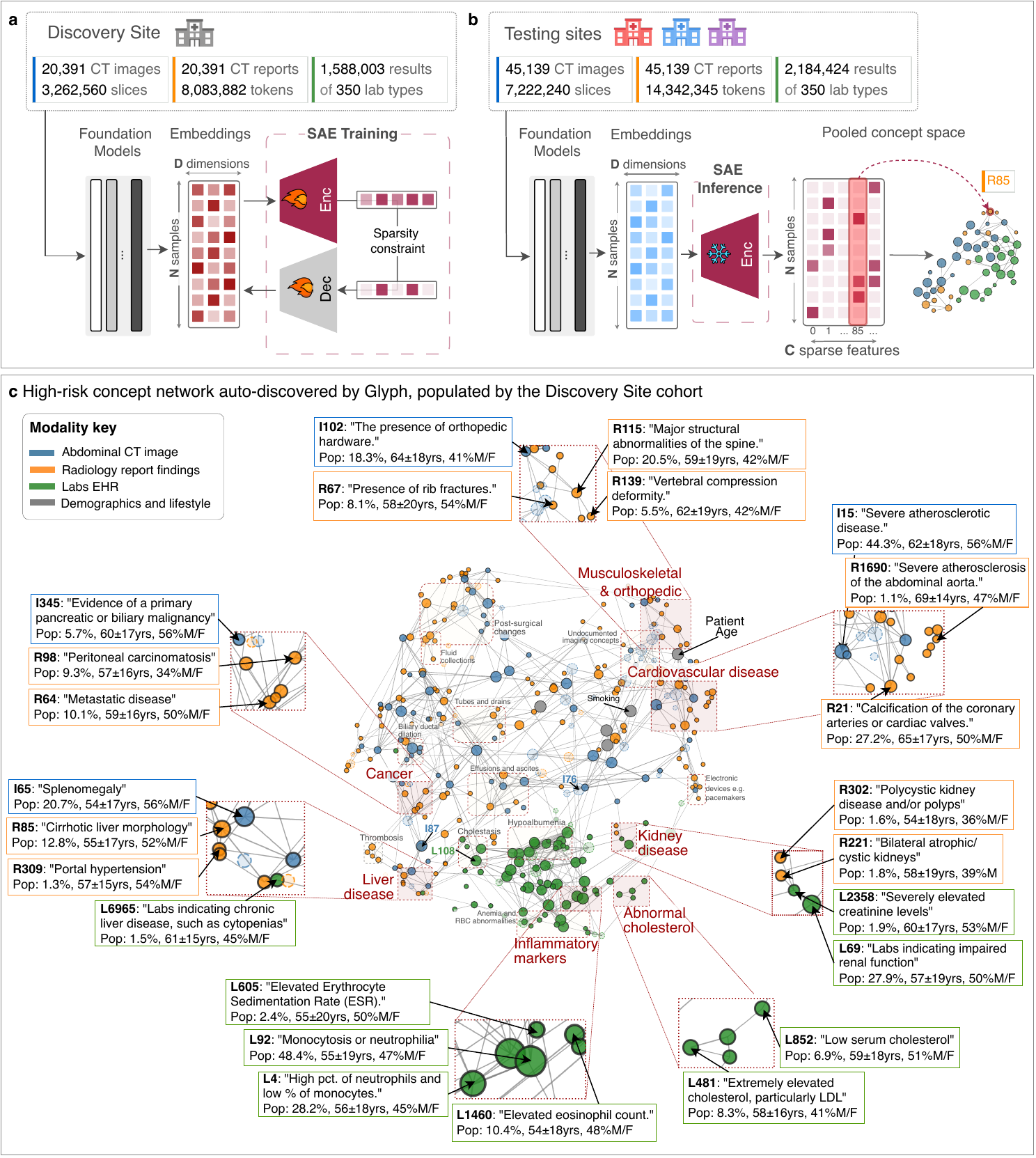}
    \caption{The Glyph framework. An interactive version is available at \url{https://stanfordmimi.github.io/Glyph-research/}.
    \textbf{a} 
    At the Discovery Site, Glyph trains sparse autoencoders (SAEs) to decompose foundation model (FM) embeddings into sparse concepts.
    \textbf{b} 
    Each SAE encoder is then frozen as a concept extractor and its decoder discarded.
    Concepts are re-extracted zero-shot in held-out internal and external cohorts to assess reproducibility of their interpretations, prognostic associations, and confounder analyses.
    \textbf{c} The discovered concept space maps the clinical characteristics and disease burden of patients undergoing abdominal CT at the Discovery Site, alongside routine variables such as age and smoking status.
    We show 397 concepts (9.26\% of the 4,285 candidate features) that were associated with at least one of the 141 prognostic outcomes after Bonferroni correction.
    Edges show correlations between concepts in the top 2.5\%.
    Clinically related concepts were found to cluster into subregions (dashed boxes) corresponding to liver, renal, and cardiovascular disease, and systemic inflammatory and cholestatic illness.
    Node size scales with the logarithm of the number of patients who share the specific concept.
    A confounder-adjusted version of this network is shown in Figure~\ref{figure:extended:6}.}
    \label{figure:2}
\end{figure}

\section{Results}
Glyph was deployed in a multimodal \textit{Discovery Site} cohort of 14,652 patients, comprising 20,391 3D abdominal CT volumes (3,262,560 preprocessed slices), radiology reports (8,083,882 tokens), and 1,588,003 lab entries spanning 350 types (Figure~\ref{figure:1}a).
These three modalities were embedded using the Merlin image encoder \cite{blankemeier2026merlin}, Qwen3-Embedding-8B \cite{zhang2025qwen3}, and Labs FM, respectively (Figure~\ref{figure:1}b).
Glyph then decomposes these embeddings into sets of sparse features (Figure~\ref{figure:1}c). 
We identified when Glyph concepts rediscovered established prognostic associations, and when they reveal complementary associations absent from the literature or not discoverable by automated PheWAS-style discovery analysis (Figure~\ref{figure:1}d).

To enable this analysis, we developed a rigorous pipeline to process sparse features in embeddings as concepts useful for discovery (Figure~\ref{figure:1}e--g).
To maximize discovery yield, we used the Discovery Site validation set to identify 141 disease outcomes whose onset 1--4 years after abdominal CT was most predictable from FM embeddings (see Methods, Section~\ref{methods:outcome_selection}).
These outcomes spanned hepatic, renal, cardiovascular, metabolic, and neurological disease.
To surface high-risk concepts, Glyph filters features to those with at least one significant prognostic association (Figure \ref{figure:1}e), before assessing their independent risk adjusted for specific confounders (Figure \ref{figure:1}f), and estimating each feature's clinical interpretation (Figure \ref{figure:1}g).
At each step, we assessed the reproducibility of these analyses across three held-out cohorts comprising 26,976 patients and 45,139 CT studies, with 7,222,240 preprocessed CT slices, 14,342,345 radiology-report tokens, and 2,184,424 lab entries (Figure \ref{figure:1}h).
This pipeline led to the set of 397 high-risk concepts that can be used in downstream discovery workflows (Figure \ref{figure:1}i).

\subsection{Glyph discovered a clinical concept space in FM embeddings}
\label{section:results:clinical_concept_space}
To first identify a pool of candidate features, Glyph fits one SAE architecture per modality (Figure~\ref{figure:2}a).
On the Discovery Site, SAEs identified 3,315 features in embeddings of reports, 343 in images, and 627 in labs, totaling 4,285 features.
Once features are learned, they can be re-extracted elsewhere without recalibration (Figure ~\ref{figure:2}b).
Specifically, FMs were reapplied to extract embeddings of data in the Internal Test, External Site \#1 and External Site \#2 cohorts.
The trained SAEs were then applied zero-shot to each site's embeddings to re-extract the same set of 4,285 features.
Overall, these features preserved the majority of the original embeddings' variance and predictive performance (see Extended Figure \ref{figure:extended:pipeline} and Section~\ref{methods:saes}).
These features were then filtered by Glyph's discovery pipeline, explored in detail in Sections \ref{section:results:association_study}, \ref{section:results:confounder_study} and Section \ref{supplementary:autointerp}, yielding the set of 397 validated, high-risk \textit{concepts}.
Textual interpretations of these concepts were then attempted via a contrastive approach detailed in Section~\ref{supplementary:autointerp} and illustrated in Figure~\ref{figure:8}.
Throughout, we refer to concepts by a modality-prefixed index: \textcolor{modReport}{\textbf{R}} for radiology report, \textcolor{modImaging}{\textbf{I}} for imaging, and \textcolor{modLabs}{\textbf{L}} for laboratory concepts.

High-risk concepts clustered into neighborhoods reflecting liver, kidney, cardiovascular, musculoskeletal, and malignant disease, as well as inflammatory states and abnormal cholesterol (Figure~\ref{figure:2}c).
Within the liver and kidney neighborhoods, granular concepts from different modalities overlapped.
Concepts for fluid accumulation, including effusions and ascites, lay nearby and were closely correlated with laboratory concepts reflecting hypoalbuminemia, reflecting known comorbidities to liver cirrhosis \cite{schuppan2008liver}. 
Laboratory concepts also provided information complementary to imaging and reports, such as evidence of abnormal cholesterol, situated adjacent to kidney disease \cite{vaziri2006dyslipidemia}, and inflammatory markers.
Other regions contained high-risk imaging concepts with no corresponding report concepts, suggesting radiological features that are not routinely documented.
The concept space also isolated the presence of medical devices, including tubes and drains, and postsurgical artifacts, that could confound predictions based on whole embeddings.
It also integrated established variables not derived from SAEs, such as patient age and smoking status, alongside the extracted concepts, situating these adjacent to cardiovascular and musculoskeletal disease, which are also known correlates \cite{oliveira2020co}.
These 397 high-risk concepts are explored for their discovery potential in the remainder of the paper.

\begin{figure}[htbp]
    \centering
    \includegraphics[height=0.79\textheight]{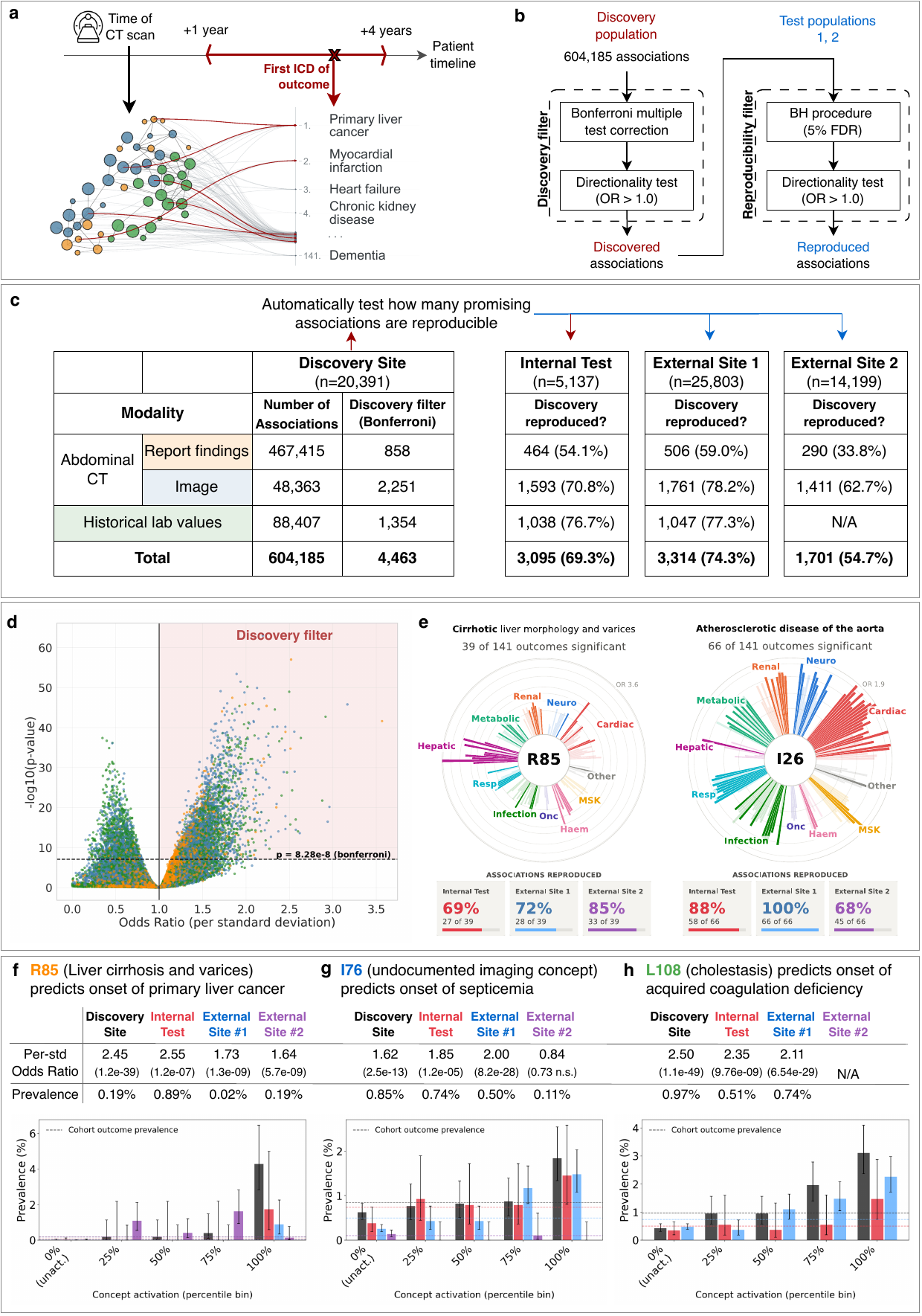}
    \caption{Glyph concept-wide association study.
    \textbf{a} Each Glyph concept is tested for association to each of 141 future disease outcomes within 1 to 4 years, yielding 604,185 concept-outcome pairs evaluated using per-standard-deviation odds ratios.
    \textbf{b} A \textit{Discovery filter} ($p \leq 0.05$ adjusted for multiple testing via Bonferroni correction, equivalent to $p \leq 8.28\times10^{-8}$, and OR $> 1.0$) applied on the Discovery Site yields 4,463 (0.739\%) promising associations.
    \textbf{c} Number of promising associations per modality on the Discovery Site, totalling 4,463; these are fixed and tested for reproducibility without calibration on the Internal Test, External Site \#1, and External Site \#2 cohorts using a 5\% FDR threshold (Benjamini--Hochberg), with the percentage replicating shown per modality and cohort.
    \textbf{d} Volcano plot of all concept-outcome associations and the Discovery filter selection.
    \textbf{e} Risk profiles and reproducibility statistics for one organ-specific and one systemic-risk concept.
    \textbf{f--h} Sample counts and observed probabilities of future disease onset, with confidence intervals, across concept-activation bins.}
    \label{figure:3}
\end{figure}

\subsection{Glyph identified reproducible markers of future disease risk}
\label{section:results:association_study}
Having constructed the concept space, we tested whether it contained any significant, reproducible high-risk markers for future disease onset.
Because the cohort consisted of patients who underwent abdominal CT, associations were expected to be enriched for abdominal pathology. 
ICD codes provided broad, cross-institutional outcome labels but may be affected by undercoding, miscoding, delayed entry, and institutional differences in coding practices.
Associations may also reflect clinical workflows, motivating the confounder analysis in Section~\ref{section:results:confounder_study}.

To identify high-risk concepts, we first computed per-standard-deviation odds ratios (ORs) between all pairings of 4,285 features and 141 outcomes, yielding 604,185 candidate associations at the Discovery Site.
These comprised 467,415 radiology report, 48,363 imaging, and 88,407 laboratory associations (Figure~\ref{figure:3}a; see Methods, Section~\ref{section:methods:or}).
To set a stringent threshold for evidence, we applied the \textit{Discovery filter}, requiring a Bonferroni-corrected $p \leq 8.28\times10^{-8}$ and $\mathrm{OR} > 1.0$ (Figure~\ref{figure:3}b).
This analysis uncovered 4,463 significant associations (0.739\% of all pairs), shown in the volcano plot in Figure \ref{figure:3}d. 
These associations spanned radiology reports (858; 0.184\%), imaging (2,251; 4.65\%), and laboratory measurements (1,354; 1.53\%), establishing that the concept space contains signals of future disease across all three modalities.
This narrowed the space to the 397 high-risk concepts with at least one significant association used in the remainder of the analyses (as shown in Figure~\ref{figure:2}c).

Having identified thousands of significant associations, we then assessed whether any were reproducible in the internal and external sites.
This was tested using a less conservative \textit{Reproducibility filter}, affording a 5\% false discovery rate via the Benjamini--Hochberg procedure (see Methods, Section~\ref{section:methods:reproducibility}).
Of the 4,463 promising associations, 3,095 (69.3\%) were reproduced without calibration on the Internal Test cohort of 3,667 patients (see Figure~\ref{figure:3}c).
Remarkably, a greater number, 3,314 (74.3\%), were reproduced at the secondary site, External Site \#1, reflecting its larger cohort of 17,224 patients and indicating that associations generalized across institutions.
Moreover, External Site \#2 which was out of state, reproduced 1,701 (54.7\%) of the radiology-report and imaging associations, reinforcing the generalizability of these findings across institutions.
Overall, this showed that Glyph automatically surfaced concepts that were reproducibly predictive of future disease across patient populations.

This prompted us to investigate the nature of these associations, and whether they reflected insights in the literature.
Less than a third of abdominal CT patients at External Site \#2 had a single recorded lab result, so associations from labs concepts on this cohort were not run.
These rediscoveries recapitulated well established risk factors across hepatic, renal, and cardiovascular disease, and reveal several systemic risk factors in our data.

\textbf{Rediscovery of hepatic disease progression:}
Among the strongest predictors of hepatic disease were L108 (cholestasis), I65 (splenomegaly), I21 (ascites) and R85 (liver cirrhosis and varices).
R85 predicted future primary liver cancer with odds ratios and p-values of 2.45 $p=1.2\times10^{-39}$, 2.55 $p=1.2\times10^{-7}$, 1.73 $p=1.3\times10^{-9}$, 1.64 $p=5.7\times10^{-9}$ at the Discovery Site, Internal Test, External Site \#1 and External Site \#2, respectively (Figure \ref{figure:3}f). It was also associated with portal hypertension (OR 2.19--3.57), with both associations replicating across all four cohorts. 
These findings accord with established complications of cirrhosis~\cite{vogel2022hepatocellular,de2022baveno}.
R85 was associated with the onset of 39 of 141 outcomes (Figure \ref{figure:3}e), tracing the recognised natural history of cirrhosis through jaundice, hepatic necrosis, varices, splenomegaly and coagulopathy. Associations also extended to renal, respiratory and infection diagnoses, consistent with reported evidence of these conditions in patients with cirrhosis~\cite{maiwall2020incidence,nahon2017bacterial,foreman2003cirrhosis}.

L108 also recovered the majority of hepatic outcomes.
These included future primary liver cancer (OR 2.08--2.57) and portal hypertension (OR 1.63--2.92). Li and Apte linked bile-acid accumulation in cholestasis to liver injury and cancer~\cite{li2015bile}, but did not evaluate any prognostic relationships.
Reproducible associations extended to future liver transplantation and coagulation (see Figure \ref{figure:3}h).

\textbf{Rediscovery of renal disease progression:}
The strongest marker of renal disease was L84 (azotemia), which reproducibly traced the stages and complications of chronic kidney disease, including chronic renal failure (2.02--2.52), progression to stage IV disease (1.77--2.73), initiation of renal dialysis (2.04--3.21), eventual kidney transplantation (2.53--2.87), and the metabolic consequences of uremia in anaemia of chronic kidney disease (1.87--2.44) and hyperkalaemia (1.80--2.07)~\cite{stevens2024kdigo}.
L84 also extended beyond renal disease to cardiovascular, metabolic, infectious, respiratory, hepatic and haematological outcomes. Elevated blood urea nitrogen (BUN), a measure of azotemia, predicted mortality after acute coronary syndromes in prior work, although that study did not examine later diagnoses across disease groups~\cite{kirtane2005serum}.

By comparison, L69 (impaired renal function, defined by increased creatinine and BUN and reduced eGFR) was far more specific, as its seven significant associations were all renal (against 11 for L84 among 76 spanning every disease group).
Its effects were smaller on all four shared outcomes (1.09--1.54 versus 1.77--2.87), it did not predict dialysis initiation or acute renal failure.
Notably, urinary protein loss was not isolated by any single concept, despite albuminuria being a second axis of chronic kidney disease staging alongside eGFR~\cite{stevens2024kdigo}; nephrotic-range values were distributed among concepts dominated by hypoalbuminaemia (L34, L45, L61, L87).

Unlike the hepatic rediscoveries, no radiological concept captured renal disease progression, consistent with the weak association between visible kidney structure and renal function \cite{grenier2016radiologic}.
Specifically, R398 (kidney atrophy) and R302 (polycystic kidney disease) reached only modest effect sizes (1.16--1.41) and replicated inconsistently.
By contrast, I109 (aortic calcification) and L127 (cytopenia) were associated with 10 and 12 of 13 renal outcomes, respectively, with strong reproduction across cohorts. 
Prior studies established abdominal aortic calcification as a cardiovascular prognostic marker in CKD~\cite{leow2021prognostic}; here it also marked later renal diagnoses more consistently than kidney-specific imaging findings.

\textbf{Rediscovery of cardiovascular risk markers:}
Glyph recovered early cardiovascular markers of calcification and atherosclerosis.
Specifically, R21 (calcification of the coronary arteries or cardiac valves) was a prominent report-derived cardiovascular predictor, reproducibly predicting myocardial infarction (OR 1.25--1.70), heart failure with preserved (1.58--1.83) and reduced (1.40--1.92) ejection fraction, and coronary atherosclerosis (1.52--1.97). Prospective studies have linked coronary and mitral annular calcification to subsequent cardiovascular events~\cite{detrano2008coronary,fox2003mitral}.
Both R21 and R3 (moderate to severe atherosclerotic disease) replicated strongly at External Sites \#1 and \#2 (R21: 100\% and 73.7\%; R3: 91.7\% and 86.1\%), well above the 33.8\% cohort average at External Site \#2.

Imaging concepts outperformed report-based concepts: the imaging group, comprising I109 (abdominal aortic calcification), I15 (abdominopelvic atherosclerosis) and I26 (aortic atherosclerosis; Figure~\ref{figure:3}e), covered 23 of 28 cardiovascular outcomes (82.1\%), against 19 of 28 (67.9\%) for the report-based group, and replicated more consistently across the Internal Test and External Sites \#1 and \#2 (96.1\%, 98.7\% and 93.4\% versus 89.6\%, 97.9\% and 70.8\%).
Imaging concepts also more strongly recovered atherosclerosis's known systemic vascular effects, covering more outcomes and with larger effect sizes than report-based concepts across renal (11 of 13, 84.6\% vs. 7 of 13, 53.8\%) and neurocognitive outcomes (8 of 12, 66.7\% vs. 6 of 12, 50.0\%) \cite{pickhardt2020automated}, including vascular dementia \cite{bos2015atherosclerotic}.
Atherosclerosis and vascular calcification concepts (I26, I106, R21 and R420) also showed associations beyond cardiovascular disease, including with femoral and vertebral fractures \cite{schulz2004aortic} and chondrocalcinosis.

Most inflammatory concepts (L0, L3, L4, L8, L92) did not have substantial associations with cardiovascular outcomes, which was unexpected given the established role of inflammation in atherosclerosis~\cite{libby2021changing}.
L68, more indicative of infection (a blood-count pattern with elevated neutrophils, white cells, monocytes and platelets) than auto-immune inflammation, was associated with 5 of the 28 cardiovascular outcomes. 
It reproducibly predicted ischaemic heart disease (OR 1.17--1.33) and myocardial infarction (1.21--1.34).
L121 captured delayed cardiac conduction on ECG and was associated with later arrhythmia and heart failure, consistent with prospective evidence linking prolonged QRS duration to incident heart failure~\cite{dhingra2006electrocardiographic}. L1488 similarly captured abnormal RR intervals, but neither concept's adjusted associations reproduced consistently, and both may reflect ECG testing prompted by suspected cardiac disease.
Concepts capturing abnormal cholesterol (L182, L184, L481, L852) did not emerge as major cardiovascular risk markers in this cohort, despite the established causal role of LDL cholesterol in atherosclerotic cardiovascular disease~\cite{ference2017low}, potentially due to widespread use of statins in this cohort.

\textbf{Concepts encoding systemic risk markers:}
Beyond organ-specific associations, several concepts behaved as broad markers of systemic risk, each associated with more than half of the outcomes tested and spanning nearly every disease group.
I87 (uncharacterized hepatobiliary pattern) showed the broadest prognostic profile, with associations across 96 of 141 outcomes (68.1\%) and all 11 disease groups.
Although I87 was correlated with prior cholecystectomy and, to a lesser extent, cirrhotic morphology, portal hypertension and prior liver transplantation, these findings neither fully characterized the imaging phenotype nor fully explained its prognostic associations (Section \ref{section:results:confounder_study}).
91 of its 96 associations stemming from I87 were reproducible at Internal Test (95\%), 93 at External Site \#1 (97\%) and 73 at External Site \#2 (76\%).
A targeted investigation into I87 is shown in Figure \ref{figure:extended:I87}.

L127 (cytopenia) and L108 (cholestasis) followed with associations across 89 (63.1\%) and 87 outcomes (61.7\%), respectively, each spanning 10 of 11 disease groups.
L108 extended beyond liver disease to predict future infection, electrolyte disturbance, renal failure and cardiac failure.
Cholemic nephropathy offers one possible explanation for these renal associations in patients with severe liver disease~\cite{brasen2019cholemic,van2013bile}.
L127 (cytopenia) likewise reached cardiovascular, infectious, renal and metabolic outcomes.
L84 (azotemia) followed with 76 associations (53.9\%) and spanned all 11 groups.
This breadth generalized across cohorts, as L108 reproduced 82 of its 87 associations at the Internal Test cohort (94\%) and 84 (97\%) at External Site \#1, L127 79 (89\%) and 86 (97\%), and L84 69 (91\%) and 75 (99\%), surpassing the average rates of 69.3\% and 74.3\%.

Concepts related to fluid accumulation surfaced as systemic risk markers. 
Specifically, I21 (ascites) was associated with 68 outcomes across 10 groups and carried the largest single effect of any systemic concept (OR up to 3.17), while I25 (combined ascites, anasarca and effusion) reached 62 outcomes across all 11 groups, and R101 (pleural effusion with ascites) 30.
A second family encoded hypoalbuminaemia and anaemia (L34, L59, L77, L82, L87, L91, L124), each associated with 39 to 71 outcomes across 10 disease groups with median odds ratios of 1.42 to 1.62.

\textbf{Infection and sepsis:}
L127 (cytopenia) \cite{warny2018lymphopenia} and L108 (cholestasis) were each associated with 14 of 22 subsequent infectious and inflammatory outcomes, although neither was specific to this outcome group.
I76 (severely distended bowel pattern) was associated with 5 of 20 outcomes, including bacterial pneumonia, sepsis and septicemia (Figure \ref{figure:3}g), despite the imaging phenotype not being consistently documented in the corresponding radiology reports.
Visual review identified severely distended, fluid-filled bowel loops (Figure~\ref{figure:8}d--e; see Section~\ref{supplementary:autointerp}).
This recapitulates prior evidence linking bowel obstruction to bacterial translocation and postoperative septic complications~\cite{sagar1995intestinal}.

\textbf{Rediscovery of metastatic sequelae}
R64 (metastatic disease) predicted systemic complications of advanced malignancy, including cachexia and pancytopenia (Discovery ORs 1.63--1.85, both reproducing at Internal Test and External Site \#1), capturing the wasting and marrow failure of disseminated disease~\cite{biswas2020understanding}, alongside neutropenia, bacteremia, and septic shock (ORs 1.45--1.58, same replication pattern), reflecting common infectious complications~\cite{williams2004hospitalized}. 
Overall, while 25 of R64's 32 significant associations (78\%) replicated at Internal Test and 26 (81\%) at External Site \#1, none reproduced at External Site \#2, likely due to differences in coding practices. 

\begin{figure}[htbp]
    \centering
    \includegraphics[width=0.8\textwidth]{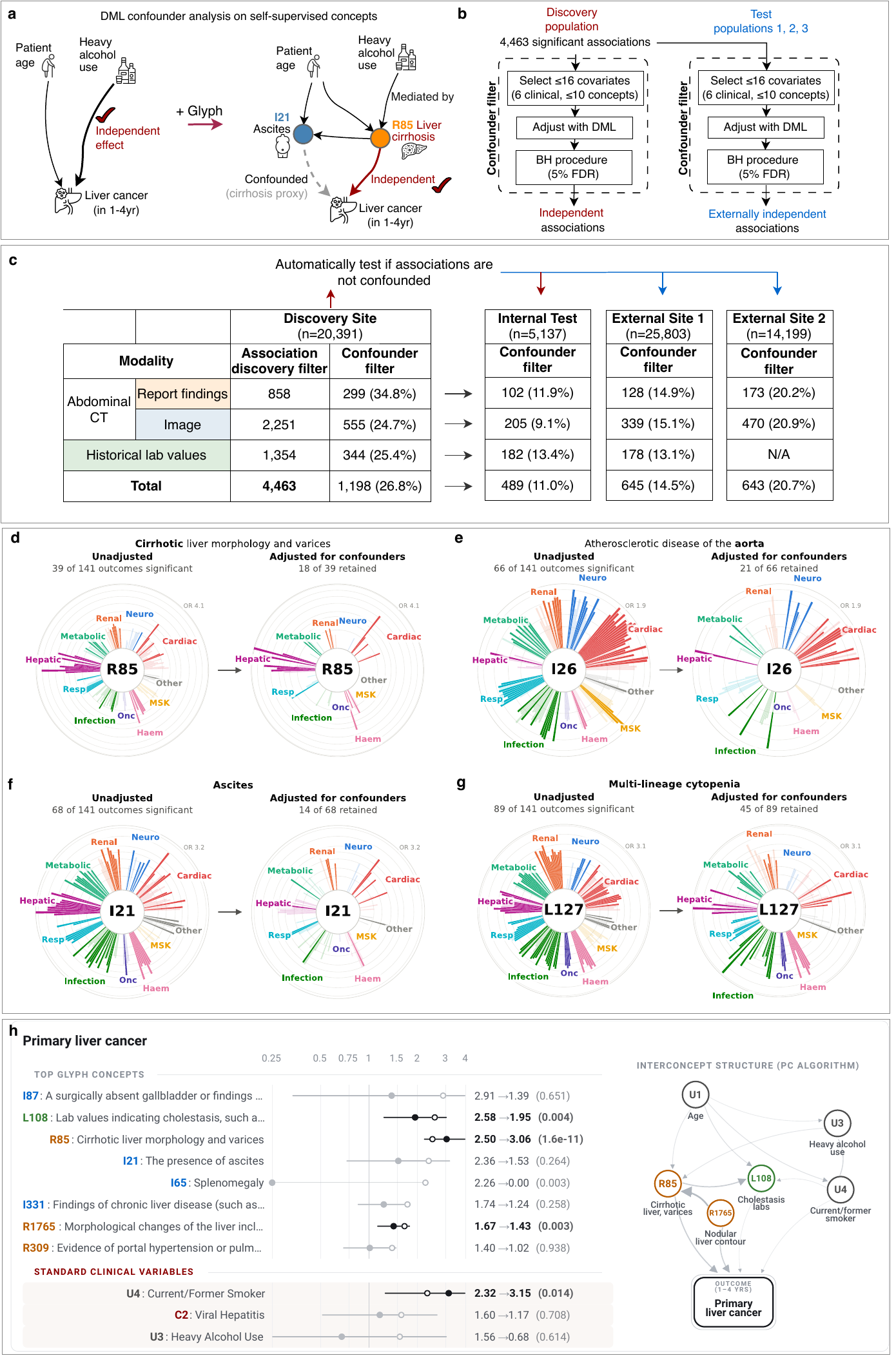}
    \caption{Confounder-adjusted association study.
    \textbf{a} Glyph concepts act as variables that can be compared with one another and with established clinical variables in a unified feature space, enabling confounder analysis of FM predictions.
    \textbf{b} For each of the 4,463 promising associations, the ten most strongly associated concepts across other modalities are automatically selected as candidate confounders, alongside six universal clinical covariates.
    \textbf{c} Number of associations per modality retaining a significant, adjusted OR $> 1.0$ after confounder adjustment per site.
    \textbf{d--g} Association profiles show that after adjustment I21 (ascites) loses most associations, L127 (cytopenia) remains broadly associated, R85 (cirrhotic liver morphology and varices) retains hepatic associations, and I26 (aortic atherosclerosis) retains cardiovascular and neurocognitive associations.
    \textbf{h} An example of a knowledge-guided confounder analysis for primary liver cancer.}
    \label{figure:4}
\end{figure}

\subsection{Glyph enabled confounder analysis to identify independent disease markers}
\label{section:results:confounder_study}
Having identified significant, reproducible disease markers, we assessed whether they persisted after adjustment by potential confounders. 
A concept encoding ascites may appear to predict liver cancer because underlying cirrhosis can cause both fluid accumulation and an increased risk of liver cancer (see Figure \ref{figure:4}a).
The majority of FM outputs are only tested for predictive correlation to an outcome, and rarely test if this prediction is potentially confounded.
This extends the scientific reach of foundation models from correlation towards independence analysis.
Glyph concepts, due to being both interpretable and sparse, can be tested for independence alongside known clinical covariates and risk markers in confounder analysis frameworks.
As in the association study, these results identify associations that persist after adjustment for measured clinical variables and other concepts, rather than establish causality.

We adjusted each significant association for confounders by applying double machine learning (DML) (see Figure \ref{figure:4}b and Methods, Section~\ref{section:methods:dml}).
16 confounders were drawn per-outcome, including the top 10 significantly associated concepts, and six universal clinical covariates covering age, sex, race, BMI, alcohol use, and smoking.
Applying the \textit{Confounder filter} to the 4,463 promising associations on the Discovery Site, 1,198 (26.8\%) retained an adjusted OR $> 1.0$ (Figure~\ref{figure:4}c).
Fewer associations were retained at the Internal Test Site and External Sites \#1 and \#2, potentially reflecting lower statistical power, variation in FM reproducibility, and differences in population distributions.
At External Site \#2, 643 of 3,109 associations (20.7\%) survived the \textit{Confounder filter}.
In the following, we study which significant associations withstood adjustment, while also noting groups of concepts that were most heavily attenuated. 
Unless otherwise specified, results refer to the Discovery Site.

\textbf{Independence of hepatic disease risk markers:}
R85 (cirrhotic liver morphology and varices) was associated with 9 of 11 hepatic outcomes at the Discovery Site, and 6 of these 9 remained significant after confounder adjustment (see Figure \ref{figure:4}d).
Among these were known consequences of cirrhosis, including portal hypertension (adjusted OR 2.46 $p=3.0\times10^{-4}$, 2.35 $p=1.4\times10^{-1}$, 3.10
$p=1.2\times10^{-3}$, 1.77 $p=2.3\times10^{-9}$), liver abscesses (2.98 $p=1.5\times10^{-16}$, 3.15 $p=2.7\times10^{-8}$, 3.19 $p=1.1\times10^{-7}$, 1.87
$p=1.4\times10^{-11}$), primary liver cancer (4.07 $p=1.2\times10^{-71}$, 2.13 $p=4.1\times10^{-7}$, 1.79 $p=1.9\times10^{-8}$, 1.68 $p=2.4\times10^{-7}$) and liver
transplantation (2.11 $p=1.6\times10^{-8}$, 2.06 $p=1.0\times10^{-7}$, 2.15 $p=1.0\times10^{-10}$, 0.99 $p=9.8\times10^{-1}$), with reproduction across all three held-out cohorts in the majority of cases.
By contrast, I21 (ascites) and I25 (ascites, anasarca or pleural effusion) retained only one of 10 and none of eight Discovery Site associations, respectively, after adjustment that included R85 (liver cirrhosis).
This pattern is consistent with fluid accumulation being less specific to hepatic disease than cirrhotic morphology and varices~\cite{de2022baveno}.

\textbf{Independence of renal disease risk markers:}
As in hepatic disease, confounder adjustment concentrated the renal signal on known markers of renal disease.
L84 (azotemia) retained 8 of its 11 Discovery Site renal associations, covering chronic renal failure, stage IV disease, kidney transplantation, oliguria and anuria, hypertensive chronic kidney disease, secondary hyperparathyroidism, acute renal failure and the composite chronic kidney disease outcome, with effects broadly unchanged by adjustment (ORs 1.71--2.53 vs. adjusted ORs 1.27--2.95).
L69 (impaired renal function) retained 6 of its 7 associations, all renal, at consistently smaller effects (ORs 1.24--1.44 vs. adjusted ORs 1.45--1.85), remaining highly concentrated on the kidney.
The broader I109 (aortic calcification) and L127 (cytopenia) markers retained 4 of 10 and 4 of 12 renal associations, respectively, indicating a mixed profile.
At Discovery, R398 (kidney atrophy) retained its sole renal association with hypertensive chronic kidney disease after adjustment. R302 (polycystic kidney disease) retained associations with stage III chronic kidney disease and kidney transplantation, but not oliguria and anuria.
In contrast, markers of fluid accumulation and general illness lost their renal links almost entirely, with I21 (ascites) falling from 7 significant renal associations to just 3.

\textbf{Independence of cardiovascular risk markers:}
Similarly, R21 (calcification of the coronary arteries and cardiac valves) lost 32 of its 38 associations after adjustment, including those with hypertension and atrial flutter across all cohorts.
I26 (aortic atherosclerosis) retained 11 of 21 cardiovascular and 4 of 7 neurocognitive associations after adjustment, while most associations with other outcome groups were attenuated (Figure~\ref{figure:4}e).
Some vascular–musculoskeletal associations attenuated after adjustment, while others persisted, including R21 (coronary/valvular calcification) with femoral fracture and chondrocalcinosis, I106 (atherosclerosis with colonic pathology) with femoral fracture, and R420 (coronary calcification/atherosclerosis) with chondrocalcinosis.
We then tested R21 in the discovery cohort using a radiologist-specified set of cardiovascular risk factors: age, sex, smoking, alcohol use, BMI, diabetes, hypertension, chronic kidney disease, sleep apnea, and statin use extracted from medication orders.
Moreover, we grouped atherosclerosis and calcification concepts within each modality.
The imaging-derived group (5 concepts combined) remained associated with future dementia after adjustment (OR 2.28 vs. adjusted OR 1.41, $p=5.2\times10^{-6}$), whereas the report-derived group (7 concepts) did not (1.61 vs. 0.93, $p=0.44$).
This suggests that imaging captured information about future dementia risk beyond that recorded in radiology reports.

\textbf{Independence of systemic risk markers:}
Many concepts previously identified as encoding systemic risk were revealed to be largely confounded. 
The most heavily discounted associations related to fluid accumulation.
Specifically, I21 (ascites) lost 54 of its 68 (79\%) associations under adjustment, among the highest attrition of any concept and well above the cohort-wide average of 73\% (see Figure \ref{figure:4}f).
Similarly, R101 (pleural effusion and ascites) lost 90\% and I25 (combined ascites, anasarca, and effusion) lost 82\%.

In contrast, I87 exhibited the largest number of associations both before and after automated confounder adjustment.
It retained 52 of its 96 associations (54\%) spanning all 11 outcome groups, including 8 of 8 respiratory, 9 of 18 cardiovascular and 7 of 11 hepatic associations.
I87 was most strongly correlated with prior cholecystectomy, but targeted analyses suggested that this did not explain its risk profile.
As shown in Figure \ref{figure:extended:I87}, associations persisted in targeted analyses stratified by additional potential confounders, including cirrhotic liver morphology, cholestasis, liver transplantation, cholecystectomy and portal hypertension, considered individually and in combination, suggesting that these factors do not fully explain I87’s prognostic signal.
These findings raise the possibility that Merlin has learned a previously unrecognized imaging phenotype associated with systemic disease risk, whose clinical meaning and potential significance warrant further characterization.

Another image-based finding, I65 (splenomegaly) retained 30 of its 48 associations (63\%) after adjustment, spanning eight outcome groups.
These included subsequent primary liver cancer, consistent with the association between incidentally detected splenomegaly and later liver cancer diagnosis reported in a recent population-based cohort study \cite{juhl2026incidentally}.
This association persisted across all four cohorts after adjustment for the clinical variables and concepts reflecting liver dysfunction, coagulopathy and ascites, but was attenuated when adjusting for cirrhosis and other liver-related findings in a targeted analysis (as shown in Figure \ref{figure:4}h).

Some systemic labs markers maintained a majority of their associations after adjustment.
Specifically, L108 (cholestasis) maintained 49 of its 87 associations (56\%), covering 9 of the 11 outcome groups.
Abnormal liver tests have been linked to haemodynamic status in patients with established heart failure~\cite{van2010abnormal}, but did not assess whether cholestasis preceded later cardiac diagnoses, as observed here.
L127 (cytopenia) retained 45 of 89 associations (51\%) after adjustment (see Figure \ref{figure:4}g). 
A prospective population study linked lymphopenia, a subtype of cytopenia, to later infection, including sepsis ~\cite{warny2018lymphopenia}.
L127 captures a broader cytopenia pattern associated with outcomes across disease groups.
L84 (azotemia) was concentrated to a more specific risk profile.
It retained 28 of 76 associations (37\%), with associations attenuating in respiratory, cardiovascular and oncologic disease but persisting for renal, metabolic, and haematologic outcomes.
In contrast, all concepts involving hypoalbuminaemia (L34, L59, L77, L82, L87, L91, L124) were heavily attenuated, retaining only 0--17.1\% of their associations.
Hypoalbuminaemia is an established independent predictor of mortality \cite{goldwasser1997association}. 
However analysis in our cohorts is more consistent with the role of hypoalbuminaemia as a manifestation of underlying illness \cite{don2004poor}.

\textbf{Independence of infection risk markers:}
At Discovery, L127 (cytopenia) retained 10 of its 14 infection associations after adjustment, compared with 8 of 14 for L108 (cholestasis).
L91 (severe hypoalbuminaemia) retained associations with sepsis and bacteremia.
I54 (intra-abdominal or pelvic abscess) retained associations with bacteremia and drug-resistant infections, indicating that localized infectious findings retained prognostic information beyond accompanying abnormalities.

\textbf{Independence of metastatic sequelae:} 
R64 (metastatic disease) maintained adjusted associations with cachexia~\cite{biswas2020understanding} at Discovery and Internal Test, acute renal failure~\cite{launay2007prevalence} at External Site \#1, and sepsis~\cite{williams2004hospitalized} across these three cohorts, consistent with reported systemic sequelae of advanced cancer.

\begin{figure}[htbp]
    \centering
    \includegraphics[width=0.9\textwidth]{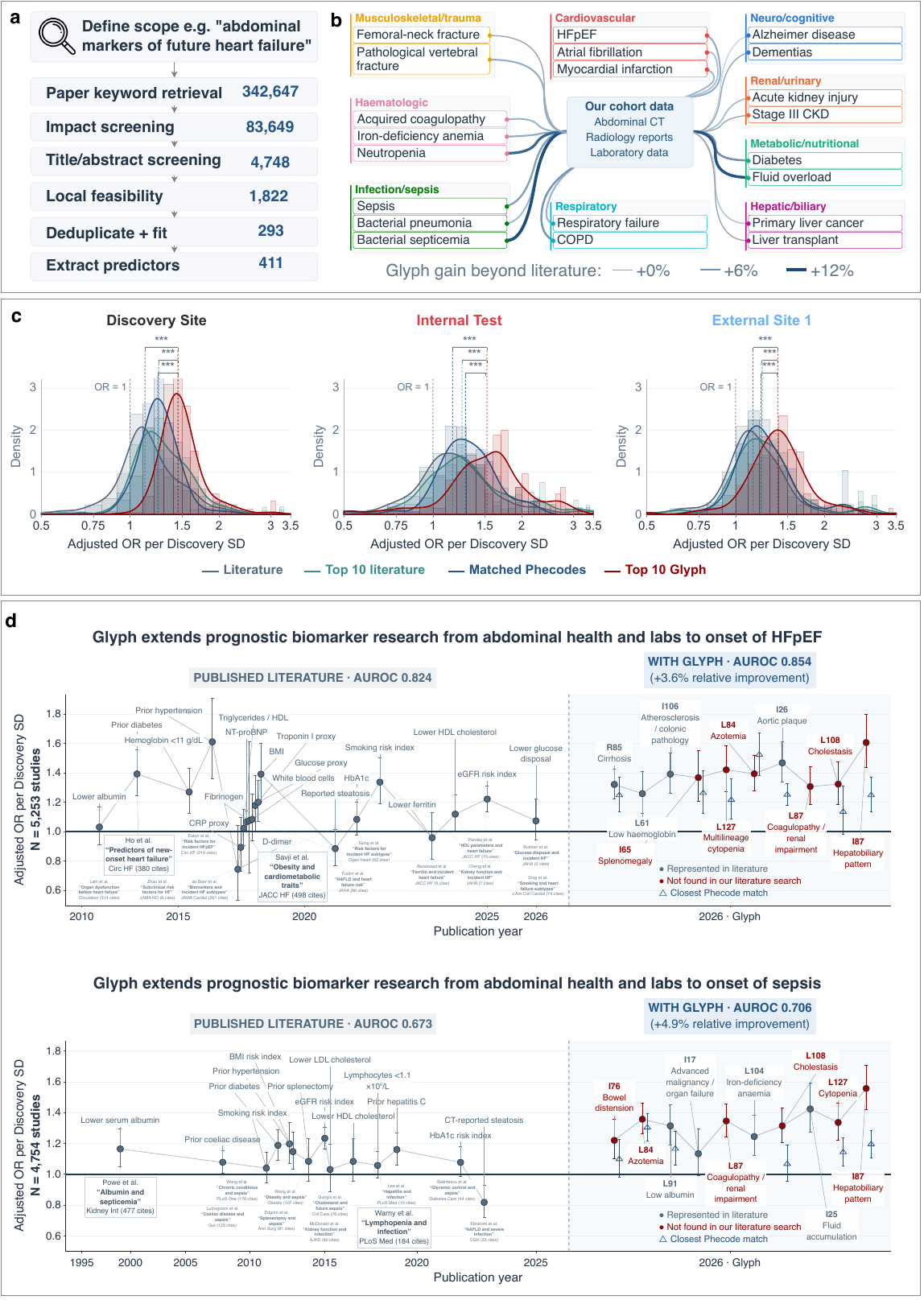}
    \caption{\textbf{Glyph identifies associations beyond the clinical literature and adds prognostic value.}
    \textbf{a} Screening 342,647 PubMed papers across 21 outcomes yielded 293 representative studies and 411 literature-derived predictors.
    \textbf{b} Adding ten Glyph concepts per outcome improved held-out AUC beyond literature markers, with the largest gains in metabolic, haematologic, and infectious outcomes.
    \textbf{c} On average, Glyph concepts outperform literature markers, even when top performing markers are selected on the Discovery Set.
    \textbf{d} Glyph concepts provide discoveries that complement published clinical research.
    The additional prognostic benefit varied across axes, as illustrated by the improvement for HFpEF (top) and the larger gain for sepsis (bottom).
    }
    \label{figure:5}
\end{figure}

\subsection{Glyph finds novel associations beyond the clinical literature}
\label{section:results:systematic_literature_comparison}

Having established that Glyph concepts identify significant, reproducible markers of disease risk, we next asked whether they represent novel discoveries beyond those reported in the clinical literature.
Specifically, we evaluated 21 future outcomes spanning multiple disease groups.
To characterize published abdominal imaging and laboratory markers for these outcomes, an agent acting as a co-scientist systematically identified and filtered 342,647 unique publications using the PubMed API \cite{sayers2009eutilities} and Paperclip \cite{gxl_paperclip_2026} (see Methods for additional details).
This yielded 411 literature-derived markers that could be tested in our data, averaging 19.6 per axis, which we then systematically re-evaluated.
These included both singular measurements, such as low albumin, and composite clinical predictors such as FIB-4 \cite{blanco2021assessing}, which combines age, transaminases and platelet count.
Finally, to assess whether Glyph concepts add prognostic value beyond these markers, we selected the top ten Glyph concepts per outcome using evidence from the Discovery Site, excluding concepts that primarily reflect care intensity such as R399 (pacemaker).
We then fit one model using the literature-derived markers alone and another combining them with the selected Glyph concepts to test whether Glyph concepts identified prognostic information beyond published markers.

Results of percentage gains in AUC are presented in Figure \ref{figure:5}b.
Both the literature baseline and Glyph enhanced model were fit at the Discovery Site, and evaluated on the pooled Internal Test and External Site \#1 cohorts.
We found that, per-outcome, literature baselines achieved 0.749 AUC (95\% CI 0.737--0.761) on held-out data, and that ten Glyph concepts significantly improved this to 0.774 (95\% CI 0.762--0.785, $p=2.66\times10^{-28}$).
Observed benefits were most concentrated in metabolic, haematologic and infectious outcomes, averaging a relative AUROC improvement of 6.64\% (95\% CI 5.26--8.01\%).
These groups also had 5.0 fewer papers on average (10.9 vs 15.9) than the other groups evaluated, indicating they may be under-researched.

We then explored which concepts were driving gains over literature markers.
The top ten Glyph concepts per outcome, selected at the Discovery Site, reproduced more frequently in the pooled held-out cohort (64.8\%) than the literature-derived marker set (28.2\%, $p=1.86\times10^{-18}$), the top ten performing literature markers (41.2\%, $p=2.54\times10^{-6}$), and corresponding Phecodes (54.8\%, $p=0.062$).
After adjustment for age, sex and race, reproduction required BH-adjusted $q<0.05$ and the same effect direction as at the Discovery Site.
Median confounder-adjusted odds ratios in the pooled evaluation cohort were also significantly higher for Glyph (1.31) than for the literature-derived marker set (1.10, $p=5.0\times10^{-5}$), the top ten
literature markers (1.16, $p=5.0\times10^{-5}$), and corresponding Phecodes (1.19, $p=5.0\times10^{-5}$; Figure~\ref{figure:5}c).
A direct comparison between Glyph concepts and all Phecodes is provided in Supplementary Section~\ref{supplementary:phecodes_comparison} and Figure~\ref{figure:extended:phecode_baseline}.
Overall, this suggests that Glyph captures prognostic information in abdominal CT cohorts, accessible through FM embeddings, that is either underrepresented in the established clinical literature, or is difficult to measure with conventional clinical markers.

To review specific axes in detail, we plotted markers according to their publication dates and assessed which of the ten Glyph concepts were represented among the literature-derived markers.
Adding Glyph concepts did not significantly improve AUROC across every axis.
For example, no significant improvement was observed for dementia (0.871 to 0.870, $p=0.80$) or Alzheimer's disease (0.840 to 0.842, $p=0.71$), suggesting little to no additional prognostic information beyond the literature-derived markers for these outcomes.
In contrast, for heart failure with preserved ejection fraction (HFpEF; Figure~\ref{figure:5}d), Glyph identified risk markers not represented among the literature-derived markers, including I65 (splenomegaly), L84 (azotemia), L127 (cytopenia), L108 (cholestasis), and I87 (uncharacterized hepatobiliary pattern).
Adding Glyph concepts significantly improved AUROC by 3.6\% from 0.824 to 0.854 ($p=4.7\times10^{-7}$).
Similarly, along the abdominal health to sepsis axis, adding Glyph concepts significantly increased AUROC by 4.9\% from 0.673 to 0.706 ($p=3.2\times10^{-7}$; Figure~\ref{figure:5}e).
These gains demonstrate additional prognostic information beyond the evaluated literature-derived markers, including contributions from imaging findings not routinely documented in reports, such as I76.

\begin{figure}[!ht]
    \centering
    \includegraphics[width=0.95\textwidth]{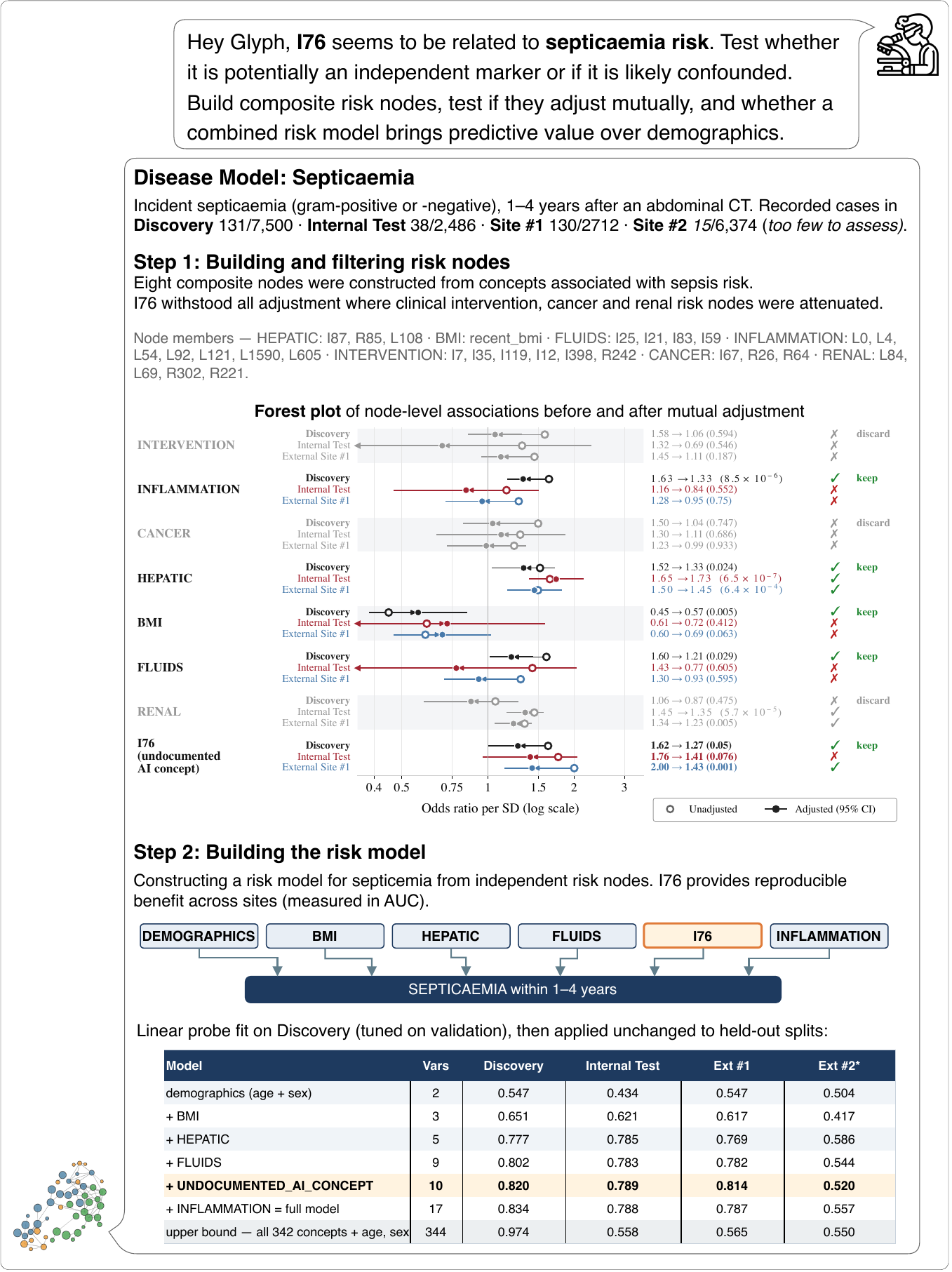}
    \caption{The MCP enables more complex analyses that build on a foundational layer of Glyph concepts. In this example, concepts were combined into nodes that competed against one as independent predictors for inclusion in a disease model of septicemia. I76 (an undocumented imaging concept linked to septicemia) provided a predictive contribution independent of hepatic disease markers, fluid accumulation, BMI, inflammatory markers, malignancy, and evidence of clinical intervention, and contributed to prognostic AUC predictive across three cohorts.}
    \label{figure:I76_workflow}
\end{figure}

\subsection{Glyph MCP provides an interface for open-ended investigation}

\begin{figure}[!ht]
        \centering
        \includegraphics[width=0.9\textwidth]{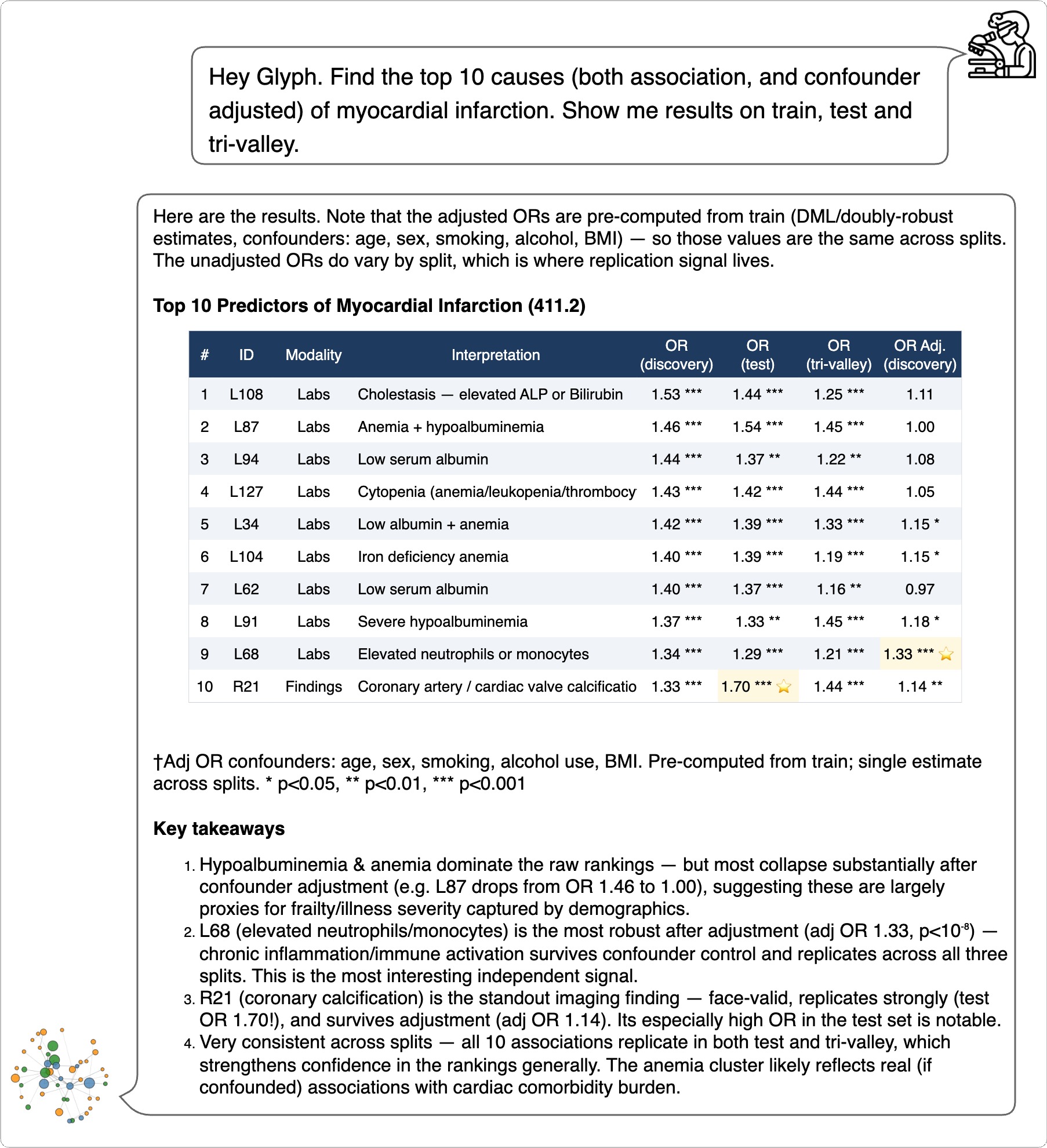}
        \caption{The Glyph MCP brings a new capability; the ability to immediately perform, user-specified requests about high-dimensional unstructured data, such as imaging. Here, Claude Code returns the most highly associated concepts for future risk of myocardial infarction, the majority of which are labs concepts. Responses are grounded by the Glyph layer.}
        \label{figure:extended:1}
    \end{figure}

Finally, we demonstrate how Glyph concepts support open-ended, agentic discovery via the Glyph MCP, through a set of illustrative workflows for agents and clinicians. The MCP enables more complex, user-guided discovery workflows. \textbf{Targeted disease modeling:} Beyond automatic discovery, Glyph concepts can be manually assembled into composite features to test a user-defined hypothesis about a disease, for example testing whether an undocumented imaging concept repeatedly linked to septicemia is confounded by body composition or existing disease markers (Figure~\ref{figure:I76_workflow}). \textbf{Concept ranking for a target outcome:} A user can query the concept space directly in natural language, for example asking for the top associations for future myocardial infarction; the Glyph MCP resolves this to the underlying outcome and returns the concepts with the largest prognostic association among those surviving Bonferroni correction, predominantly drawn from laboratory panels in this example (Extended Figure~\ref{figure:extended:1}).

\begin{figure}[!ht]
        \centering
        \includegraphics[width=0.9\textwidth]{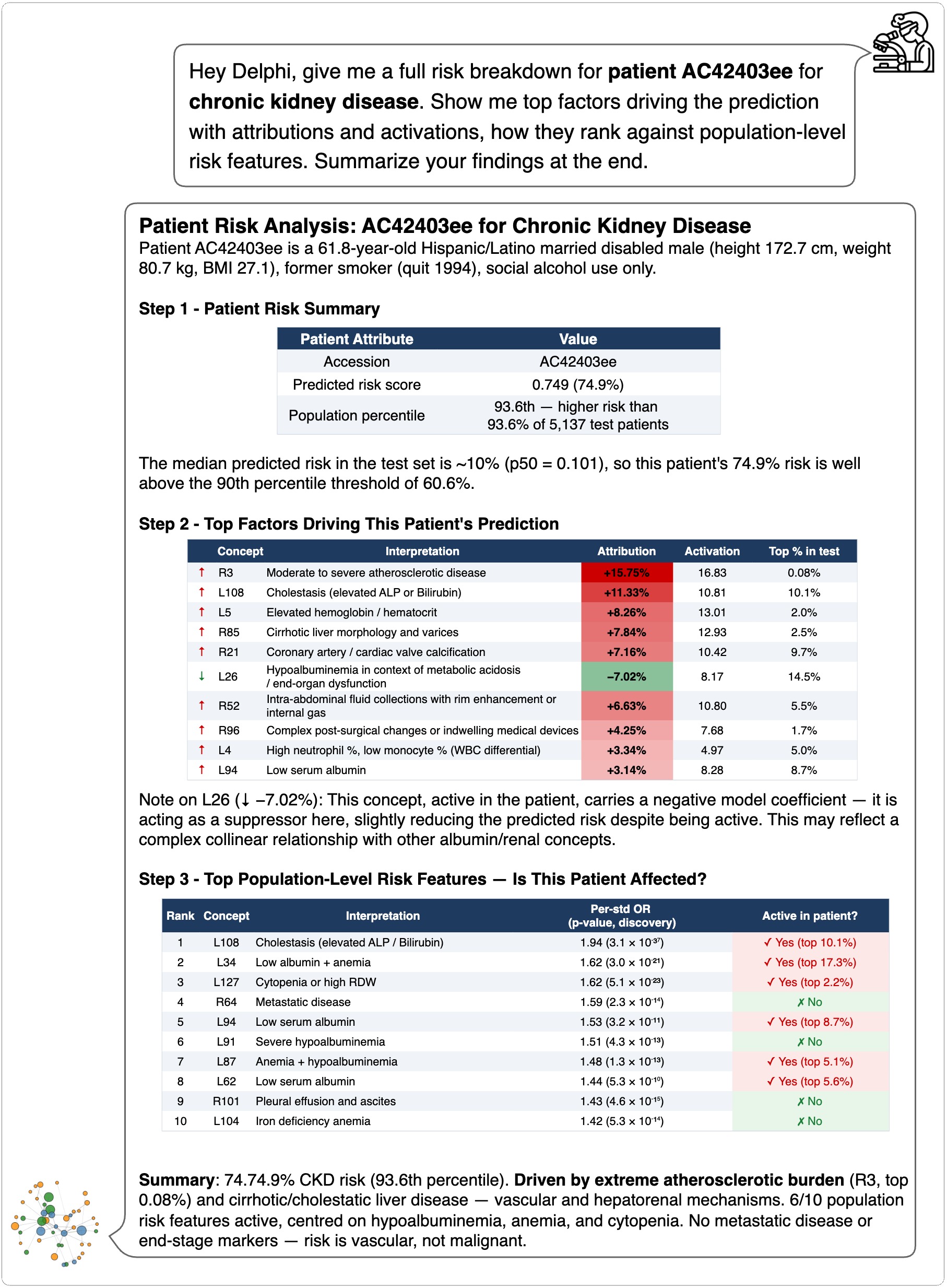}
        \caption{Glyph generates individualised risk explanations for a given outcome. For a query patient (AC42403ee, 74.9\% predicted CKD risk, 93.6th percentile), the system surfaces the concepts most responsible for that prediction alongside their attribution scores, activation levels, and population rarity. Cross-referencing against the top population-level risk features reveals which epidemiologically validated signals are active in this patient, grounding the individual explanation in cohort-level evidence.}
        \label{figure:extended:4}
    \end{figure}

\begin{figure}[!ht]
        \centering
        \includegraphics[width=0.9\textwidth]{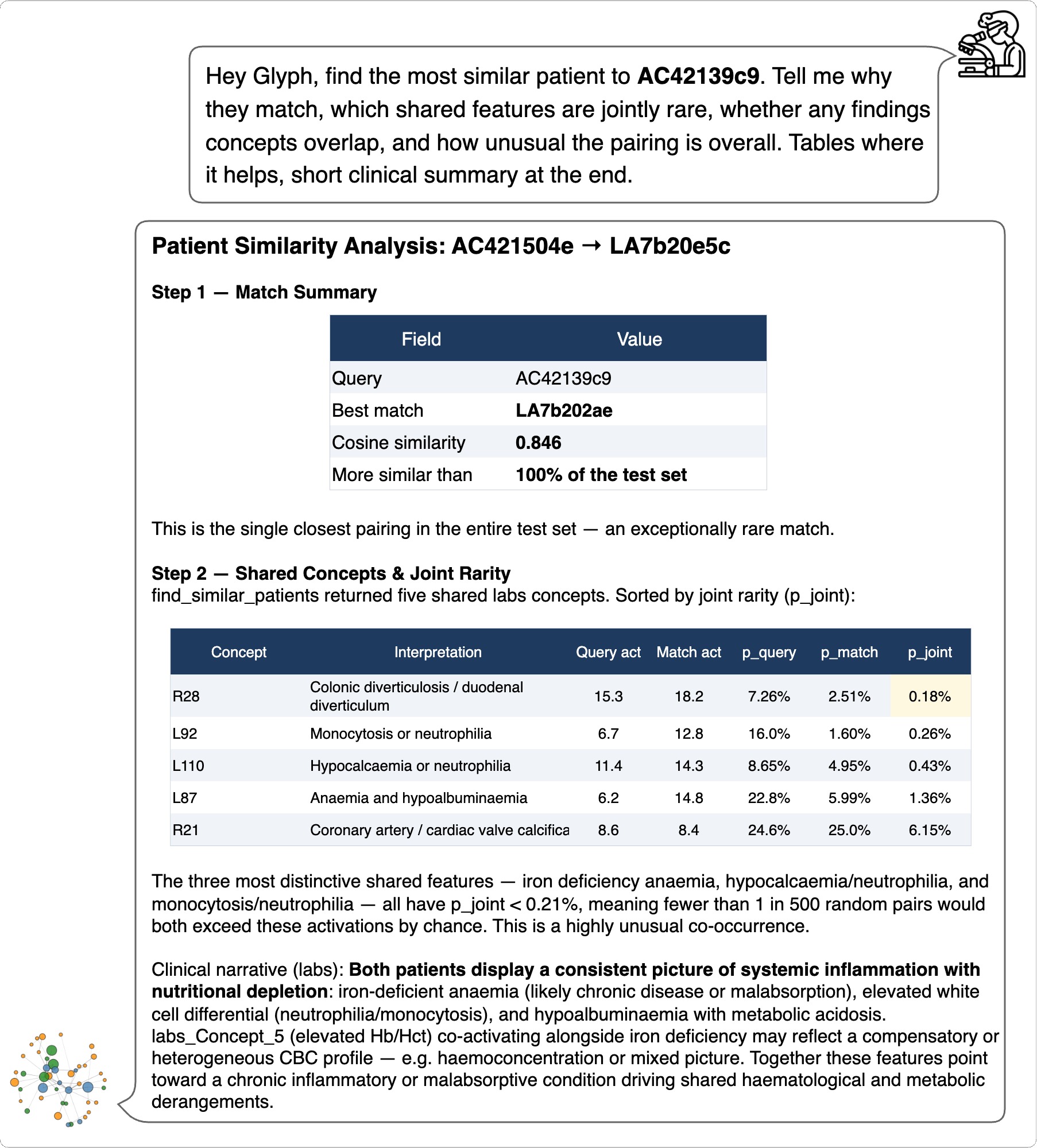}
        \caption{The Glyph concept layer is flexible to other forms of discovery beyond prognosis. Here, Glyph performs a patients-like-me scenario where the most similar patient to a query is found. Their similarity is then explained in terms of shared concepts, and the rarity of such a co-occurrence at their respective levels of activation.}
        \label{figure:extended:2}
    \end{figure}

Outside of discovery, Glyph also brings explainability to typical use of foundation models: \textbf{Individualized risk explanation:} Beyond population-level associations, Glyph concepts can explain an individual patient's risk prediction, attributing it to the specific concepts driving that prediction and contextualizing them against population-level evidence (Extended Figure~\ref{figure:extended:4}). \textbf{Patients-like-me:} Given a query patient, Glyph retrieves the most similar patient in the cohort and explains their similarity in terms of shared concepts and the rarity of their joint activation, illustrating how the concept space supports case-based reasoning beyond simple embedding distance (Extended Figure~\ref{figure:extended:2}).

\section{Discussion}
In this paper we introduced Glyph, a method for transforming foundation model (FM) embeddings into engines of scientific discovery.
At its core, Glyph supposes that FM embeddings, trained on large volumes of clinical data, encode established clinical markers alongside unexplored features that could reveal new insights into disease mechanisms.
To extract these features, Glyph builds on mechanistic interpretability and establishes their value as coded, interpretable clinical concepts for scientific investigation.
In our dataset, the resulting set of extracted concepts enabled rediscovery of established organ-specific and systemic risk markers associated with the onset of future disease.
In addition, we were able to automatically establish that many of these findings were reproducible across three sites without model retraining.
Transforming embeddings into separate, interpretable and measurable concepts enabled adjustment for potential confounders identified directly from the data, beyond routinely collected covariates or those selected from prior clinical knowledge.
Applying Glyph on large clinical datasets revealed complementary associations and risk factors not reported in established clinical literature.
Moreover, complementing literature-derived markers with Glyph concepts significantly improved prognostic prediction across several disease axes.
Together, these findings demonstrate that interpretability can extend beyond a mechanistic understanding of neural networks, enabling features learned implicitly by FMs to interact directly with existing clinical knowledge and available clinical outcomes.

Glyph rediscovered known patterns of disease progression across hepatic, renal, and cardiovascular axes.
It also recovered known cross-system associations, e.g. linking renal decline to metabolic comorbidities or valvular atherosclerosis to cardiovascular and neurological disease.
Our observations also reflected the complementary strengths of different clinical modalities.
Specifically, laboratory concepts more consistently captured renal disease progression than kidney-related imaging findings.
Similarly, imaging-derived findings often showed more reproducible associations than findings documented in radiology reports, particularly for atherosclerotic and hepatic risk markers.
Confounder adjustment further characterized discovered associations by attenuating associations known to involve nonspecific markers of illness severity, such as fluid accumulation and hypoalbuminemia.
It also emphasized published prognostic relationships, including associations of imaging-derived vascular findings with neurological outcomes \cite{pickhardt2020automated}, renal findings with downstream cardiovascular outcomes \cite{ndumele2023cardiovascular}, and the renal consequences of cholestatic liver disease \cite{ghallab2025enteronephrohepatic}.
Together, the recovery of known disease relationships in patients undergoing abdominal CT supported the validity of the concept space as a foundation for the exploration of novel hypotheses.

Beyond rediscoveries, Glyph expanded the clinical picture of disease risk, revealing prognostic information beyond what established literature-derived markers capture.
In particular, comparisons across modalities highlighted laboratory markers such as cholestasis and cytopenia.
After adjustment, these markers retained broad systemic risk profiles, while other laboratory concepts showed more specific associations.
Azotemia, for example, retained a predominantly renal, metabolic, and hematological risk profile.
Beyond laboratory markers, Glyph uncovered previously uncharacterized imaging phenotypes within CT embeddings, including a mixed hepatobiliary pattern whose adjusted associations remained significant across all outcome groups.
Furthermore, complementing literature-derived markers with a selection of top performing Glyph concepts reproducibly improved prognostic prediction, particularly in metabolic, hematological and infectious outcomes, indicating that existing published markers lack prognostic information presently available in abdominal CT and laboratory data.
Together, these findings demonstrate a practical approach to exploring a systems-level perspective on disease \cite{barabasi2011network,barnett2012epidemiology} using routinely acquired clinical data.
They also motivate evaluating several laboratory markers and imaging phenotypes as potential targets for opportunistic or incidental screening.

Comparison with coded phenotypes further characterized the clinical information captured by Glyph.
Phecodes, which group ICD codes into clinically meaningful categories, provide an established basis for automated PheWAS discovery \cite{denny2013systematic}.
In some cases, directly matched Phecodes showed the same disease associations as Glyph concepts, but with overall smaller effects that often did not reach significance.
For these clinical findings, the stronger signal captured by Glyph enabled both rediscoveries and candidate discoveries that were not detected using diagnosis codes alone.
Where Phecodes yielded stronger associations, they reflected related conditions captured in the broader clinical record, beyond the scope of abdominal CT and laboratory data.
Across the selected outcomes, this broader coverage yielded more significant associations overall, whereas Glyph associations had larger median effects and higher reproduction rates.
Together, these findings suggest that Glyph complements the coded record by making additional phenotypic information in imaging and laboratory data accessible for discovery.

Several clinical and technical considerations are important in interpreting our findings.
Primarily, our selection criteria restricted the cohorts to patients undergoing abdominal CT across acute and outpatient settings.
This selection likely enriched the study population for abdominal illness, influencing both the associations detected and their estimated prognostic value.
For example, enrichment for abdominal disease in this cohort may have contributed to the prominence of markers of hepatobiliary disease and renal dysfunction, such as cholestasis and azotemia.
In addition, enrichment for abdominal disease could partly explain why hepatobiliary abnormalities featured among the strongest adjusted markers of cardiovascular outcomes, alongside established vascular, renal and haematological risk markers.
Future work could reproduce these analyses in broader patient populations to assess whether the observed associations and relative prognostic value of these markers are preserved.

Furthermore, outcome data used in our study warrant further consideration.
We used Phecodes derived from ICD codes as outcome data to standardize evaluation across institutions \cite{denny2013systematic}.
However, such coded data may introduce noise and uncertainty, e.g. regarding the precise timing of observed disease patterns.
Consequently, some associations may capture existing but undocumented disease rather than risk of future disease onset, while cases never assigned the corresponding code may be missed entirely.
Refining outcome labels through clinician adjudication of longitudinal records against predefined diagnostic criteria could correct diagnosis and timing, but would require extensive chart review, potentially supported by automated search of clinical notes.
Ultimately, prospective studies using predefined diagnostic criteria and systematic follow-up would be required for stronger validation of these outcomes.

Glyph enabled systematic confounder adjustment across a broad set of clinical concepts.
Associations that persist after adjustment indicate that related concepts provide prognostic information beyond clinical covariates and other identified concepts, but do not establish that they represent independent disease processes or causal drivers of disease on a retrospective dataset.
Broader clinical coverage, including additional imaging, clinical notes, categorical laboratory results, flowsheets, medication use, and genetic factors, could help account for omitted factors and reveal further confounders and candidate markers.
Furthermore, the fixed correlation threshold used in our analysis was designed to exclude potential mediators, downstream consequences through which the tested concept could affect the outcome, but adjustment results were sensitive to the chosen threshold.
Future work should guide confounder selection using knowledge of disease mechanisms, rather than statistics alone \cite{vanderweele2019principles}.

Future work could extend the systematic comparison with published literature presented in Section~\ref{section:results:systematic_literature_comparison} beyond the selected outcomes.
Although it requires substantial token usage, this assessment provides a framework for distinguishing established relationships recovered by Glyph, published associations not detected in our cohorts, and potentially novel findings.
Future work could also investigate how concepts change over time and how their combinations contribute to disease risk.
The analyses shown in this work related concepts measured around individual CT examinations to future outcomes.
Repeated measurement could reveal when concepts emerge, how they evolve, and how their relationships change through disease progression, providing temporal evidence to guide causal investigation.
Combined with more complete treatment histories, these trajectories could support investigation of changes associated with clinical intervention.
Changes in concept activation could also be evaluated as quantitative indicators of disease state or treatment response by comparing their sensitivity with existing clinical measures.
Beyond these temporal extensions, our PheWAS analysis tested concepts individually and did not assess their combined prognostic information.
Future work could examine combinations and interactions between concepts to identify more specific clinical phenotypes and risk associations not apparent from individual markers.

Technical considerations and insights arise both from the source embeddings and from how their features are extracted and interpreted.
Text embeddings yielded finer-grained concepts than image embeddings, even though radiology reports provide a compressed description of the images.
This observation motivates future work in designing image encoders for fine-grained concept discovery, alongside methods for resolving more specific findings within existing embeddings.
We also extracted embeddings separately by modality, which simplified attribution of concepts to their source and examination of relationships across modalities.
However, some predictive patterns may be inherently multimodal, motivating joint embeddings that could capture combinations of imaging, laboratory, and report findings not fully represented in any single modality.
One insight from our confounder analyses is that prediction models built on FM embeddings can draw on both potentially actionable markers and confounded proxies for illness.
Risk scores derived from the full FM embedding may therefore identify patients at risk without reliably guiding how that risk can be reduced.

Textual interpretations describe the clinical findings associated with a feature, but do not necessarily establish its precise clinical meaning.
In a subset of extracted concepts, automated interpretation required a combination of a small number of distinct clinical or imaging findings, suggesting that not all concepts were monosemantic.
It remains unclear whether this reflects \textit{polysemanticity} in the FM embeddings, failure of the SAE to separate distinct features, or a property of underlying data, in which certain features frequently co-occur making feature separation difficult.
Clinical knowledge could guide when related concepts should be grouped or separated into more specific findings, with Glyph providing an efficient basis for refining and testing these definitions.
Future work could also complement our global evaluation with local comparisons against neighboring concepts and clinically similar cases that do not activate the concept in question.
These comparisons could help distinguish defining findings from co-occurring data properties, severity, or clinical context.

In conclusion, Glyph provides a framework for extracting knowledge from foundation models, enabling reproducible, population-scale discovery of disease markers and associations.
Glyph can be applied to any dataset represented by FM embeddings to rapidly investigate a chosen discovery axis, using an agentic co-scientist framework to assess the novelty and added value of extracted findings against the published literature.
This enables researchers to automatically validate data-driven insights, such as those presented here, or discover and share new concept networks and insights from their own datasets.
Overall, we envision that Glyph can accelerate clinical science by establishing a standard for sharing validated FM features as data-driven phenotypes, connecting discoveries across datasets with existing clinical knowledge.

\clearpage
\bibliography{main,clinical}

\begin{thebibliography}{10}
\urlstyle{rm}
\expandafter\ifx\csname url\endcsname\relax
  \def\url#1{\texttt{#1}}\fi
\expandafter\ifx\csname urlprefix\endcsname\relax\def\urlprefix{URL }\fi
\expandafter\ifx\csname doiprefix\endcsname\relax\def\doiprefix{DOI: }\fi
\providecommand{\bibinfo}[2]{#2}
\providecommand{\eprint}[2][]{\url{#2}}

\bibitem{wang2023scientific}
\bibinfo{author}{Wang, H.} \emph{et~al.}
\newblock \bibinfo{journal}{\bibinfo{title}{Scientific discovery in the age of artificial intelligence}}.
\newblock {\emph{\JournalTitle{Nature}}} \textbf{\bibinfo{volume}{620}}, \bibinfo{pages}{47--60} (\bibinfo{year}{2023}).

\bibitem{denny2013systematic}
\bibinfo{author}{Denny, J.~C.} \emph{et~al.}
\newblock \bibinfo{journal}{\bibinfo{title}{Systematic comparison of phenome-wide association study of electronic medical record data and genome-wide association study data}}.
\newblock {\emph{\JournalTitle{Nature biotechnology}}} \textbf{\bibinfo{volume}{31}}, \bibinfo{pages}{1102--1111} (\bibinfo{year}{2013}).

\bibitem{bai2020population}
\bibinfo{author}{Bai, W.} \emph{et~al.}
\newblock \bibinfo{journal}{\bibinfo{title}{A population-based phenome-wide association study of cardiac and aortic structure and function}}.
\newblock {\emph{\JournalTitle{Nature medicine}}} \textbf{\bibinfo{volume}{26}}, \bibinfo{pages}{1654--1662} (\bibinfo{year}{2020}).

\bibitem{gottweis2025towards}
\bibinfo{author}{Gottweis, J.} \emph{et~al.}
\newblock \bibinfo{journal}{\bibinfo{title}{Towards an ai co-scientist}}.
\newblock {\emph{\JournalTitle{arXiv preprint arXiv:2502.18864}}}  (\bibinfo{year}{2025}).

\bibitem{swanson2025virtual}
\bibinfo{author}{Swanson, K.}, \bibinfo{author}{Wu, W.}, \bibinfo{author}{Bulaong, N.~L.}, \bibinfo{author}{Pak, J.~E.} \& \bibinfo{author}{Zou, J.}
\newblock \bibinfo{journal}{\bibinfo{title}{The virtual lab of ai agents designs new sars-cov-2 nanobodies}}.
\newblock {\emph{\JournalTitle{Nature}}} \textbf{\bibinfo{volume}{646}}, \bibinfo{pages}{716--723} (\bibinfo{year}{2025}).

\bibitem{blankemeier2026merlin}
\bibinfo{author}{Blankemeier, L.} \emph{et~al.}
\newblock \bibinfo{journal}{\bibinfo{title}{Merlin: a computed tomography vision--language foundation model and dataset}}.
\newblock {\emph{\JournalTitle{Nature}}} \textbf{\bibinfo{volume}{652}}, \bibinfo{pages}{1318--1328} (\bibinfo{year}{2026}).

\bibitem{xiao2025viral}
\bibinfo{author}{Xiao, Q.} \emph{et~al.}
\newblock \bibinfo{journal}{\bibinfo{title}{Viral oncogenesis in cancer: from mechanisms to therapeutics}}.
\newblock {\emph{\JournalTitle{Signal transduction and targeted therapy}}} \textbf{\bibinfo{volume}{10}}, \bibinfo{pages}{151} (\bibinfo{year}{2025}).

\bibitem{sedlakova2023challenges}
\bibinfo{author}{Sedlakova, J.} \emph{et~al.}
\newblock \bibinfo{journal}{\bibinfo{title}{Challenges and best practices for digital unstructured data enrichment in health research: A systematic narrative review}}.
\newblock {\emph{\JournalTitle{PLOS digital health}}} \textbf{\bibinfo{volume}{2}}, \bibinfo{pages}{e0000347} (\bibinfo{year}{2023}).

\bibitem{schockel2020developments}
\bibinfo{author}{Sch{\"o}ckel, L.} \emph{et~al.}
\newblock \bibinfo{journal}{\bibinfo{title}{Developments in x-ray contrast media and the potential impact on computed tomography}}.
\newblock {\emph{\JournalTitle{Investigative radiology}}} \textbf{\bibinfo{volume}{55}}, \bibinfo{pages}{592--597} (\bibinfo{year}{2020}).

\bibitem{jumper2021highly}
\bibinfo{author}{Jumper, J.} \emph{et~al.}
\newblock \bibinfo{journal}{\bibinfo{title}{Highly accurate protein structure prediction with alphafold}}.
\newblock {\emph{\JournalTitle{nature}}} \textbf{\bibinfo{volume}{596}}, \bibinfo{pages}{583--589} (\bibinfo{year}{2021}).

\bibitem{varadi2022alphafold}
\bibinfo{author}{Varadi, M.} \emph{et~al.}
\newblock \bibinfo{journal}{\bibinfo{title}{Alphafold protein structure database: massively expanding the structural coverage of protein-sequence space with high-accuracy models}}.
\newblock {\emph{\JournalTitle{Nucleic acids research}}} \textbf{\bibinfo{volume}{50}}, \bibinfo{pages}{D439--D444} (\bibinfo{year}{2022}).

\bibitem{merchant2023scaling}
\bibinfo{author}{Merchant, A.} \emph{et~al.}
\newblock \bibinfo{journal}{\bibinfo{title}{Scaling deep learning for materials discovery}}.
\newblock {\emph{\JournalTitle{Nature}}} \textbf{\bibinfo{volume}{624}}, \bibinfo{pages}{80--85} (\bibinfo{year}{2023}).

\bibitem{jensen2014temporal}
\bibinfo{author}{Jensen, A.~B.} \emph{et~al.}
\newblock \bibinfo{journal}{\bibinfo{title}{Temporal disease trajectories condensed from population-wide registry data covering 6.2 million patients}}.
\newblock {\emph{\JournalTitle{Nature communications}}} \textbf{\bibinfo{volume}{5}}, \bibinfo{pages}{4022} (\bibinfo{year}{2014}).

\bibitem{gottweis2026accelerating}
\bibinfo{author}{Gottweis, J.} \emph{et~al.}
\newblock \bibinfo{journal}{\bibinfo{title}{Accelerating scientific discovery with co-scientist}}.
\newblock {\emph{\JournalTitle{Nature}}} \bibinfo{pages}{1--3} (\bibinfo{year}{2026}).

\bibitem{moor2023foundation}
\bibinfo{author}{Moor, M.} \emph{et~al.}
\newblock \bibinfo{journal}{\bibinfo{title}{Foundation models for generalist medical artificial intelligence}}.
\newblock {\emph{\JournalTitle{Nature}}} \textbf{\bibinfo{volume}{616}}, \bibinfo{pages}{259--265} (\bibinfo{year}{2023}).

\bibitem{zhou2023foundation}
\bibinfo{author}{Zhou, Y.} \emph{et~al.}
\newblock \bibinfo{journal}{\bibinfo{title}{A foundation model for generalizable disease detection from retinal images}}.
\newblock {\emph{\JournalTitle{Nature}}} \textbf{\bibinfo{volume}{622}}, \bibinfo{pages}{156--163} (\bibinfo{year}{2023}).

\bibitem{ding2025multimodal}
\bibinfo{author}{Ding, T.} \emph{et~al.}
\newblock \bibinfo{journal}{\bibinfo{title}{A multimodal whole-slide foundation model for pathology}}.
\newblock {\emph{\JournalTitle{Nature medicine}}} \textbf{\bibinfo{volume}{31}}, \bibinfo{pages}{3749--3761} (\bibinfo{year}{2025}).

\bibitem{bricken2023monosemanticity}
\bibinfo{author}{Bricken, T.} \emph{et~al.}
\newblock \bibinfo{journal}{\bibinfo{title}{Towards monosemanticity: Decomposing language models with dictionary learning}}.
\newblock {\emph{\JournalTitle{Transformer Circuits Thread}}}  (\bibinfo{year}{2023}).
\newblock \bibinfo{note}{Https://transformer-circuits.pub/2023/monosemantic-features/index.html}.

\bibitem{gao2024scaling}
\bibinfo{author}{Gao, L.} \emph{et~al.}
\newblock \bibinfo{journal}{\bibinfo{title}{Scaling and evaluating sparse autoencoders}}.
\newblock {\emph{\JournalTitle{arXiv preprint arXiv:2406.04093}}}  (\bibinfo{year}{2024}).

\bibitem{templeton2024scaling}
\bibinfo{author}{Templeton, A.} \emph{et~al.}
\newblock \bibinfo{journal}{\bibinfo{title}{Scaling monosemanticity: Extracting interpretable features from claude 3 sonnet}}.
\newblock {\emph{\JournalTitle{Transformer Circuits Thread}}}  (\bibinfo{year}{2024}).

\bibitem{zhang2025qwen3}
\bibinfo{author}{Zhang, Y.} \emph{et~al.}
\newblock \bibinfo{journal}{\bibinfo{title}{Qwen3 embedding: Advancing text embedding and reranking through foundation models}}.
\newblock {\emph{\JournalTitle{arXiv preprint arXiv:2506.05176}}}  (\bibinfo{year}{2025}).

\bibitem{schuppan2008liver}
\bibinfo{author}{Schuppan, D.} \& \bibinfo{author}{Afdhal, N.~H.}
\newblock \bibinfo{journal}{\bibinfo{title}{Liver cirrhosis}}.
\newblock {\emph{\JournalTitle{The Lancet}}} \textbf{\bibinfo{volume}{371}}, \bibinfo{pages}{838--851} (\bibinfo{year}{2008}).

\bibitem{vaziri2006dyslipidemia}
\bibinfo{author}{Vaziri, N.}
\newblock \bibinfo{journal}{\bibinfo{title}{Dyslipidemia of chronic renal failure: the nature, mechanisms, and potential consequences}}.
\newblock {\emph{\JournalTitle{American Journal of Physiology-Renal Physiology}}}  (\bibinfo{year}{2006}).

\bibitem{oliveira2020co}
\bibinfo{author}{Oliveira, C.~B.} \emph{et~al.}
\newblock \bibinfo{journal}{\bibinfo{title}{Co-occurrence of chronic musculoskeletal pain and cardiovascular diseases: a systematic review with meta-analysis}}.
\newblock {\emph{\JournalTitle{Pain Medicine}}} \textbf{\bibinfo{volume}{21}}, \bibinfo{pages}{1106--1121} (\bibinfo{year}{2020}).

\bibitem{vogel2022hepatocellular}
\bibinfo{author}{Vogel, A.}, \bibinfo{author}{Meyer, T.}, \bibinfo{author}{Sapisochin, G.}, \bibinfo{author}{Salem, R.} \& \bibinfo{author}{Saborowski, A.}
\newblock \bibinfo{journal}{\bibinfo{title}{Hepatocellular carcinoma}}.
\newblock {\emph{\JournalTitle{The Lancet}}} \textbf{\bibinfo{volume}{400}}, \bibinfo{pages}{1345--1362} (\bibinfo{year}{2022}).

\bibitem{de2022baveno}
\bibinfo{author}{De~Franchis, R.} \emph{et~al.}
\newblock \bibinfo{journal}{\bibinfo{title}{Baveno vii--renewing consensus in portal hypertension}}.
\newblock {\emph{\JournalTitle{Journal of hepatology}}} \textbf{\bibinfo{volume}{76}}, \bibinfo{pages}{959--974} (\bibinfo{year}{2022}).

\bibitem{maiwall2020incidence}
\bibinfo{author}{Maiwall, R.} \emph{et~al.}
\newblock \bibinfo{journal}{\bibinfo{title}{Incidence, risk factors, and outcomes of transition of acute kidney injury to chronic kidney disease in cirrhosis: a prospective cohort study}}.
\newblock {\emph{\JournalTitle{Hepatology}}} \textbf{\bibinfo{volume}{71}}, \bibinfo{pages}{1009--1022} (\bibinfo{year}{2020}).

\bibitem{nahon2017bacterial}
\bibinfo{author}{Nahon, P.} \emph{et~al.}
\newblock \bibinfo{journal}{\bibinfo{title}{Bacterial infection in compensated viral cirrhosis impairs 5-year survival (anrs co12 cirvir prospective cohort)}}.
\newblock {\emph{\JournalTitle{Gut}}} \textbf{\bibinfo{volume}{66}}, \bibinfo{pages}{330--341} (\bibinfo{year}{2017}).

\bibitem{foreman2003cirrhosis}
\bibinfo{author}{Foreman, M.~G.}, \bibinfo{author}{Mannino, D.~M.} \& \bibinfo{author}{Moss, M.}
\newblock \bibinfo{journal}{\bibinfo{title}{Cirrhosis as a risk factor for sepsis and death: analysis of the national hospital discharge survey}}.
\newblock {\emph{\JournalTitle{Chest}}} \textbf{\bibinfo{volume}{124}}, \bibinfo{pages}{1016--1020} (\bibinfo{year}{2003}).

\bibitem{li2015bile}
\bibinfo{author}{Li, T.} \& \bibinfo{author}{Apte, U.}
\newblock \bibinfo{title}{Bile acid metabolism and signaling in cholestasis, inflammation, and cancer}.
\newblock In \emph{\bibinfo{booktitle}{Advances in pharmacology}}, vol.~\bibinfo{volume}{74}, \bibinfo{pages}{263--302} (\bibinfo{publisher}{Elsevier}, \bibinfo{year}{2015}).

\bibitem{stevens2024kdigo}
\bibinfo{author}{Stevens, P.~E.} \emph{et~al.}
\newblock \bibinfo{journal}{\bibinfo{title}{Kdigo 2024 clinical practice guideline for the evaluation and management of chronic kidney disease}}.
\newblock {\emph{\JournalTitle{Kidney international}}} \textbf{\bibinfo{volume}{105}}, \bibinfo{pages}{S117--S314} (\bibinfo{year}{2024}).

\bibitem{kirtane2005serum}
\bibinfo{author}{Kirtane, A.~J.} \emph{et~al.}
\newblock \bibinfo{journal}{\bibinfo{title}{Serum blood urea nitrogen as an independent marker of subsequent mortality among patients with acute coronary syndromes and normal to mildly reduced glomerular filtration rates}}.
\newblock {\emph{\JournalTitle{Journal of the American College of Cardiology}}} \textbf{\bibinfo{volume}{45}}, \bibinfo{pages}{1781--1786} (\bibinfo{year}{2005}).

\bibitem{grenier2016radiologic}
\bibinfo{author}{Grenier, N.}, \bibinfo{author}{Merville, P.} \& \bibinfo{author}{Combe, C.}
\newblock \bibinfo{journal}{\bibinfo{title}{Radiologic imaging of the renal parenchyma structure and function}}.
\newblock {\emph{\JournalTitle{Nature Reviews Nephrology}}} \textbf{\bibinfo{volume}{12}}, \bibinfo{pages}{348--359} (\bibinfo{year}{2016}).

\bibitem{leow2021prognostic}
\bibinfo{author}{Leow, K.} \emph{et~al.}
\newblock \bibinfo{journal}{\bibinfo{title}{Prognostic value of abdominal aortic calcification: a systematic review and meta-analysis of observational studies}}.
\newblock {\emph{\JournalTitle{Journal of the American Heart Association}}} \textbf{\bibinfo{volume}{10}}, \bibinfo{pages}{e017205} (\bibinfo{year}{2021}).

\bibitem{detrano2008coronary}
\bibinfo{author}{Detrano, R.} \emph{et~al.}
\newblock \bibinfo{journal}{\bibinfo{title}{Coronary calcium as a predictor of coronary events in four racial or ethnic groups}}.
\newblock {\emph{\JournalTitle{New England Journal of Medicine}}} \textbf{\bibinfo{volume}{358}}, \bibinfo{pages}{1336--1345} (\bibinfo{year}{2008}).

\bibitem{fox2003mitral}
\bibinfo{author}{Fox, C.~S.} \emph{et~al.}
\newblock \bibinfo{journal}{\bibinfo{title}{Mitral annular calcification predicts cardiovascular morbidity and mortality: the framingham heart study}}.
\newblock {\emph{\JournalTitle{Circulation}}} \textbf{\bibinfo{volume}{107}}, \bibinfo{pages}{1492--1496} (\bibinfo{year}{2003}).

\bibitem{pickhardt2020automated}
\bibinfo{author}{Pickhardt, P.~J.} \emph{et~al.}
\newblock \bibinfo{journal}{\bibinfo{title}{Automated ct biomarkers for opportunistic prediction of future cardiovascular events and mortality in an asymptomatic screening population: a retrospective cohort study}}.
\newblock {\emph{\JournalTitle{The Lancet Digital Health}}} \textbf{\bibinfo{volume}{2}}, \bibinfo{pages}{e192--e200} (\bibinfo{year}{2020}).

\bibitem{bos2015atherosclerotic}
\bibinfo{author}{Bos, D.} \emph{et~al.}
\newblock \bibinfo{journal}{\bibinfo{title}{Atherosclerotic calcification is related to a higher risk of dementia and cognitive decline}}.
\newblock {\emph{\JournalTitle{Alzheimer's \& Dementia}}} \textbf{\bibinfo{volume}{11}}, \bibinfo{pages}{639--647} (\bibinfo{year}{2015}).

\bibitem{schulz2004aortic}
\bibinfo{author}{Schulz, E.}, \bibinfo{author}{Arfai, K.}, \bibinfo{author}{Liu, X.}, \bibinfo{author}{Sayre, J.} \& \bibinfo{author}{Gilsanz, V.}
\newblock \bibinfo{journal}{\bibinfo{title}{Aortic calcification and the risk of osteoporosis and fractures}}.
\newblock {\emph{\JournalTitle{The Journal of Clinical Endocrinology \& Metabolism}}} \textbf{\bibinfo{volume}{89}}, \bibinfo{pages}{4246--4253} (\bibinfo{year}{2004}).

\bibitem{libby2021changing}
\bibinfo{author}{Libby, P.}
\newblock \bibinfo{journal}{\bibinfo{title}{The changing landscape of atherosclerosis}}.
\newblock {\emph{\JournalTitle{Nature}}} \textbf{\bibinfo{volume}{592}}, \bibinfo{pages}{524--533} (\bibinfo{year}{2021}).

\bibitem{dhingra2006electrocardiographic}
\bibinfo{author}{Dhingra, R.} \emph{et~al.}
\newblock \bibinfo{journal}{\bibinfo{title}{Electrocardiographic qrs duration and the risk of congestive heart failure: the framingham heart study}}.
\newblock {\emph{\JournalTitle{Hypertension}}} \textbf{\bibinfo{volume}{47}}, \bibinfo{pages}{861--867} (\bibinfo{year}{2006}).

\bibitem{ference2017low}
\bibinfo{author}{Ference, B.~A.} \emph{et~al.}
\newblock \bibinfo{journal}{\bibinfo{title}{Low-density lipoproteins cause atherosclerotic cardiovascular disease. 1. evidence from genetic, epidemiologic, and clinical studies. a consensus statement from the european atherosclerosis society consensus panel}}.
\newblock {\emph{\JournalTitle{European heart journal}}} \textbf{\bibinfo{volume}{38}}, \bibinfo{pages}{2459--2472} (\bibinfo{year}{2017}).

\bibitem{brasen2019cholemic}
\bibinfo{author}{Br{\"a}sen, J.~H.} \emph{et~al.}
\newblock \bibinfo{journal}{\bibinfo{title}{Cholemic nephropathy causes acute kidney injury and is accompanied by loss of aquaporin 2 in collecting ducts}}.
\newblock {\emph{\JournalTitle{Hepatology}}} \textbf{\bibinfo{volume}{69}}, \bibinfo{pages}{2107--2119} (\bibinfo{year}{2019}).

\bibitem{van2013bile}
\bibinfo{author}{Van~Slambrouck, C.~M.}, \bibinfo{author}{Salem, F.}, \bibinfo{author}{Meehan, S.~M.} \& \bibinfo{author}{Chang, A.}
\newblock \bibinfo{journal}{\bibinfo{title}{Bile cast nephropathy is a common pathologic finding for kidney injury associated with severe liver dysfunction}}.
\newblock {\emph{\JournalTitle{Kidney international}}} \textbf{\bibinfo{volume}{84}}, \bibinfo{pages}{192--197} (\bibinfo{year}{2013}).

\bibitem{warny2018lymphopenia}
\bibinfo{author}{Warny, M.}, \bibinfo{author}{Helby, J.}, \bibinfo{author}{Nordestgaard, B.~G.}, \bibinfo{author}{Birgens, H.} \& \bibinfo{author}{Bojesen, S.~E.}
\newblock \bibinfo{journal}{\bibinfo{title}{Lymphopenia and risk of infection and infection-related death in 98,344 individuals from a prospective danish population-based study}}.
\newblock {\emph{\JournalTitle{PLoS medicine}}} \textbf{\bibinfo{volume}{15}}, \bibinfo{pages}{e1002685} (\bibinfo{year}{2018}).

\bibitem{sagar1995intestinal}
\bibinfo{author}{Sagar, P.} \emph{et~al.}
\newblock \bibinfo{journal}{\bibinfo{title}{Intestinal obstruction promotes gut translocation of bacteria}}.
\newblock {\emph{\JournalTitle{Diseases of the colon \& rectum}}} \textbf{\bibinfo{volume}{38}}, \bibinfo{pages}{640--644} (\bibinfo{year}{1995}).

\bibitem{biswas2020understanding}
\bibinfo{author}{Biswas, A.~K.} \& \bibinfo{author}{Acharyya, S.}
\newblock \bibinfo{journal}{\bibinfo{title}{Understanding cachexia in the context of metastatic progression}}.
\newblock {\emph{\JournalTitle{Nature Reviews Cancer}}} \textbf{\bibinfo{volume}{20}}, \bibinfo{pages}{274--284} (\bibinfo{year}{2020}).

\bibitem{williams2004hospitalized}
\bibinfo{author}{Williams, M.~D.} \emph{et~al.}
\newblock \bibinfo{journal}{\bibinfo{title}{Hospitalized cancer patients with severe sepsis: analysis of incidence, mortality, and associated costs of care}}.
\newblock {\emph{\JournalTitle{Critical care}}} \textbf{\bibinfo{volume}{8}}, \bibinfo{pages}{R291} (\bibinfo{year}{2004}).

\bibitem{juhl2026incidentally}
\bibinfo{author}{Juhl, A.~R.} \emph{et~al.}
\newblock \bibinfo{journal}{\bibinfo{title}{Incidentally detected splenomegaly and risk of hematologic cancer and liver disease}}.
\newblock {\emph{\JournalTitle{JAMA oncology}}} \textbf{\bibinfo{volume}{12}}, \bibinfo{pages}{275--284} (\bibinfo{year}{2026}).

\bibitem{van2010abnormal}
\bibinfo{author}{Van~Deursen, V.} \emph{et~al.}
\newblock \bibinfo{journal}{\bibinfo{title}{Abnormal liver function in relation to hemodynamic profile in heart failure patients}}.
\newblock {\emph{\JournalTitle{Journal of cardiac failure}}} \textbf{\bibinfo{volume}{16}}, \bibinfo{pages}{84--90} (\bibinfo{year}{2010}).

\bibitem{goldwasser1997association}
\bibinfo{author}{Goldwasser, P.} \& \bibinfo{author}{Feldman, J.}
\newblock \bibinfo{journal}{\bibinfo{title}{Association of serum albumin and mortality risk}}.
\newblock {\emph{\JournalTitle{Journal of clinical epidemiology}}} \textbf{\bibinfo{volume}{50}}, \bibinfo{pages}{693--703} (\bibinfo{year}{1997}).

\bibitem{don2004poor}
\bibinfo{author}{Don, B.~R.} \& \bibinfo{author}{Kaysen, G.}
\newblock \bibinfo{journal}{\bibinfo{title}{Poor nutritional status and inflammation: serum albumin: relationship to inflammation and nutrition}}.
\newblock {\emph{\JournalTitle{Seminars in Dialysis}}} \textbf{\bibinfo{volume}{17}}, \bibinfo{pages}{432--437} (\bibinfo{year}{2004}).

\bibitem{launay2007prevalence}
\bibinfo{author}{Launay-Vacher, V.} \emph{et~al.}
\newblock \bibinfo{journal}{\bibinfo{title}{Prevalence of renal insufficiency in cancer patients and implications for anticancer drug management: the renal insufficiency and anticancer medications (irma) study}}.
\newblock {\emph{\JournalTitle{Cancer}}} \textbf{\bibinfo{volume}{110}}, \bibinfo{pages}{1376--1384} (\bibinfo{year}{2007}).

\bibitem{sayers2009eutilities}
\bibinfo{author}{Sayers, E.~W.}
\newblock \bibinfo{title}{A general introduction to the e-utilities}.
\newblock In \emph{\bibinfo{booktitle}{Entrez Programming Utilities Help}} (\bibinfo{publisher}{National Center for Biotechnology Information}, \bibinfo{address}{Bethesda, MD}, \bibinfo{year}{2009}).
\newblock \bibinfo{note}{Updated November 17, 2022}.

\bibitem{gxl_paperclip_2026}
\bibinfo{author}{{Generative Expert Labs}}.
\newblock \bibinfo{title}{{Paperclip}}.
\newblock \bibinfo{howpublished}{Scientific literature search software} (\bibinfo{year}{2026}).
\newblock \bibinfo{note}{Version 0.7.130, accessed October 19, 2026}.

\bibitem{blanco2021assessing}
\bibinfo{author}{Blanco-Grau, A.} \emph{et~al.}
\newblock \bibinfo{journal}{\bibinfo{title}{Assessing liver fibrosis using the fib4 index in the community setting}}.
\newblock {\emph{\JournalTitle{Diagnostics}}} \textbf{\bibinfo{volume}{11}}, \bibinfo{pages}{2236} (\bibinfo{year}{2021}).

\bibitem{ndumele2023cardiovascular}
\bibinfo{author}{Ndumele, C.~E.} \emph{et~al.}
\newblock \bibinfo{journal}{\bibinfo{title}{Cardiovascular-kidney-metabolic health: a presidential advisory from the american heart association}}.
\newblock {\emph{\JournalTitle{Circulation}}} \textbf{\bibinfo{volume}{148}}, \bibinfo{pages}{1606--1635} (\bibinfo{year}{2023}).

\bibitem{ghallab2025enteronephrohepatic}
\bibinfo{author}{Ghallab, A.} \emph{et~al.}
\newblock \bibinfo{journal}{\bibinfo{title}{Enteronephrohepatic circulation of bile acids and therapeutic potential of systemic bile acid transporter inhibitors}}.
\newblock {\emph{\JournalTitle{Journal of hepatology}}}  (\bibinfo{year}{2025}).

\bibitem{barabasi2011network}
\bibinfo{author}{Barab{\'a}si, A.-L.}, \bibinfo{author}{Gulbahce, N.} \& \bibinfo{author}{Loscalzo, J.}
\newblock \bibinfo{journal}{\bibinfo{title}{Network medicine: a network-based approach to human disease}}.
\newblock {\emph{\JournalTitle{Nature reviews genetics}}} \textbf{\bibinfo{volume}{12}}, \bibinfo{pages}{56--68} (\bibinfo{year}{2011}).

\bibitem{barnett2012epidemiology}
\bibinfo{author}{Barnett, K.} \emph{et~al.}
\newblock \bibinfo{journal}{\bibinfo{title}{Epidemiology of multimorbidity and implications for health care, research, and medical education: a cross-sectional study}}.
\newblock {\emph{\JournalTitle{The Lancet}}} \textbf{\bibinfo{volume}{380}}, \bibinfo{pages}{37--43} (\bibinfo{year}{2012}).

\bibitem{vanderweele2019principles}
\bibinfo{author}{VanderWeele, T.~J.}
\newblock \bibinfo{journal}{\bibinfo{title}{Principles of confounder selection}}.
\newblock {\emph{\JournalTitle{European journal of epidemiology}}} \textbf{\bibinfo{volume}{34}}, \bibinfo{pages}{211--219} (\bibinfo{year}{2019}).

\bibitem{muennighoff2023mteb}
\bibinfo{author}{Muennighoff, N.}, \bibinfo{author}{Tazi, N.}, \bibinfo{author}{Magne, L.} \& \bibinfo{author}{Reimers, N.}
\newblock \bibinfo{title}{Mteb: Massive text embedding benchmark}.
\newblock In \emph{\bibinfo{booktitle}{Proceedings of the 17th Conference of the European Chapter of the Association for Computational Linguistics}}, \bibinfo{pages}{2014--2037} (\bibinfo{year}{2023}).

\bibitem{bussmann2025learning}
\bibinfo{author}{Bussmann, B.}, \bibinfo{author}{Nabeshima, N.}, \bibinfo{author}{Karvonen, A.} \& \bibinfo{author}{Nanda, N.}
\newblock \bibinfo{journal}{\bibinfo{title}{Learning multi-level features with matryoshka sparse autoencoders}}.
\newblock {\emph{\JournalTitle{arXiv preprint arXiv:2503.17547}}}  (\bibinfo{year}{2025}).

\bibitem{thapa2026multimodal}
\bibinfo{author}{Thapa, R.} \emph{et~al.}
\newblock \bibinfo{journal}{\bibinfo{title}{A multimodal sleep foundation model for disease prediction}}.
\newblock {\emph{\JournalTitle{Nature Medicine}}} \textbf{\bibinfo{volume}{32}}, \bibinfo{pages}{752--762} (\bibinfo{year}{2026}).

\bibitem{chernozhukov2018double}
\bibinfo{author}{Chernozhukov, V.} \emph{et~al.}
\newblock \bibinfo{journal}{\bibinfo{title}{Double/debiased machine learning for treatment and structural parameters}}.
\newblock {\emph{\JournalTitle{The Econometrics Journal}}} \textbf{\bibinfo{volume}{21}}, \bibinfo{pages}{C1--C68} (\bibinfo{year}{2018}).

\bibitem{liu2020double}
\bibinfo{author}{Liu, M.}, \bibinfo{author}{Zhang, Y.} \& \bibinfo{author}{Zhou, D.}
\newblock \bibinfo{journal}{\bibinfo{title}{Double/debiased machine learning for logistic partially linear model}}.
\newblock {\emph{\JournalTitle{arXiv preprint arXiv:2009.14461}}}  (\bibinfo{year}{2020}).

\bibitem{tan2019doubly}
\bibinfo{author}{Tan, Z.}
\newblock \bibinfo{journal}{\bibinfo{title}{On doubly robust estimation for logistic partially linear models}}.
\newblock {\emph{\JournalTitle{Statistics \& Probability Letters}}} \textbf{\bibinfo{volume}{155}}, \bibinfo{pages}{108577} (\bibinfo{year}{2019}).

\bibitem{bouzid2025insights}
\bibinfo{author}{Bouzid, K.} \emph{et~al.}
\newblock \bibinfo{journal}{\bibinfo{title}{Insights into a radiology-specialised multimodal large language model with sparse autoencoders}}.
\newblock {\emph{\JournalTitle{arXiv preprint arXiv:2507.12950}}}  (\bibinfo{year}{2025}).

\bibitem{bello2022dagma}
\bibinfo{author}{Bello, K.}, \bibinfo{author}{Aragam, B.} \& \bibinfo{author}{Ravikumar, P.}
\newblock \bibinfo{journal}{\bibinfo{title}{Dagma: Learning dags via m-matrices and a log-determinant acyclicity characterization}}.
\newblock {\emph{\JournalTitle{Advances in Neural Information Processing Systems}}} \textbf{\bibinfo{volume}{35}}, \bibinfo{pages}{8226--8239} (\bibinfo{year}{2022}).

\bibitem{zheng2018dags}
\bibinfo{author}{Zheng, X.}, \bibinfo{author}{Aragam, B.}, \bibinfo{author}{Ravikumar, P.~K.} \& \bibinfo{author}{Xing, E.~P.}
\newblock \bibinfo{journal}{\bibinfo{title}{Dags with no tears: Continuous optimization for structure learning}}.
\newblock {\emph{\JournalTitle{Advances in Neural Information Processing Systems}}} \textbf{\bibinfo{volume}{31}} (\bibinfo{year}{2018}).

\end{thebibliography}

\newpage

\section{Methods}
\label{methods:}

\subsection{Datasets}
\label{methods:datasets}

All datasets are partitioned into a Discovery Site and a series of Testing sets.
The Discovery Site is used to train sparse autoencoders, select disease outcomes, interpret concepts, and rank associations.
Testing sets are held out entirely and used solely to assess whether insights replicate in patient cohorts that played no part in discovery, in order of increasing institutional distance from Stanford.
All datasets were collected under IRB approval with a waiver of informed consent owing to retrospective use.

\textbf{Discovery Site -- Stanford Main hospital.}
\label{methods:datasets:discovery}
The Discovery Site cohort comprises consecutive abdominal CT examinations performed at Stanford academic medical center from December 2012 to October 2018, yielding 20,391 CT studies from 14,652 unique patients (3,262,560 CT slices, 1,588,003 lab entries, and 22,781,540 historical diagnoses).
No patient contributed studies to more than one split.
We divide the cohort into a train set of 15,331 studies from 11,009 patients (2,452,960 slices, 6,083,609 report tokens, 1,194,451 lab entries, 17,017,170 diagnoses) and a validation set of 5,060 studies from 3,643 patients (809,600 slices, 2,000,273 report tokens, 393,552 lab entries, 5,764,373 diagnoses).
Embedding normalization statistics are computed on the Discovery Site.
Finally, all datasets contain abdominal CT images and radiology reports that capture the current health of the patient; laboratory analyses are restricted to the Discovery Site, Internal Test and External Site \#1. All have longitudinal ICD codes with at least 4 years of follow-up for association studies, and the six universal clinical covariates, including race.

CT preprocessing follows \cite{blankemeier2026merlin}.
From each study, the series with the greatest number of axial slices was selected to maximize information content, and a contrast-phase detection algorithm confirmed that 97\% of selected series were portal venous phase.
Volumes were preprocessed by orienting axes (left--right, posterior--anterior, inferior--superior), resampling to 1.5\,mm in-plane resolution and 3\,mm slice thickness, clipping and normalizing Hounsfield units from $[-1000, 1000]$ to $[0, 1]$, and center-cropping to $224 \times 224 \times 160$ voxels.

Report preprocessing follows \cite{blankemeier2026merlin}.
Only the findings section of each report was used, as it provides the most granular per-organ-system observations. It covers the lower thorax, liver and biliary tree, gallbladder, spleen, pancreas, adrenal glands, kidneys and ureters, gastrointestinal tract, peritoneal cavity, pelvic organs, vasculature and lymph nodes, and musculoskeletal system.

Longitudinal lab measurements were collected from the EHR for each patient.
Only numerical lab types were retained, defined as those where at least 90\% of recorded measurements were numeric, resulting in 350 lab types (Supplementary Table~\ref{supplementary:lab_types}).
To reduce noise in dense measurement periods, a quarterly downsampling strategy was applied, retaining only the last recorded value per patient, lab type, and calendar quarter, using only measurements taken within the year prior to the CT scan date.
Each lab value was z-score normalized per lab type and clamped to $[-3, 3]$ to limit the influence of extreme outliers.

For each patient encounter associated with a CT scan, all ICD-9 and ICD-10 codes assigned during that encounter were collected and mapped to hierarchical phenotypes using PheWAS Phecode mapping \cite{denny2013systematic}, collapsing 16,553 distinct ICD codes to 1,692 phenotype categories.

\textbf{Internal Test cohort -- Stanford Main hospital.}
\label{methods:datasets:internal_test}
The Internal Test cohort contains 5,137 CT studies from 3,667 unique patients drawn from the same institution as the Discovery Site (821,920 CT slices, 405,802 lab entries, 6,134,090 diagnoses).
These patients were held out from all stages of discovery and represent the first test of reproducibility under matched distributional conditions.

\textbf{External Site \#1.}
\label{methods:datasets:tri_valley}
The External Site \#1 cohort contains 25,803 CT studies from 17,224 patients at a community hospital outside Stanford (4,128,480 CT slices, 1,764,079 lab entries, 19,610,547 diagnoses).
No data from this cohort was used at any stage of discovery or model training, and the shift in patient population and clinical practice makes it a more demanding test of generalization.

\textbf{External Site \#2 -- Mayo Clinic.}
\label{methods:datasets:mayo}
The Mayo Clinic cohort consists of patients across four Mayo Clinic hospitals, comprising 14,199 CT studies from 6,085 patients.
No data from this cohort was used at any stage of discovery or model training, and as an out-of-state health system it represents the greatest institutional distance from the Discovery Site.

\begin{table}[h]
\centering
\small
\resizebox{\textwidth}{!}{%
\begin{tabular}{l cc cc cc cc}
    \toprule
    & \multicolumn{2}{c}{Discovery (Train + Val)} & \multicolumn{2}{c}{Internal Test} & \multicolumn{2}{c}{External Site \#1} & \multicolumn{2}{c}{External Site \#2} \\
    \cmidrule(lr){2-3} \cmidrule(lr){4-5} \cmidrule(lr){6-7} \cmidrule(lr){8-9}
    Dataset Characteristics & n & \% & n & \% & n & \% & n & \% \\
    \midrule
    Accessions (Abdominal CTs) & 20,391 & -- & 5,137 & -- & 25,803 & -- & 14,199 & -- \\
    Unique Patients & 14,652 & -- & 3,667 & -- & 17,224 & -- & 6,085 & -- \\
    Age (mean) & 54.1 $\pm$ 18.9 & -- & 54.5 $\pm$ 19.1 & -- & 60.6 $\pm$ 18.9 & -- & 62.1 $\pm$ 16.9 & -- \\
    \multicolumn{9}{l}{Sex} \\
    \quad Female & 11,407 & 55.9 & 2,927 & 57.0 & 14,039 & 54.4 & 6,994 & 49.3 \\
    \quad Male & 8,982 & 44.0 & 2,210 & 43.0 & 11,764 & 45.6 & 7,205 & 50.7 \\
    \quad Other & 2 & 0.01 & 0 & 0.00 & 0 & 0.00 & 0 & 0.00 \\
    \multicolumn{9}{l}{Smoking Status} \\
    \quad Current Smoker & 1,460 & 7.2 & 353 & 6.9 & 1,380 & 5.3 & 1,633 & 11.5 \\
    \quad Heavy Alcohol Use & 2,419 & 11.9 & 667 & 13.0 & 2,875 & 11.1 & 2,327 & 16.4 \\
    BMI (mean) & 26.9 $\pm$ 6.9 & -- & 26.9 $\pm$ 6.9 & -- & 27.4 $\pm$ 6.4 & -- & 27.1 $\pm$ 2.7 & -- \\
    \multicolumn{9}{l}{Laboratory values (6mo prior)} \\
    \quad Lab values & 1,934,671 & -- & 494,575 & -- & 1,772,983 & -- & 145,868 & -- \\
    \quad Accessions with lab values & 19,845 & 97.3 & 5,011 & 97.5 & 24,710 & 95.8 & 4,155 & 29.3 \\
    \bottomrule
\end{tabular}

}
\caption{Summary of dataset characteristics across the Discovery, Internal Test, External Site \#1, and External Site \#2 (Mayo Clinic) cohorts, including the number of unique patients and abdominal CTs, age, sex distribution, smoking status, heavy alcohol use, and BMI. Values are reported as counts (n) and percentages where applicable.}
\label{methods:table:dataset_characteristics}
\end{table}

\subsection{Foundation Models}
\label{methods:fms}

Each data modality is encoded by the most performant foundation model selected at the time of the study.
FM weights are frozen throughout all downstream analyses; the SAEs and association pipelines operate entirely on static embeddings produced by these models.

\paragraph{CT imaging} For CT images, we use Merlin \cite{blankemeier2026merlin}, a 3D inflated ResNet-152 pretrained on abdominal CT volumes paired with radiology reports and EHR diagnosis codes.
Merlin processes each CT volume and produces a 2,048-dimensional embedding that serves as the input to all downstream concept extraction and association analyses.

\paragraph{CT radiology report} For radiology report findings, we use Qwen3-Embedding-8B \cite{zhang2025qwen3}, a general-domain text embedding model selected as the top-performing model on the MTEB leaderboard at the time of this study \cite{muennighoff2023mteb}.
Each findings section is encoded into a single 4,096-dimensional vector.

\paragraph{Laboratory values}
We use EHR Labs FM, an in-house masked autoencoder pretrained on laboratory histories from the Discovery Site. The encoder is a three-layer Feature Tokenizer Transformer with four attention heads and a hidden dimension of 768 that jointly encodes laboratory values, test types and measurement times. For each history, values are randomly masked at a rate sampled uniformly between 0 and 90\%, with at least one value masked. Laboratory-specific prediction heads reconstruct the masked values using a mean squared error objective. Training uses AdamW with a learning rate of $10^{-4}$, a batch size of 256 and 1,500 optimisation steps, with checkpoint selection based on validation reconstruction loss. For downstream analyses, the reconstruction heads are discarded and the encoder remains frozen. Its output embeddings are averaged over non-padding tokens to produce a single 768-dimensional vector.

\subsection{Sparse Autoencoders}
\label{methods:saes}

FM embeddings are dense and polysemantic: individual dimensions typically encode mixtures of unrelated clinical concepts simultaneously, making direct interpretation infeasible.
Sparse autoencoders address this by learning a larger, overcomplete dictionary of directions in the embedding space, where each sample activates only a small subset of dictionary entries.
This sparsity encourages each entry to correspond to a single, distinct clinical concept.

\textbf{Architecture.}
For each modality, a separate SAE operates on the z-normalized FM embedding $\tilde{x} = (x - \mu)/\sigma$, where statistics are computed on the Discovery train set.
Z-normalization is critical so that the SAE learns relative structure in the embedding space rather than the distance of all points from the origin.
A linear encoder maps $\tilde{x}$ to pre-activation scores over a dictionary of $n = 8{,}192$ features,
\begin{equation}
\hfill z = \tilde{x} W_{\text{enc}} + b_{\text{enc}}, \quad W_{\text{enc}} \in \mathbb{R}^{d \times n} \hfill
\end{equation}
and Top-$K$ sparsity retains only the $K$ highest-scoring features, zeroing the rest to produce the sparse representation $z_{\text{sparse}} = \text{TopK}(z, K)$.
A linear decoder then reconstructs the embedding,
\begin{equation}
\hfill \hat{x} = z_{\text{sparse}} W_{\text{dec}} + b_{\text{dec}}, \quad W_{\text{dec}} \in \mathbb{R}^{n \times d}. \hfill
\end{equation}
The decoder is initialized as the transpose of the encoder, and its columns are constrained to unit norm throughout training, with the gradient component parallel to each column removed after each step to prevent shrinkage.
We ablate five sparsity levels, $K \in \{5, 10, 20, 40, 80\}$.

\textbf{Matryoshka SAEs.}
Standard Top-$K$ SAEs tend to produce only fine-grained features, a phenomenon known as feature splitting, neglecting broader concepts higher in the clinical hierarchy.
To address this, we additionally train Matryoshka SAEs \cite{bussmann2025learning}, which impose nested reconstruction losses across an ordered set of prefix sizes $\mathcal{M}$,
\begin{equation}
\hfill \mathcal{L}_{\text{Matryoshka}} = \frac{1}{|\mathcal{M}|} \sum_{m \in \mathcal{M}} \left\| \tilde{x} - z_{\text{sparse}}[{:}m]\, W_{\text{dec}}[{:}m, :] - b_{\text{dec}} \right\|_2^2. \hfill
\end{equation}
Smaller prefixes are forced to encode coarser concepts while larger prefixes introduce progressively finer distinctions, producing a hierarchical concept space.
We test a standard Top-$K$ SAE and two Matryoshka SAE configurations, with nested dictionary sizes $\mathcal{M}_1 = \{128, 512, 2048, 8192\}$ and $\mathcal{M}_2 = \{256, 1024, 4096, 16384\}$.

\textbf{Training.}
SAEs are trained on the Discovery train set by minimizing MSE reconstruction loss, $\mathcal{L} = \|\tilde{x} - \hat{x}\|_2^2$, using the Adam optimizer with a learning rate of $5 \times 10^{-4}$, a batch size of 2,000, and a maximum of 1,000 steps with early stopping based on validation loss.
One SAE is trained per modality.

\textbf{Feature extraction and configuration selection.}
Once trained, SAE weights are frozen and sparse features are inferred on all datasets by passing each FM embedding through the pretrained SAE.
Each sparse feature is then z-normalized by subtracting its mean and dividing by its standard deviation, both computed on the Discovery Site, providing a consistent basis for comparison across modalities and datasets.
Based on the ablation results described in Supplementary Section~\ref{section:results:architectural_ablation} and Figure~\ref{figure:6}, we select \(K = 40\) with the \(\mathcal{M} = \{128, 512, 2048, 8192\}\) Matryoshka architecture as a single consistent configuration across all modalities.
The resulting sparse feature vectors from each modality are concatenated to form a unified multimodal concept space, in which each sample is represented by a single sparse vector spanning imaging, report, and laboratory concepts.

\subsection{Automated Interpretability}
\label{methods:autointerp}

By design, sparse features activate for only a small subset of samples.
This sparsity creates the precondition for interpretability: if a feature fires for a restricted, coherent cohort, it may encode a single identifiable clinical phenomenon.
To convert features into named concepts at scale, we automate the interpretation process using a large language model (LLM), eliminating the need for manual inspection of thousands of features.

\textbf{Generative interpretation.}
For each SAE feature, we identify the samples in the Discovery Site whose activation exceeds the 75th percentile of all positive activations, forming a high-activating cohort.
We then present $n_\text{high} = 20$ randomly sampled high-activating samples to an LLM, alongside $n_\text{non} = 20$ randomly sampled samples with zero or near-zero activation as a contrast reference.
The contrast is essential: it grounds the interpretation in what is \emph{specific} to the concept rather than what is common across the patient population.
The LLM is instructed to reason via explicit chain-of-thought before producing a single-sentence summary of the form ``\textit{This cohort is characterized by [interpretation]}.''
We use Gemini 2.5 Pro via a PHI-safe GCP Vertex instance.
Full prompts are provided in Supplementary Section~\ref{supplementary:autointerp_prompts}.

\textbf{Discriminative validation.}
Validation serves two purposes: assessing whether the interpretation faithfully describes the feature, and testing whether this description generalises to held-out sets.
For each concept, we sample $n_\text{high} = 20$ and $n_\text{non} = 20$ from the test set, then crucially we shuffle the combined pool.
The LLM receives the textual interpretation together with the shuffled sample data, is informed that exactly half the samples belong to the cohort, and must classify each sample as either belonging to the concept cohort (1) or not (0).
Discrimination accuracy, calculated by the fraction of samples correctly classified, is computed on the Internal Test set as a measure of monosemanticity, and separately on each external cohort to assess cross-site and -institutional generalization.

\subsection{Outcome selection}
\label{methods:outcome_selection}

\begin{figure}[!ht]
    \centering
    \includegraphics[width=0.9\textwidth]{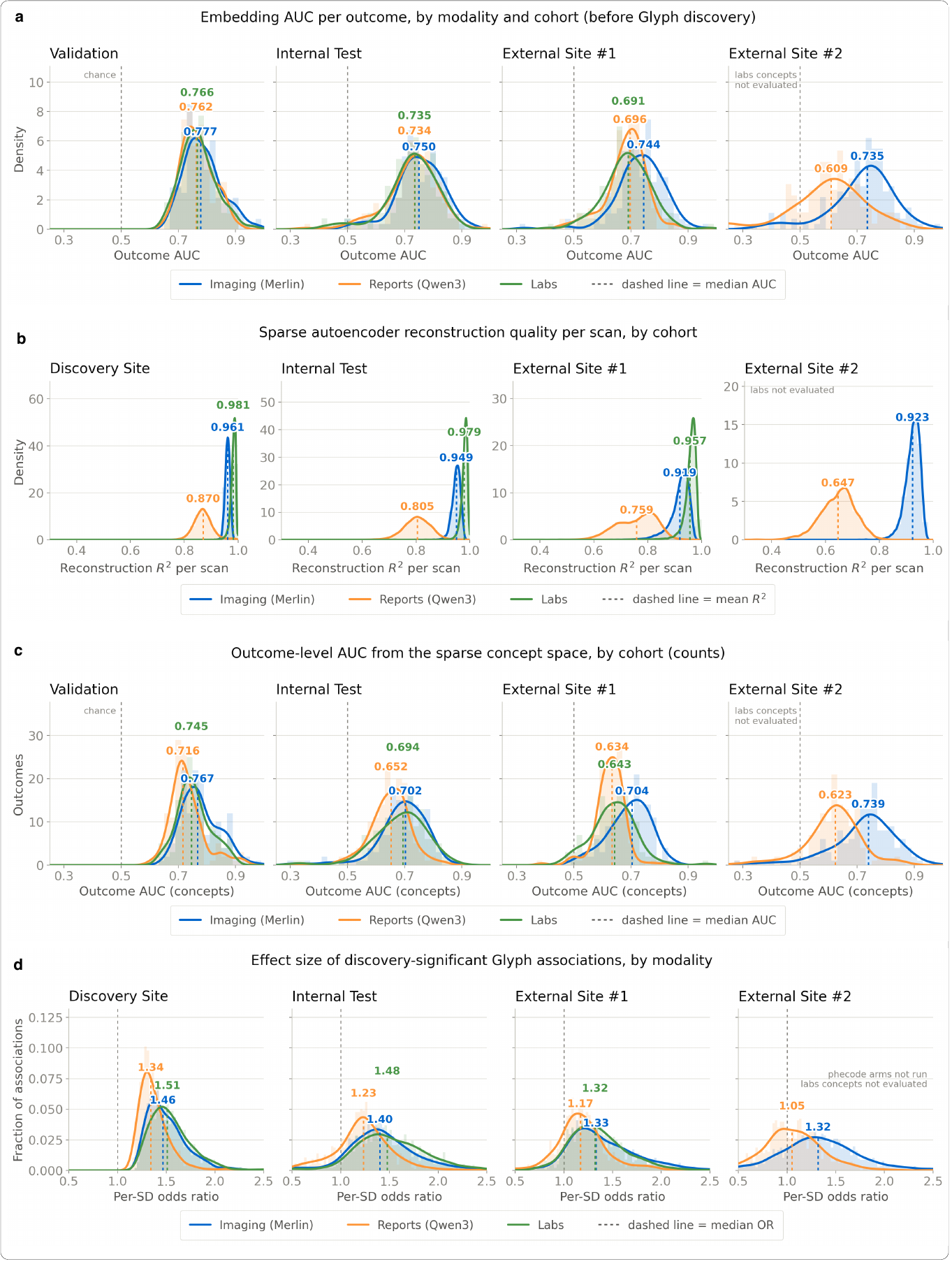}
    \caption{
    \textbf{a,} Outcomes were selected by ranking Phecode prediction AUCs on the Discovery Site validation set.
    After excluding non-disease codes, 141 of the top 150 outcomes were retained.
    Without refitting, imaging probes transferred with minimal loss across cohorts, whereas report performance declined at External Site \#2.
    \textbf{b,} SAEs fitted at the Discovery Site were applied unchanged across cohorts.
    Reconstruction remained high for imaging and labs but declined steadily for reports.
    \textbf{c,} Replacing dense embeddings with sparse concepts preserved most predictive information, with less loss for imaging than reports.
    \textbf{d,} Median odds ratios of significant associations followed a similar per-site trend to the predictive performance, maintaining high OR for imaging- and labs-derived concepts and steadily dropping for radiology report-derived concepts.
    }
    \label{figure:extended:pipeline}
\end{figure}

To maximize the potential for rediscoveries and new discoveries, we use an automated approach to select the highest signal outcome targets for exploration using the Discovery Site.
We first map 16,553 unique ICD-9 and ICD-10 codes to 1,692 hierarchical phenotypes termed Phecodes \cite{denny2013systematic}.
From these we define a candidate pool of 1,278 outcomes with support in our data.
Studies were labeled as progressors if the first recorded diagnosis of an outcome occurred 1--4 years after the study date. 
Controls were selected as having no recorded diagnosis of that outcome, and at least four years of follow-up.
All other studies, including those with a recorded diagnosis of the outcome before the CT date, were excluded from that outcome’s analysis.
Following \cite{thapa2026multimodal}, we then compute on the Discovery Site the aggregate rank (by reciprocal rank fusion) of the ability of each modality and model combination to predict each outcome.
The 150 highest-ranked outcomes were then manually reviewed, and nine excluded as not interpretable as clinical disease endpoints.
These comprised non-specific catch-all Phecodes (for example, \textit{other disorders of the liver}), abnormal-finding and administrative codes (for example, \textit{other abnormal blood chemistry}), and device or procedure status codes (for example, \textit{ileostomy status}).
This left the 141 outcomes used in this study, covering conditions across oncological, cardiovascular, metabolic, and acute disease categories.
To see the performance of each FM on this set, see Figure \ref{figure:extended:pipeline}.

\subsection{Association study}
\label{section:methods:or}

Taking inspiration from the multi-stage design of genome-wide association studies~\cite{denny2013systematic}, we estimate the association between each Glyph concept and each future disease outcome as a per-standard-deviation odds ratio (OR), enabling direct comparison across modalities whose activations operate on different scales.

\textbf{Standardization.}
Concept activations are z-normalised using a per-site mean and standard deviation, so that each OR reflects the multiplicative change in outcome odds per one standard deviation increase in concept activation.

\textbf{OR computation.}
For each concept-outcome pair, we fit an unadjusted logistic regression model on the Discovery Site,
\begin{equation}
\hfill
\text{logit}(P(Y=1)) = \alpha + \beta \tilde{c},
\hfill
\end{equation}
where $\tilde{c}$ is the standardised concept activation.
The per-standard-deviation OR is defined as $\exp(\beta)$, and 95\% confidence intervals are reported as $\exp(\beta \pm 1.96 \cdot \mathrm{SE})$.

\textbf{Statistical significance.}
We assess significance using the Wald $p$-value derived from the asymptotic covariance matrix, computed as the Moore-Penrose pseudo-inverse of the Fisher information matrix,
\begin{equation}
\hfill
I(\beta) = X^\top W X,
\hfill
\end{equation}
where $W$ is a diagonal matrix with entries $p_i(1 - p_i)$.
The $z$-statistic $z = \beta / \mathrm{SE}$ yields a two-tailed $p$-value. Each CT study was analyzed as a separate observation to capture the patient’s clinical state at the time of imaging, with independence assumed across studies.

\textbf{Discovery filter.}
To concentrate downstream analysis on statistically supported, directionally consistent associations, we apply a Bonferroni correction across all concept-outcome pairs, setting the threshold at $p_{\text{Bonferroni}} = 0.05 / N_{\text{pairs}} = 8.28 \times 10^{-8}$.
We additionally require OR $> 1.0$ to restrict attention to predictive rather than protective associations.
Associations satisfying both criteria on the Discovery Site form the set of promising associations carried forward into all subsequent analyses.

\subsection{Reproducibility filter}
\label{section:methods:reproducibility}

Having identified significant associations on the Discovery Site, we assess whether they represent real, generalizable phenomena by re-evaluating them on held-out cohorts.
Concept activations on held-out data are computed by applying the same frozen SAE weights without modification, requiring no manual labeling or recalibration of new samples.

For each promising association, we re-compute the same OR on each held-out cohort and apply the Benjamini-Hochberg (BH) procedure at a 5\% false discovery rate across the set of re-evaluated associations.
An association is considered reproducible if the BH-adjusted $p$-value is significant and directionality is maintained (OR $> 1.0$).

\begin{table}[h]
\centering
\small
\begin{tabular}{l l r}
    \toprule
    Cohort & Correction & Threshold \\
    \midrule
    Discovery (Train) & Bonferroni & $8.28 \times 10^{-8}$ \\
    Internal Test & Benjamini--Hochberg (5\% FDR) & $3.46 \times 10^{-2}$ \\
    External Site \#1 & Benjamini--Hochberg (5\% FDR) & $3.72 \times 10^{-2}$ \\
    External Site \#2 & Benjamini--Hochberg (5\% FDR) & $2.44 \times 10^{-2}$ \\
    \bottomrule
\end{tabular}

\caption{Significance thresholds for the association filters. The Discovery Site uses a Bonferroni-corrected threshold across all tested concept-outcome pairs; each held-out cohort uses its own Benjamini--Hochberg threshold (5\% FDR) computed over the promising associations re-evaluated on that cohort.}
\label{methods:table:association_thresholds}
\end{table}

\subsection{Confounder adjustment}
\label{section:methods:dml}

A predictive association between a concept and an outcome may be driven by a shared confounder rather than by a direct relationship.
To assess whether associations persist after adjustment for measured confounders, we apply double machine learning (DML) with cross-fitting~\cite{chernozhukov2018double}.
This allows us to account for nonlinear relationships and interactions among measured confounders that logistic regression with linear covariate terms may miss.
The estimator uses a \textit{Neyman-orthogonal} estimating equation, whose expectation is locally insensitive to first-order errors in the nuisance functions at their true values.
Orthogonality therefore reduces sensitivity to nuisance estimation errors, while cross-fitting limits overfitting bias.

Because our outcomes are binary, we use a logistic partially linear model,
\begin{equation}
\text{logit}\,P(Y = 1 \mid T, W) = \theta T + r(W),
\end{equation}
for concept activation $T$, binary outcome $Y$, and confounder set $W$, where $\theta$ is the adjusted log-odds ratio of interest and $r(W)$ is the baseline log-odds function.
We estimate the nuisance functions using XGBoost with five-fold outer and five-fold inner cross-fitting, so that each observation is evaluated using nuisance models fitted without that observation.
All nuisance construction, including inner cross-fitting, is confined to the corresponding outer training fold.
Models use 100 trees with maximum depth four. Each CT study is treated as an independent observation, and activations and continuous covariates are standardized using Discovery training statistics.

\textbf{Baseline log-odds ($r$).}
We estimate $r(W)$ using the Full Model Refitting procedure of Liu et al.~\cite{liu2020double}.
Within each outer training fold, a joint model of $Y$ on $(T,W)$ produces inner out-of-fold predictions of the full conditional probability.
Their logits serve as an intermediate target for estimating a preliminary exposure coefficient and the baseline function $r(W)$, accounting for the concept's contribution to the full predictor.

\textbf{Residualized activation ($\tilde{T}$).}
For the sigmoid-based score used here, orthogonality requires a variance-weighted conditional mean of the concept activation~\cite{tan2019doubly},
\begin{equation}
m(W) =
\frac{\mathbb{E}[v(T,W)T \mid W]}
     {\mathbb{E}[v(T,W) \mid W]},
\qquad
v(T,W) = \mu(T,W)\{1-\mu(T,W)\},
\end{equation}
where $\mu(T,W)=P(Y=1\mid T,W)$.
We estimate $m(W)$ by regressing $T$ on $W$, with variance weights calculated from the inner out-of-fold predictions of the full outcome model.
The fitted weighted regressor is applied to the outer held-out fold to obtain residuals $\tilde{T}_i=T_i-\hat{m}(W_i)$.

The adjusted effect $\hat{\theta}$ is then obtained by solving the orthogonal score equation
\begin{equation}
\sum_i \left(Y_i-\sigma(\theta T_i+\hat{r}(W_i))\right)\tilde{T}_i=0,
\end{equation}
by root-finding, where $\sigma$ is the logistic sigmoid function.
Standard errors are calculated using the corresponding sandwich estimator.
We report the adjusted OR as $\exp(\hat{\theta})$, with a 95\% Wald confidence interval and a two-sided Wald $p$-value testing $\theta=0$.

\textbf{Confounder selection.}
For each association, confounders are drawn from two sources.
The six universal clinical covariates are patient age, sex, BMI, alcohol use, smoking status and race, with race represented by indicator variables.
In addition, up to ten conceptual confounders are automatically selected per outcome from the significant Glyph associations ranked by unadjusted OR. 
These are selected independently within each dataset to account for differences between institutions and populations.
This also provides the opportunity to test confounders from other modalities, though we disallow confounders from the radiology report to interfere with those from the image, and vice versa, as they were derived from the same data source.
To limit overadjustment for overlapping representations of the tested concept and potential mediators, we applied a correlation threshold.
We selected an absolute Pearson correlation threshold of 0.35 through qualitative inspection to balance excluding likely downstream consequences against retaining potential confounders, and excluded candidates at or above this threshold.
For example, when testing ascites, a broader concept combining ascites, anasarca, and pleural effusion was excluded from the adjustment set (r=0.61).
Similarly, when testing aortic atherosclerosis, a broader concept representing abdominopelvic atherosclerosis was excluded (r=0.41).
However, related atherosclerosis and aortic calcification concepts remained eligible for adjustment when analyzing aortic atherosclerosis (r=0.31–0.34), whereas upstream liver pathology was excluded when analyzing splenomegaly (r=0.46).
The threshold therefore served as a heuristic, since correlation alone cannot distinguish overlapping representations, mediators, and confounders \cite{vanderweele2019principles}.

\textbf{Confounder filter.}
Similar to the Reproducibility filter, an association is considered to survive confounder adjustment if the BH-adjusted $p$-value (at 5\% FDR, applied to the full set of Bonferroni-significant associations) remains significant and the adjusted OR $> 1.0$.
Adjustment was evaluated for the 4,463 significant Discovery associations in each cohort, except at External Site \#2, where only the 3,109 imaging and radiology report associations were evaluated.

\begin{table}[h]
\centering
\small
\begin{tabular}{l l r}
    \toprule
    Cohort & Correction & Threshold \\
    \midrule
    Discovery (Train) & Benjamini--Hochberg (5\% FDR) & $2.24 \times 10^{-2}$ \\
    Internal Test & Benjamini--Hochberg (5\% FDR) & $1.06 \times 10^{-2}$ \\
    External Site \#1 & Benjamini--Hochberg (5\% FDR) & $9.50 \times 10^{-3}$ \\
    External Site \#2 & Benjamini--Hochberg (5\% FDR) & $1.46 \times 10^{-2}$ \\
    \bottomrule
\end{tabular}

\caption{Significance thresholds for the confounder filter. Unlike the association filters (Table~\ref{methods:table:association_thresholds}), every cohort, including Discovery, uses a Benjamini--Hochberg threshold (5\% FDR) computed over the confounder-adjusted $p$-values of the promising associations tested on that cohort.}
\label{methods:table:confounder_thresholds}
\end{table}

\subsection{Comparison with the clinical literature}
\label{section:methods:literature}
To assess whether Glyph identifies prognostic information beyond established clinical markers, we compared its concepts with risk markers reported in the literature. Here, we describe the literature search
and screening, reconstruction of published markers, and methods used to compare their associations and predictive value with Glyph. For each outcome, PubMed and Paperclip queries combined endpoint names and
synonyms with an input scope defining eligible abdominal imaging findings, laboratory measurements and clinical risk factors. Publication in a leading relevant journal or at least 50 citations was required,
retaining 83,649 outcome-specific records. Title screening advanced 14,801 records to abstract review, of which 4,748 underwent further assessment of their measurements, endpoints and local feasibility.
Original methods and tables were consulted where marker definitions were unclear. At least one reported marker could be reconstructed for 1,822 records, before deduplicating repeated measurement families and
selecting 293 representative studies. 
Markers were reconstructed using published thresholds, transformations and composite-score definitions, with approximate measurements labelled as proxies. Counting markers separately for each outcome, the final 411 entries comprised 36 explicit multi-input scores, ratios or biomarker rules
(8.8\%), 31 BMI or estimated kidney-function measures (7.5\%), and 344 single measurements, findings or recorded conditions (83.7\%). Thus, 67 markers (16.3\%) combined multiple inputs under the broader
definition including BMI and estimated kidney function. Queries, source papers, screening decisions and fitted definitions were retained for each outcome.

All prediction models used the same outcome-eligible CT studies and included age, sex and recorded race. Markers required at least 100 observed Discovery measurements and, for binary rules, at least 25
observations in each category. Missing measurements were replaced with the Discovery training mean and accompanied by a missingness indicator. Ridge logistic regression coefficients and preprocessing were
learned on Discovery training data, with regularization selected by validation AUROC and the selected model applied without refitting. Models were fitted to convergence before validation AUROC was used to select regularization. Patients did not cross fitting, tuning or evaluation splits.
We repeated fitting on five bootstrap samples of independent Discovery training studies, keeping the concept sets, validation set and held-out cohorts fixed. Reported AUROCs are the arithmetic means of the five fitted-model AUROCs, rather than AUROCs of averaged predictions. If an augmented model did not improve validation AUROC, the corresponding baseline was retained.
Evaluation used the pooled Internal Test and External Site \#1 cohorts. External Site \#2 was not included because direct access to the clinical data needed to reconstruct the literature markers was unavailable at the time of the study. AUROC confidence intervals and two-sided comparisons used paired study-level influence functions averaged across the five fits. Across-outcome means gave outcomes equal weight, retaining shared-study covariance across models and outcomes; relative-gain uncertainty used the delta method. Inference conditions on the fitted models and selected markers and does not include training or selection uncertainty.

Individual-marker associations were estimated using the orthogonal logistic DML estimator described above, adjusted for age, sex and recorded race, with five-fold outer and inner cross-fitting and independent-study sandwich uncertainty. Exposures, including binary rules, were standardized using Discovery training statistics.
Ten Glyph concepts per outcome were curated using Discovery clinically adjusted rankings and clinical interpretations, excluding concepts primarily reflecting intervention or workflow. For the size-matched OR and reproduction comparisons, up to ten literature markers per outcome were selected using the adjusted Discovery-site OR ranking and fixed across held-out cohorts. This selection did not restrict the literature baseline used in the prediction models. There were 194 selected literature markers across outcomes.
Reproduction required BH-adjusted $q<0.05$ in the pooled held-out cohort and the same effect direction as Discovery, with BH applied across the 768 Glyph, literature and matched-Phecode marker--outcome associations. Differences in reproduction rates were tested using two-sided Fisher exact tests, treating reproduction of each association as independent.
Median OR comparisons gave outcomes equal weight and used 20,000 shared-study multiplier bootstrap samples to preserve dependence across markers and outcomes, with marker sets and nuisance fits fixed. The minimum attainable bootstrap $p$-value was $1/20{,}001$. Benjamini--Hochberg correction was applied to the six displayed held-out OR comparisons, with separate families for the three Discovery and three pooled comparisons. KDEs used all supported outcomes within each marker family; pairwise median comparisons used the outcomes supported by both families.

\subsection{Targeted exploration of concepts and associations}
\label{supplementary:I87}
\begin{figure}[!ht]
    \centering
    \includegraphics[width=0.88\textwidth]{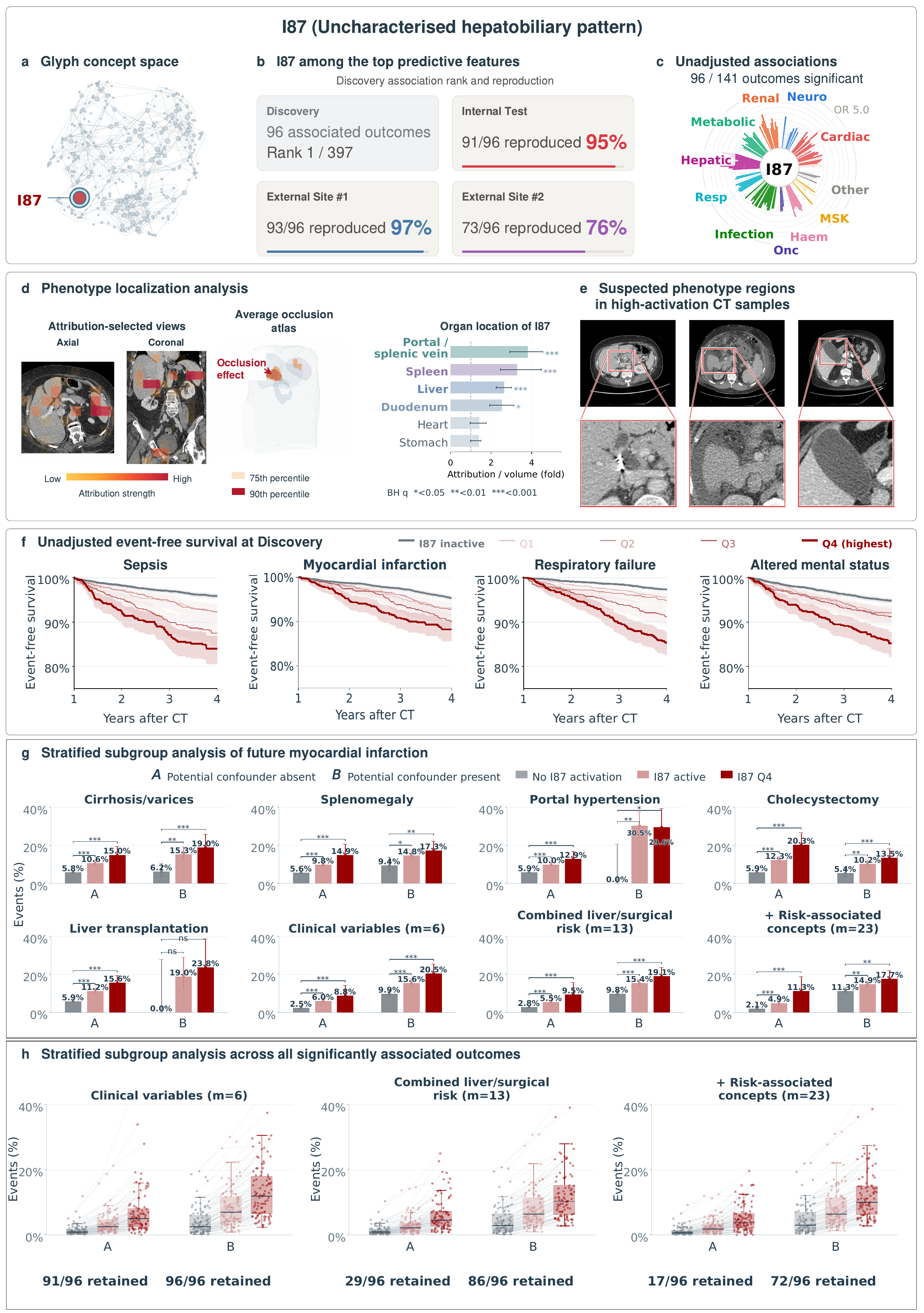}
\caption{\textbf{Investigating I87.}
  \textbf{a} I87 is situated within the Glyph concept space in the hepatic and splenic disease neighborhood, and 
  \textbf{b} demonstrated the largest number of significant associations in every cohort tested.
  \textbf{c} Associations spanned all 11 outcome groups, revealing a broad systemic risk profile.
  \textbf{d} Occlusion analysis in 15 patients localized I87 attribution to the portal/splenic veins, spleen, liver and duodenum. 
  \textbf{e} CT examples from approximately the 96th--99th activation percentiles, with suspected phenotype regions enlarged.
  \textbf{f} Kaplan--Meier curves show lower event-free survival with increasing I87 activation, comparing scans with no detected I87 activation against quartiles of positive activation. 
  Shading denotes 95\% confidence intervals.
  \textbf{g} Stratified analyses showed that I87 activation remained associated with myocardial infarction within several subgroups defined by selected clinical findings and combined risk scores. 
  \textbf{h} The analysis was extended to all outcomes significantly associated with I87, with retention varying by subgroup and covariate set.
  }
\label{figure:extended:I87}
\end{figure}

Targeted exploration of I87 combined occlusion analysis, manual inspection and stratified subgroup analyses (Figure \ref{figure:extended:I87}).
We assessed concept localization in 15 high-activation CT examinations using randomly shifted 36- and 72-mm occlusion cubes (36-mm results shown in Figure \ref{figure:extended:I87}).
Masked body voxels were replaced by the whole-scan mean intensity.
We compared I87 activation with and without each cube masked, keeping the randomly masked background unchanged.
Attribution was measured as the signed decrease in I87 activation percentile relative to positive Discovery Site activations, averaged across independent confirmation pairs and mapped to the corresponding voxels.
Organ masks were obtained from precomputed TotalSegmentator segmentations and resampled to the model input grid using nearest-neighbour interpolation.
Patient-level mean attribution within each organ was compared with the remaining imaged body using one-sided paired Wilcoxon signed-rank tests with BH correction.
Activation quartiles were defined among positive Discovery Site activations, with zero activation forming a separate reference group.
Kaplan--Meier curves were estimated from a one-year event-free landmark, with censoring at death, the end of recorded follow-up or four years after CT, and 95\% confidence intervals.

To test whether I87's risk profile could be explained by known markers, we performed stratified analyses of selected potential confounders, individually and in combination.
Individual analyses compared patients without and with each potential confounder, groups termed A and B, defined by positive concept activation or recorded baseline history.
We progressively tested alternative explanations for I87's associations, starting with clinical variables and then adding joint profiles of seven liver and surgical findings.
Finally, we added other concepts with significant associations with the same outcome that ranked highly by unadjusted OR in our data.
At each stage, these features were used to construct outcome-specific risk scores for the stratified analyses.
Scores were estimated using standardized predictors and L2-penalized logistic regression ($C=1$), with five-fold out-of-fold predictions.
For each outcome, we split patients at the median predicted risk before selecting groups based on I87 activation.
Within each subgroup, we compared 1--4-year disease incidence in patients with any positive or top-quartile I87 activation against those with no activation, reporting Wilson 95\% confidence intervals and two-sided Fisher exact tests with BH correction across the 32 displayed comparisons.
We extended this analysis to all outcomes with significant unadjusted associations with I87, using logistic regression to test standardized I87 activation separately in the lower-risk (A) and higher-risk (B) groups, adjusting for the variables used to calculate each risk score.
Associations were retained if positive and significant at BH-adjusted $q<0.05$, assessed separately for each model and subgroup.

\subsection{Data Availability}
Released concept activations and interpretations are hosted at \url{https://huggingface.co/datasets/RobbieHolland/Glyph-research}, subject to access restrictions.
The Merlin abdominal CT dataset \cite{blankemeier2026merlin}, containing de-identified CT scans and paired radiology reports, is available through the Stanford AIMI Center at \url{https://stanfordaimi.azurewebsites.net/datasets/60b9c7ff-877b-48ce-96c3-0194c8205c40}.
Access requires completion and approval of a data-use agreement, after which a secure download link is provided.
This release does not include laboratory measurements and longitudinal clinical records used in the present study.
Mayo Clinic study-level data and clinical covariates are not distributed.

\subsection{Code Availability}
Analysis and reproduction code is available at \url{https://github.com/StanfordMIMI/Glyph-research}. Pretrained SAE checkpoints and normalization statistics are available at \url{https://huggingface.co/RobbieHolland/Glyph-research}. External Site \#2 (Mayo Clinic) study-level data and clinical covariates required for confounder-adjusted refitting are not distributed.

\subsection{Open source}
\label{methods:open_source}
The pipeline described above, spanning embedding extraction, SAE training, automated interpretation, hypothesis search, and visualization, is available through the open-source library Glyph (\url{https://pypi.org/project/glyph-discovery/}).
Users can learn new sparse concepts from their own data or apply the pretrained dictionaries released with this paper using the corresponding foundation models.
Foundation models can be selected through built-in presets or supported Hugging Face model identifiers; custom models can be connected by specifying their preprocessing and embedding output.
Listing~\ref{lst:api} illustrates the Python API for learning new concepts, Listing~\ref{lst:pretrained} shows how to measure pretrained concepts, and Listing~\ref{list:skill} demonstrates running the workflow through a coding agent.

When training new dictionaries, \texttt{fit} evaluates SAE configurations and favors sparse representations that achieve a validation reconstruction $R^2$ of at least 0.85, allowing modest increases in the number of active concepts for further improvements in reconstruction.
A warning is issued if this threshold is not reached.
\texttt{hypothesis\_search} runs the association study described in Section~\ref{section:methods:or}, with optional confounder adjustment.
\texttt{visualize} produces an interactive concept-space explorer with association statistics and Bonferroni-significant concepts highlighted.
Additional agent skills support literature searches, comparisons against published markers, and concise visual reports.
Automated interpretation is optional and requires configured access to a supported language model.

\begin{listing}[t]
\centering
\begin{tcolorbox}[colback=black!3!white, colframe=black!35!white, boxrule=0.4pt, arc=1mm]
\begin{minted}[fontsize=\small, breaklines, style=friendly]{python}
from glyph import Glyph

glyph = Glyph(config_path='config', config_overrides=[
    'data.metadata=/path/to/metadata.csv',
    'data.input_cols=[image_path, report_path]',
    'data.outcome_cols=[pneumonia]',
    'data.embedding_map={image_path: dinov3, report_path: qwen3-embedding-0.6b}',
])

glyph.fit()                # selects a sparse SAE using validation R^2
glyph.hypothesis_search()  # association testing over sparse concepts
glyph.visualize(outcome='pneumonia', split='test')
\end{minted}
\end{tcolorbox}
\caption{\textbf{Learning new concepts with Glyph.}
The example uses DINOv3 for images and Qwen3-Embedding-0.6B for report text.
The \texttt{embedding\_map} connects each metadata input column to its encoder.
\texttt{fit} selects a sparse SAE configuration using validation reconstruction performance, warning if $R^2<0.85$.
\texttt{hypothesis\_search} tests sparse concepts against the specified outcomes, and \texttt{visualize} displays the interactive concept space and association results.}
\label{lst:api}
\end{listing}

\begin{listing}[t]
\centering
\begin{tcolorbox}[colback=black!3!white, colframe=black!35!white, boxrule=0.4pt, arc=1mm]
\begin{minted}[fontsize=\small, breaklines, style=friendly]{python}
from glyph import Glyph

# 1. EXTRACT A SINGLE PUBLISHED CONCEPT ON YOUR DATA
ct_concepts = Glyph.from_pretrained('ct-merlin')
i87 = ct_concepts.measure('I87', metadata='ct_metadata.csv')

# Optionally restrict extraction to one study.
i87 = ct_concepts.measure(
    'I87', metadata='ct_metadata.csv', sample_id='scan_001'
)

# 2. EXTRACT THE FULL PUBLISHED CONCEPT SPACE ON YOUR DATA
# Example: apply the paper's entire laboratory concept space.
labs_concepts = Glyph.from_pretrained('labs-loinc')
concepts = labs_concepts.measure(metadata='labs_metadata.csv')
labs_concepts.visualize()

# Select a concept from the already measured space.
l127 = concepts['L127']
\end{minted}
\end{tcolorbox}
\caption{\textbf{Extracting published concepts on new data.}
\textbf{Single-concept extraction.} Specify a concept code to measure one published concept across a cohort or for a selected study, illustrated with imaging concept I87.
\textbf{Full-space extraction.} Omit the concept code to apply the paper's entire pretrained dictionary for a modality to the supplied cohort, illustrated with the laboratory dictionary.
The resulting study-by-concept table can be visualized, and individual concepts such as L127 can be selected directly.
The metadata supplies study identifiers and input paths; Glyph computes the matching foundation-model embeddings and concept activations on the user's data while preserving the paper's concept indices.
Neither outcome labels nor SAE retraining is required for concept extraction and exploration; labels are needed for association testing and significance highlighting.}
\label{lst:pretrained}
\end{listing}

\begin{listing}[ht]
\begin{tcolorbox}[colback=black!3!white, colframe=black!35!white, boxrule=0.4pt, arc=1mm]
\begin{minted}[fontsize=\small, breaklines, style=friendly]{text}
$ claude

> Install Glyph from github.com/RobbieHolland/Glyph and its bundled skills
  for Claude Code, then run the pipeline on my cohort.
  metadata:  <path/to/metadata.csv>
  inputs:    <input_col_1, input_col_2, ...>
  outcomes:  <outcome_1, outcome_2, ...>
  encoders:  <input_col_1: model_name_or_HF_ID, input_col_2: ...>
  Report the reconstruction R^2, draw the concept map, and list the concepts
  most associated with each outcome, including odds ratios and which survive
  Bonferroni correction. Use the literature skill to identify established
  markers with supporting sources, and produce a brief visual report.

> Now apply the paper's pretrained concepts instead of fitting a new SAE.
  Use the appropriate bundle: <ct-merlin | findings-qwen3 | labs-loinc>.
  Tell me which associations reproduce in my cohort.
\end{minted}
\end{tcolorbox}
\caption{\textbf{Running Glyph from a coding agent.}
Bundled skills guide Claude Code or Codex through configuration, analysis, literature comparisons, visualization, and reporting.
The agent installs the package, configures the inputs and outcomes, runs the requested analyses, and summarizes the results.
The second prompt applies the paper's pretrained dictionaries and corresponding foundation models to a new cohort without retraining the SAE.}
\label{list:skill}
\end{listing}

\subsection{Ethics Statement}
All datasets used in this study were collected and analysed in accordance with institutional guidelines and regulations. The use of retrospective clinical data was approved by the Institutional Review Board at Stanford University, with a waiver of informed consent owing to the retrospective nature of the study. The Stanford cohort comprised abdominal CT examinations acquired between December 2012 and October 2018, with associated radiology reports, laboratory measurements, and electronic health records. Held-out cohorts, including Mayo Clinic, were accessed under the applicable institutional approvals and data-use agreements, with waivers of informed consent, and were used solely for evaluation.

\subsection{Author Contributions}
R.H. designed the study, developed the methodology and code, trained models, performed experiments and analyses, interpreted results, created figures, and wrote the manuscript.
A.K., P.W., T.N.W., and E.P.R. provided technical advice and methodological input.
M.A.E. and Andrea Fisher contributed to interpreting the clinical significance of the findings.
D.L. applied the models and performed external evaluation, coordinated by I.B.
J.P., L.F., M.P., V.S.-L., S.M., and M.S.W. contributed to manuscript writing and presentation.
Ates Fettahoglu contributed to the phenome-wide association analyses.
I.L. contributed to biomarker testing and clinical interpretation.
F.R. provided cardiovascular expertise.
V.S. advised on approach to confounder adjustment.
C.P.L. and A.S.C. provided senior scientific guidance.
S.G. contributed to study design, clinical interpretation, and supervision.

\subsection{Funding}
This work was supported by NIH grants R01 HL167974, R01 HL169345, R01 AR077604, R01 EB002524, R01 AR079431, P41 EB027060, and P50 HD118632, and the Advanced Research Projects Agency for Health (ARPA-H) Biomedical Data Fabric (AY2AX000045).
\newpage

\clearpage

\clearpage
\makeatletter
\immediate\write\@auxout{%
  \string\gdef\string\mainpagetotal{%
    \number\numexpr\value{page}-1\relax}}
\makeatother

\setcounter{page}{1}
\rfoot{\small\sffamily\bfseries\thepage/\pageref*{LastPage}}
\appendix
\section{Supplementary material}
\localtableofcontents
\newpage

\begin{figure}[htbp]
    \centering
    \includegraphics[width=0.95\textwidth]{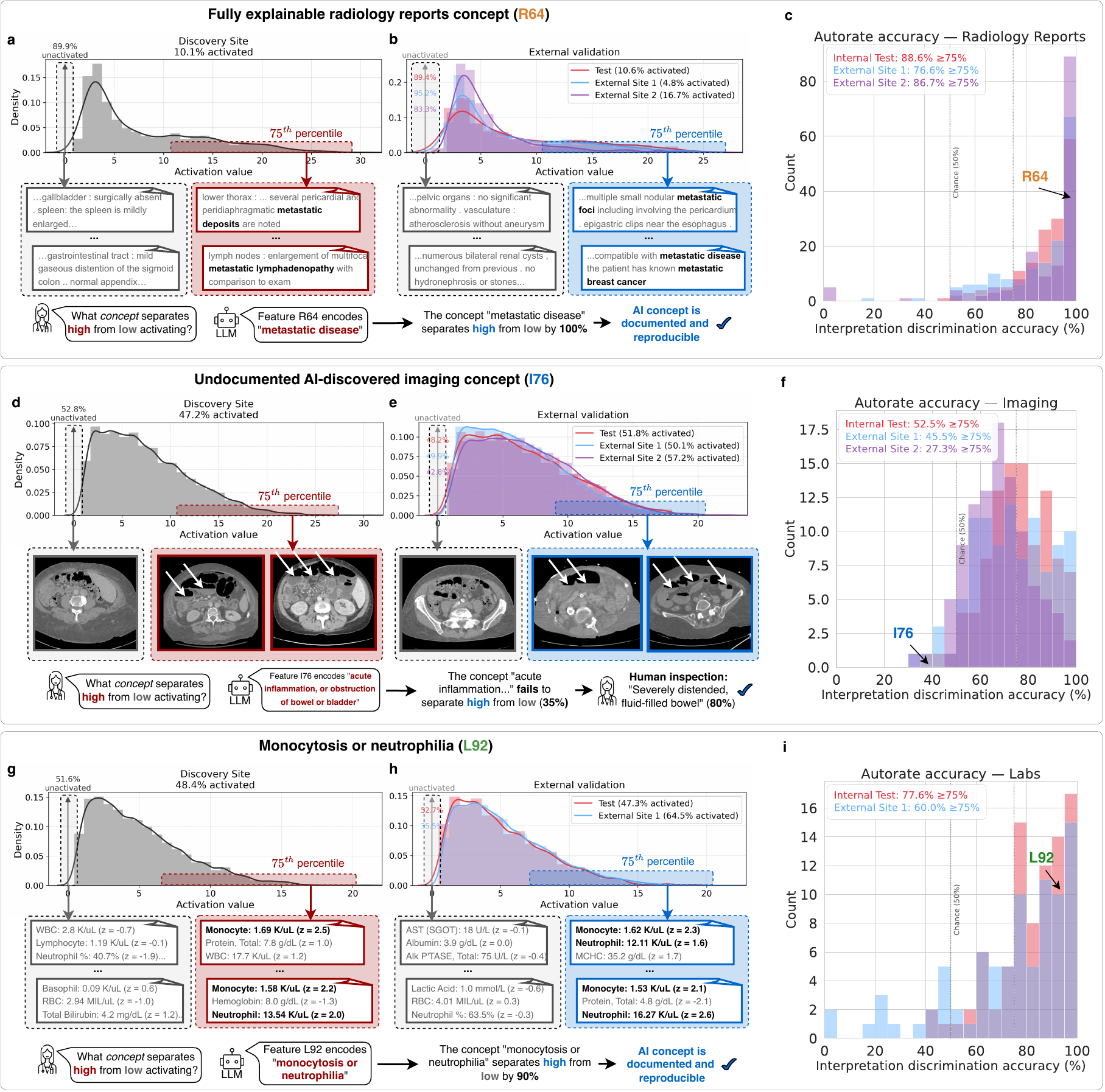}
    \caption{Glyph concepts are automatically extracted, interpreted, and
    reproduced zero-shot at external sites. 
    Each concept is interpreted on the Discovery Site by comparing top- versus non-activating examples, then re-extracted at an Internal Test cohort and two external sites where the same interpretation is re-tested on unseen patients; consistent discrimination shows both concepts and interpretations transfer across institutions.
    \textbf{a--c} A radiology report concept, labeled \textit{metastatic disease}, re-extracted zero-shot and confirmed at every site. AutoInterp accuracy is highest for reports: as human-written summaries, their concepts are near-fully interpretable but bounded in discovery potential.
    \textbf{d--e} An imaging concept whose interpretation is not consistently documented in radiology reports, yet was found to rank among the top 10\% of imaging predictors of infection, sepsis, wasting, and organ failure. 
    \textbf{f} Remarkably, 52.5\% of imaging concepts were interpretable via corresponding radiology reports.
    \textbf{g--i} A laboratory concept, labelled \textit{monocytosis or neutrophilia}, confirmed across sites.
    }
    \label{figure:8}
\end{figure}

\subsection{Interpretation of Glyph concepts and validation across sites}
\label{supplementary:autointerp}
We assessed the interpretability of Glyph's discovered features, developing a contrastive modification of AutoInterp \cite{templeton2024scaling} and interpretation testing outlined in Section \ref{methods:autointerp}.
To make analysis more token-efficient, we only interpret high-risk concepts, defined by those with at least one significant association to future disease (see Section \ref{section:results:association_study}).
Prior work highlighted the difficulty of interpreting radiological SAE features at an F1 threshold of 0.75 \cite{bouzid2025insights}, motivating our choice of a discrimination accuracy threshold of $\geq 75\%$ for considering a concept interpretable.

Using this method and threshold, 88.6\% of radiology report concepts were interpretable internally, and 76.6\% and 86.7\% of interpretations successfully generalized to External Sites \#1 and \#2, respectively (see Figure \ref{figure:8}c).
This high rate was expected, as these concepts are derived from embeddings of text that explicitly describe clinical findings.
Interpreting concepts derived from images represents a greater challenge, as Merlin likely relies on its own internal set of concepts that may not correspond to anything radiologists routinely or consistently document in reports.
Remarkably, 52.5\% of Merlin image concepts were broadly interpretable by corresponding radiology reports (see Figure \ref{figure:8}d--f).
Labs concepts were more automatically interpretable than imaging (see Figure \ref{figure:8}g--i), and captured physiological states such as monocytosis.

Investigating image concepts not interpretable by AutoInterp, we find many still carry identifiable meaning, when viewed directly in images by radiologists.
For example, we noticed I76 was not interpretable via AutoInterp, but was among the top 1\% closest imaging-derived markers to a broad swath of laboratory concepts, spanning inflammation, cholesterol, and albumin levels (annotated in Figure \ref{figure:2}c).
Strikingly, despite not being routinely documented, I76 was significantly associated with the onset of infection, sepsis, septicemia, and wasting.
Prompted by these results, radiologists visually inspected images retrieved by this concept, finding it activates in patients with severely distended, fluid-filled bowel loops (see Figure \ref{figure:8}d--e).

\subsection{Inter-concept dependency graph}
\begin{figure}[!ht]
        \centering
        \includegraphics[width=0.9\textwidth]{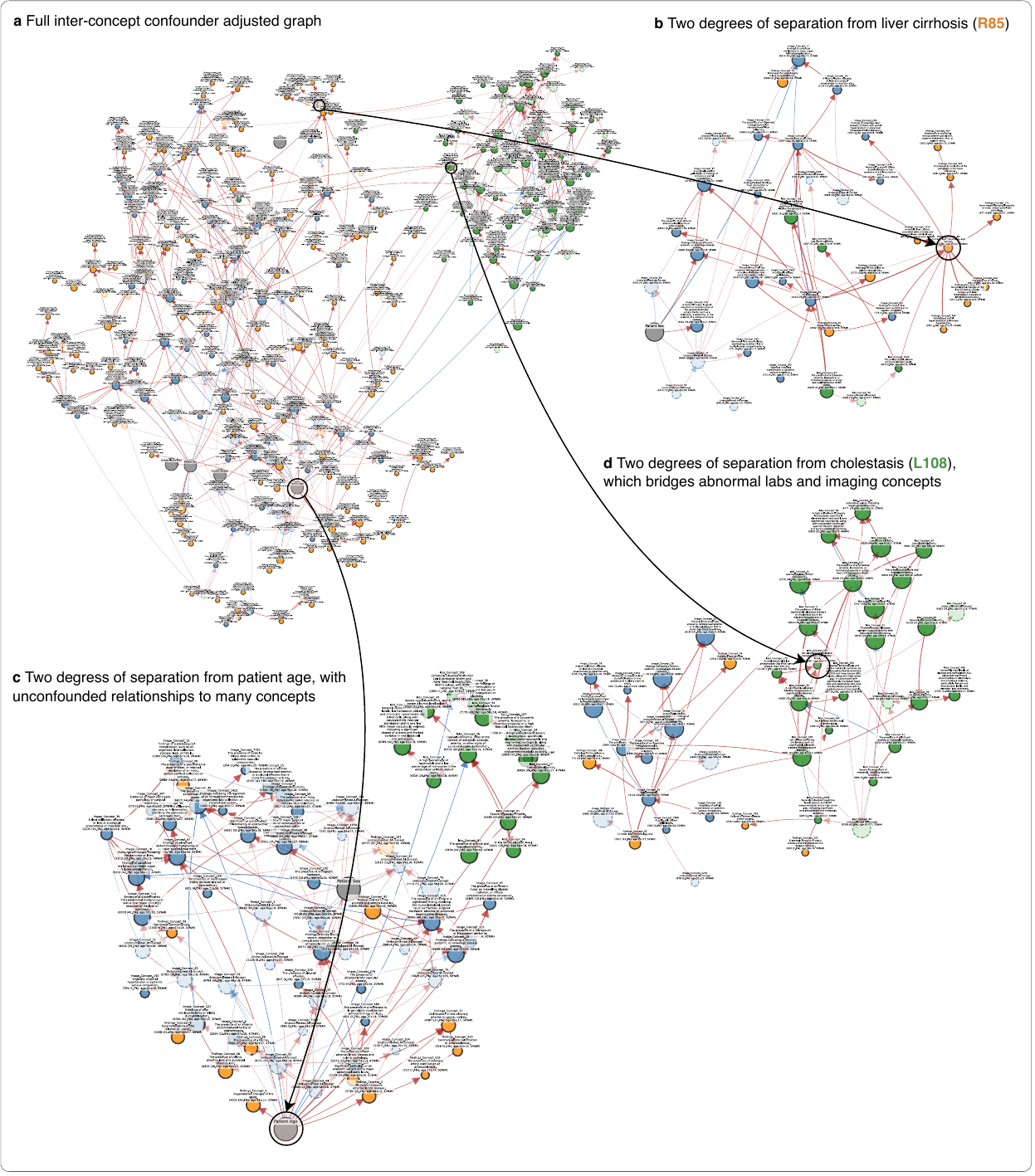}
        \caption{Inter-concept dependency graph. \textbf{a} The full adjusted network. Edges denote directed dependencies from the fitted sparse SEM, retained where the standardized coefficient satisfies $|w| \geq 0.1$; edge width scales with $|w|$ and colour denotes sign (red positive, blue negative). Dashed edges involve at least one undocumented concept. \textbf{b} Concepts within two degrees of separation from liver cirrhosis (R85), a disease hub known to link to splenomegaly (I65) and portal hypertension (R309). 
        \textbf{c} Patient age shows adjusted associations with many concepts, particularly cardiovascular disease and spinal degeneration (R1). 
        \textbf{d} In this abdominal CT cohort, labs indicating cholestasis (L108) form the major bridge between modalities, connecting abnormal lab states, such as elevated alkaline phosphatase (L33) and cytopenia (L127), to structural imaging phenotypes such as splenomegaly (I65) and intra-abdominal fluid (I105).}
        \label{figure:extended:6}
    \end{figure}

As an additional exploratory analysis, we investigated conditional relationships among concepts, complementing the concept--outcome analyses. 
We learned a sparse directed acyclic graph (DAG) over all documented and undocumented concepts, including age, sex, smoking status, heavy alcohol use and BMI as additional nodes to account for shared demographic and lifestyle variation. 
This yielded a joint map of inter-concept dependencies conditional on the fitted graph structure (Figure \ref{figure:extended:6}).
We standardized all node activations and fit a linear structural equation model $X = XW + \varepsilon$ with an $\ell_1$ sparsity penalty on the weighted adjacency $W$, subject to a continuous acyclicity constraint. 
We used DAGMA \cite{bello2022dagma}, which enforces acyclicity through a log-determinant characterization and a continuation schedule that progressively tightens the constraint toward an exact DAG, building on the continuous-optimization formulation of NOTEARS \cite{zheng2018dags}. 
Entries of $W$ below a magnitude threshold were set to zero, and the surviving weights define the directed edges. 
This provides a computationally tractable approach to exploring inter-concept structure across approximately $d^2$ candidate relationships.
The resulting graph represents conditional relationships under the fitted model and should be interpreted as exploratory. 
Including measured clinical covariates does not ensure that confounding has been eliminated, and edge directions do not establish causal relationships. 
Estimated edges may reflect population-specific structure, residual confounding, or sensitivity to model assumptions, thresholding and optimization.

\subsection{Glyph provided complementary associations to pre-coded phenotypes}
\label{supplementary:phecodes_comparison}

\begin{figure}[!ht]
    \centering
    \includegraphics[width=0.85\textwidth]{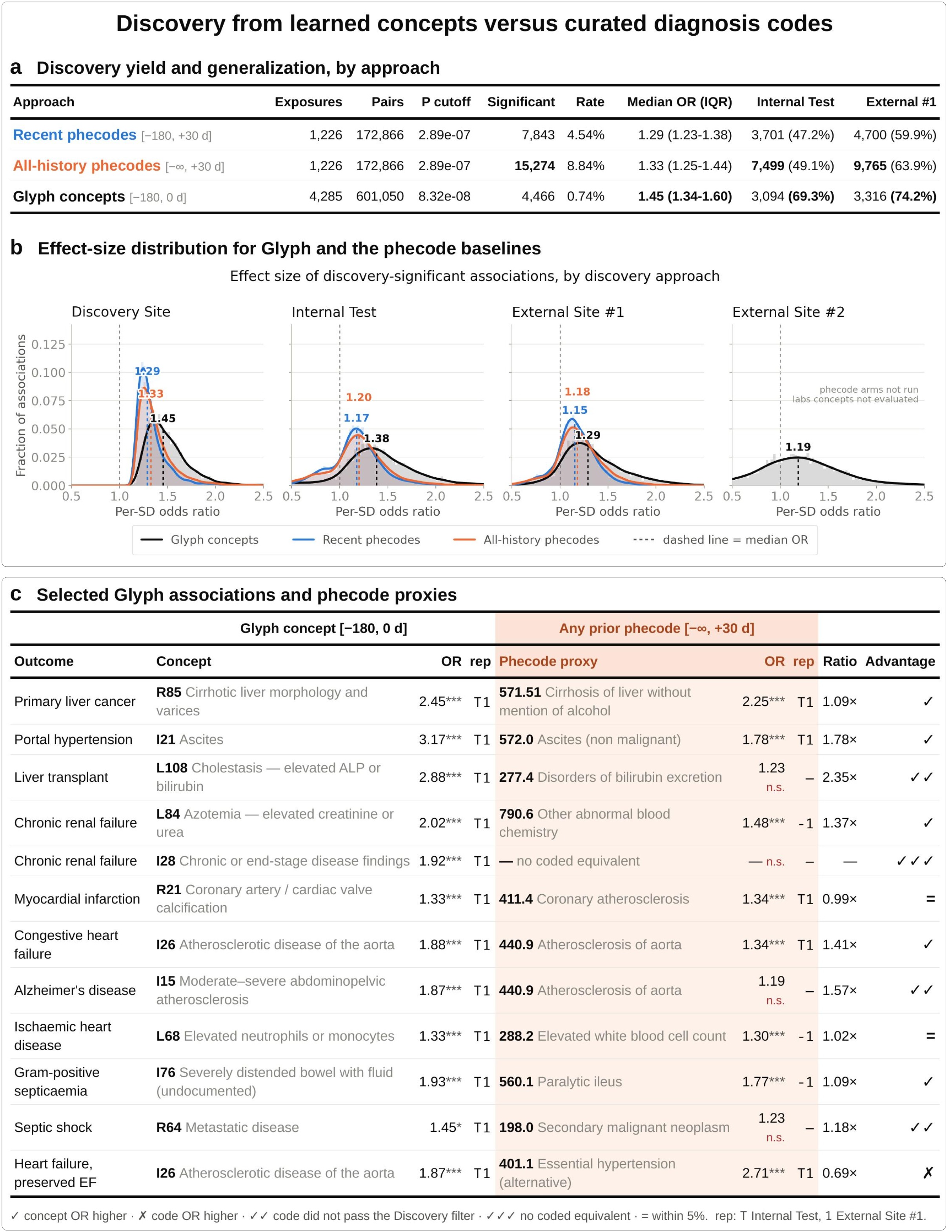}
\caption{\textbf{Glyph concepts build on pre-coded discovery.}
Phecodes group ICD codes into manually curated, clinically meaningful concepts for use in research \cite{denny2013systematic}.
We test complementary benefits of data-driven Glyph concepts over Phecodes.
\textbf{a--b} Binary Phecodes drawn from the full recorded history yield 3.4$\times$ more significant associations than Glyph concepts, but reproduce less often at Internal Test (49\% versus 69\%) and have a smaller median effect (OR 1.33 versus 1.45).
Recent Phecodes, restricted to the scan-adjacent window, also yield more associations than Glyph concepts but reproduce less often.
\textbf{c} Directly corresponding Phecodes were often less specific than imaging- or laboratory-derived Glyph concepts and yielded weaker or nonsignificant associations.
In other cases, the broader clinical history captured by Phecodes provided a stronger proxy for future risk.
For example, hypertension served as a proxy for atherosclerosis (bottom) and outperformed the corresponding Glyph concept for some outcomes.
}
\label{figure:extended:phecode_baseline}
\end{figure}

Having established that Glyph concepts yield significant and reproducible markers of future disease risk, we asked whether pre-coded PheWAS for automated clinical discovery would have identified the same discoveries \cite{denny2013systematic}.
We treated Phecodes as exposures and applied the same discovery pipeline, including multiple-test corrected Discovery and Reproducibility filters, searching among the same 141 outcomes where Glyph has the maximum expected yield.

Aggregate binary-exposure results are presented in Figure \ref{figure:extended:phecode_baseline}a--b.
Recent Phecodes recorded from $-180$ to $+30$ days relative to the scan yielded 7,843 promising associations, with significantly lower replication than Glyph concepts at Internal Test (47.2\% vs. 69.3\%, $p=9.77\times10^{-4}$) and External Site \#1 (59.9\% vs. 74.3\%, $p=1.95\times10^{-3}$)
Using each patient's entire recorded coding history increased the Phecode yield to 15,274 promising associations, approximately 3.4$\times$ the number identified by Glyph, of which 49.1\% and 63.9\% reproduced in the two held-out cohorts.
The median Discovery-site effect was smaller for both Phecode arms than for Glyph concepts (OR 1.29 for Recent Phecodes and 1.33 for all-history Phecodes, versus 1.45 for Glyph).
The greater yield reflects the broader scope of ICD codes, which span the full medical record, whereas Glyph concepts were limited to phenomena observable in abdominal CT images, and laboratory panels.
Comparable yield may therefore be possible as Glyph expands to additional modalities and examinations.

We next assessed which Glyph associations could have been identified from codes alone (see Figure \ref{figure:extended:phecode_baseline}c).
For previously reported associations, we identified if there was a matching Phecode and compared their associations directly.
Directly matched associations were often recoverable from Phecodes.
For example, Phecode 571.51 (cirrhosis of the liver) reproducibly predicted primary liver cancer (OR 2.03--2.25), independently corroborating the association identified for R85 (cirrhotic liver morphology and varices; OR 1.64--2.55).
Across the majority of comparisons with a matching Phecode, the Glyph concept had a larger odds ratio.
Some matched Phecodes did not pass the Discovery filter, and some additional Glyph associations had no corresponding Phecode.
Overall, Phecodes identified more associations across the full medical record, while Glyph concepts yielded larger median effects, reproduced more frequently, and produced stronger associations within directly matched phenotypes.
Phecodes therefore capture information beyond the modalities available to Glyph, while Glyph concepts extract more predictive signal from the modalities it does observe.

\subsection{Architectural ablations}
\label{section:results:architectural_ablation}
\begin{figure}[htbp]
    \centering
    \includegraphics[width=0.85\textwidth]{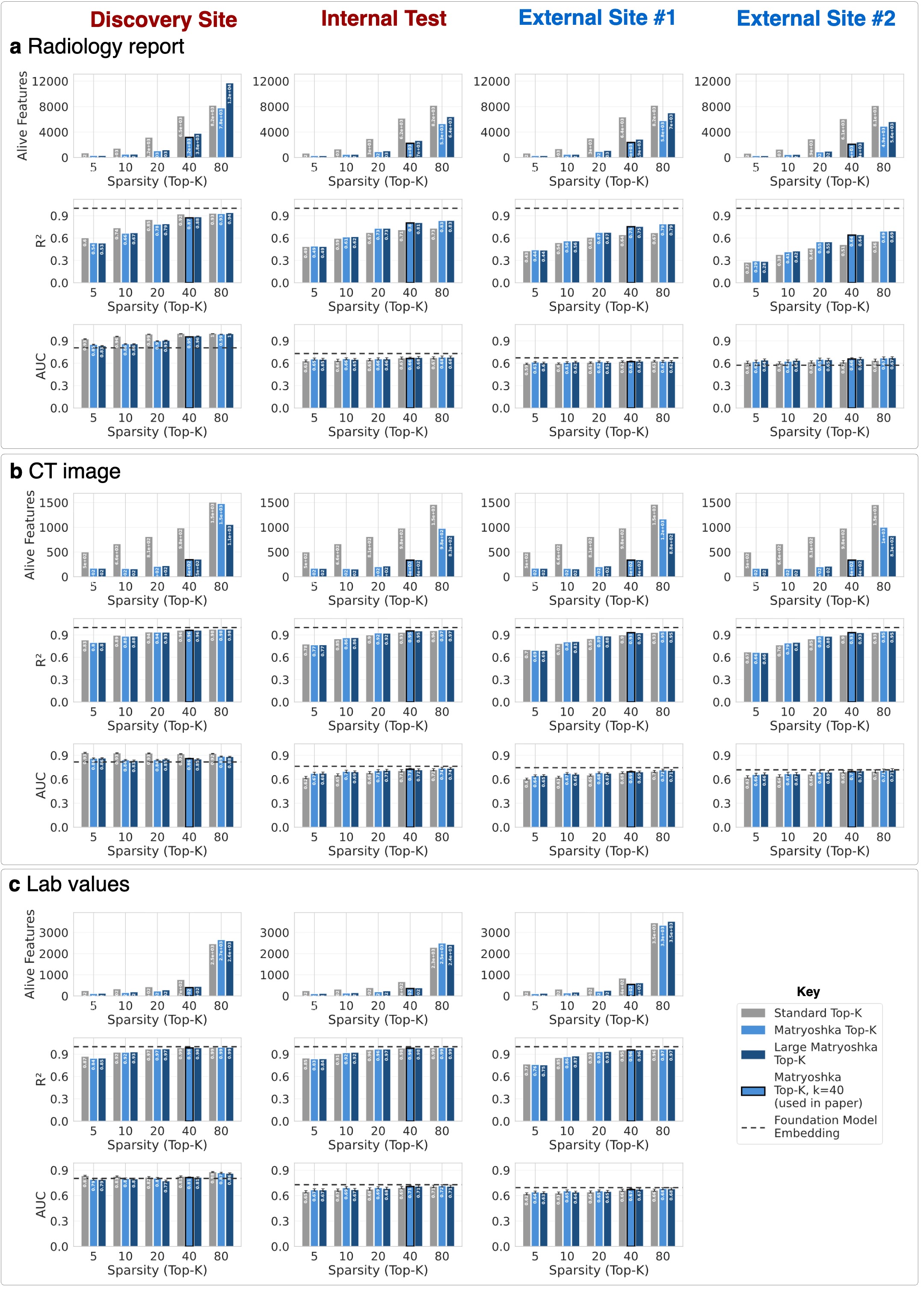}
    \caption{Architectural ablation of the sparse autoencoder across 141 disease prediction tasks.
    Results are organized by modality, with panels \textbf{a--c} corresponding to radiology reports (Qwen3-Embedding-8B), CT imaging (Merlin), and laboratory values (Labs FM), respectively.
    Within each modality, results are organized by metric, showing the number of alive features, $R^2$ between original and reconstructed foundation-model embeddings, and mean downstream AUC across the 141 disease prediction tasks.
    The number of alive features and $R^2$ are evaluated on the discovery training set and unseen external data, while downstream AUC assesses whether the Glyph concept space preserves the predictive performance of the underlying foundation models with minimal degradation.
    Across modalities, diminishing returns in reconstruction are observed beyond $K=40$, with $R^2$ improving and subsequently plateauing at this sparsity level.}
    \label{figure:6}
\end{figure}

The quality of Glyph's concept space, and hence the sensitivity of all downstream discovery, depended directly on the architecture of the sparse autoencoder.
Two parameters governed standard Top-$K$ SAEs: sparsity $K$, which set the number of features active per sample, and dictionary size $N$, the maximum number of distinct features available.
Standard SAEs, however, have a known tendency known as \textit{feature splitting} to recover only fine-grained concepts while neglecting broader phenomena higher in the conceptual hierarchy.
The Matryoshka SAE addressed this by organising features into nested groups of increasing specificity, simultaneously encoding concepts at multiple scales.
To identify the architecture best suited to Glyph's multimodal clinical data, we systematically evaluated a standard Top-$K$ SAE alongside two Matryoshka variants with nested dictionary sizes [128, 512, 2048, 8192] and [256, 1024, 4096, 16384], across five sparsity levels ($K \in \{5, 10, 20, 40, 80\}$) and all three modalities.

Across all modalities, increasing $K$ was the dominant factor for improving reconstruction quality, reducing MSE and raising $R^2$ between original and reconstructed FM embeddings (Figure~\ref{figure:6}).
All modalities exhibited diminishing returns beyond $K = 40$, beyond which additional active features yielded marginal improvements while increasing computational cost.
The number of alive features in the dictionary also grew with $K$, reflecting the richer encoding capacity afforded by denser activations.
Modality differences in the required number of features broadly tracked embedding dimensionality: radiology reports, embedded with the 4096-dimensional Qwen3-Embedding-8B, required more active features than CT images via Merlin (2048-dimensional), which in turn required more than laboratory values via the Labs FM (768-dimensional).

Matryoshka SAEs improved generalisation to external data relative to the standard Top-$K$ baseline, an advantage more clearly visible on the External Site \#1 cohort than on the internal discovery set.
The SAE trained on Discovery data maintained strong performance across all test sets, though with a modest drop on external cohorts that mirrored the reduced replication rates in Sections 2.2 and 2.3, and may partly reflect this distributional shift.
No meaningful benefit was observed between the two Matryoshka dictionary sizes, and we accordingly selected the Top-$K=40$ Matryoshka SAE with nested sizes [128, 512, 2048, 8192] for all analyses, balancing concept quality, generalisation, and computational cost, and achieved over 90\% reconstruction of the original embeddings ($R^2 > 0.90$) across all modalities and FMs.

Beyond reconstruction fidelity, a key requirement was that the sparsified concept space preserved the predictive power of the underlying foundation models, confirming that the interpretability step did not discard clinically relevant signal.
Across modalities, Glyph concepts closely matched FM predictive performance, with particularly tight agreement for CT images (AUC 0.73 vs. 0.73 on the internal test cohort, 0.71 vs. 0.70 on External Site \#1) and laboratory values (0.73 vs. 0.71; 0.66 vs. 0.65).
Radiology report concepts showed a modestly larger gap (0.72 vs. 0.63; 0.68 vs. 0.62), consistent with the greater information density of the high-dimensional Qwen3-Embedding-8B text embedding space.
Overall, predictive performance was largely preserved for imaging and laboratory concepts, with larger reductions for radiology report concepts.

\subsection{AutoInterp prompts}
\label{supplementary:autointerp_prompts}

The following prompt is populated with radiology reports findings and lab histories, before being passed to the LLM.

\begin{Verbatim}[breaklines, baselinestretch=0.8, fontsize=\small, breaksymbolleft=, breaksymbolright=]
    === Task ===:
    You are an expert radiologist specializing in abdominal CT scans and lab values. A cohort of patients were selected according to some shared characteristic specific to this cohort. You are provided radiology reports, lab histories, or both, and your task is to analyze them in order determine what are the shared feature(s). Radiology reports and lab histories may be filtered to the words, phrases or recordings most relevant to the shared characteristic to faciliate its identification. The importance level of those words, phrases or recordings should be considered in the analysis.
  
    While the patient data will contain many different pathologies, there should be some undercurrent or one or two features which run through >80% of them. These characteristics should be positive, and somewhat unusual to be among so many reporst from different patients at once. Things like 'no evidence of bowel obstruction' is so common it is not useful. The explanation should be general enough to apply to the highest number of reports in this cohort, but not so general that it is no longer specific to this cohort, and would apply to many of the reference patients from the larger dataset.
    
    To arrive at the most accurate and precise explanation of what this cohort is characterized by, you must engage in explicit chain of thought reasoning. Begin by thoroughly examining all provided data, noting any patterns or commonalities. Pay close attention to recurring terminology, described anatomical structures, and especially consistent pathological findings. Consider how these elements might interrelate to form a singular, distinctive feature that the cohort could be characterized by.
  
    As you progress through your analysis, verbalize your thought process. Explain each step of your reasoning, from initial observations to intermediate conclusions, and finally to your overall assessment. This chain of thought approach will help ensure a comprehensive and well - reasoned final explanation.
  
    After this detailed analytical process, formulate a single, specific explanation of what the cohort is characterized by. Focus on specific features or combinations of features, using 'and' to connect multiple elements if necessary and 'or' if there are multiple possibilities. Refrain from explaining the pathology itself (e.g., avoid statements like "This feature represents X, which is defined as ...").
  
    Below are the patient data, listed in order of how strongly they belong to the cohort (keeping in mind that all of them belong strongly compared to the average patient in the larger dataset):
  
    For reference, here are examples of patients randomly sampled from the much larger dataset that do not belong to this cohort:
  
    {reference_patients}
  
    Finally, the data from the cohort characterized by the feature in question:
  
    {patient_data}
    
    === Format ===:
    It is crucial that you present your final explanation in the following format, started with an asterisk:
    * This cohort is characterized by [your specific, detailed description of what the cohort is characterized by compared to the average patient in the larger dataset].
    The asterisk is absolutely essential. Your explanation must begin immediately after the asterisk, without any additional text, numbering, or preamble. The presence of this asterisk is critical for the proper processing of your response.
\end{Verbatim}

The following prompt tests the accuracy of a given interpretation for a sparse feature. It is populated with testing data, before being passed to the LLM.
\begin{Verbatim}[breaklines, baselinestretch=0.8, fontsize=\small, breaksymbolleft=, breaksymbolright=]
    Task: Distinguish samples (randomly ordered) based on those that do and do not belong to the group characterized by the group characteristics.

    Medical Findings Reports:
    {patient_data}
  
    Group characteristics:
    {interpretation}
  
    Analyze the findings reports one by one, reasoning on whether or not the characteristics apply to the report. Use the following format for each report:
    {{i}}: 1 (if belongs to the group) or 0 (if does not belong to the group). We have designed the list so that exactly half of the reports belong to the group, and half do not. Therefore, you should use the characteristics to differentiate each of the reports into two distinct groups, rather than finding perfect matches.
  
    You need to be as accurate as possible when assigning 1s and 0s. We will score the accuracy of your response at the end.
  
    === Format ===:
    It is crucial that you present your final explanation in the following format, started with an asterisk:
    * PREDICTIONS:
    1. 1 (if the sample belongs to the group) or 0 (if it does not belong to the group)
    2. 1 (if the sample belongs to the group) or 0 (if it does not belong to the group)
    ...
    {{n}}. 1 (if the sample belongs to the group) or 0 (if it does not belong to the group)  
\end{Verbatim}

\subsection{Outcome frequencies across cohorts}
\label{section:supplementary:outcome_prevalence}
Table~\ref{supplementary:outcome_prevalence} lists the 141 outcomes and their positive, negative and excluded counts, followed by the observed case percentage in each cohort. Percentages use the outcome-specific eligible population, comprising studies labelled as progressors or controls under the study's outcome definition. They are not baseline disease prevalence estimates.
\begingroup
\small
\setlength{\tabcolsep}{4pt}
\renewcommand{\arraystretch}{1.12}
\begin{longtable}{@{}>{\raggedright\arraybackslash}p{\dimexpr0.28\textwidth-8\tabcolsep\relax}*{4}{>{\centering\arraybackslash}p{0.18\textwidth}}@{}}
\caption{Outcome case frequencies by cohort, grouped by outcome category.}\label{supplementary:outcome_prevalence}\\
\toprule
Outcome & Discovery & Internal Test & External Site \#1 & External Site \#2 \\
\midrule
\endfirsthead
\multicolumn{5}{l}{\tablename~\thetable{} (continued)}\\
\toprule
Outcome & Discovery & Internal Test & External Site \#1 & External Site \#2 \\
\midrule
\endhead
\midrule
\multicolumn{5}{r}{Continued on next page}\\
\endfoot
\bottomrule
\endlastfoot

\midrule\multicolumn{5}{c}{\textbf{Hepatic / biliary}}\\*\midrule
Liver Abscess And Sequelae Of Chronic Liver Disease (571.8) & 143 /\allowbreak 7,109 /\allowbreak 8,079\newline (1.97\%) & 42 /\allowbreak 2,392 /\allowbreak 2,703\newline (1.73\%) & 126 /\allowbreak 2,558 /\allowbreak 23,119\newline (4.69\%) & 105 /\allowbreak 6,308 /\allowbreak 7,786\newline (1.64\%) \\
Acute And Subacute Necrosis Of Liver (573.4) & 127 /\allowbreak 7,387 /\allowbreak 7,817\newline (1.69\%) & 34 /\allowbreak 2,472 /\allowbreak 2,631\newline (1.36\%) & 93 /\allowbreak 2,598 /\allowbreak 23,112\newline (3.46\%) & 22 /\allowbreak 6,405 /\allowbreak 7,772\newline (0.34\%) \\
Disorders Of Bilirubin Excretion (277.4) & 122 /\allowbreak 7,306 /\allowbreak 7,903\newline (1.64\%) & 27 /\allowbreak 2,426 /\allowbreak 2,684\newline (1.10\%) & 85 /\allowbreak 2,596 /\allowbreak 23,122\newline (3.17\%) & 105 /\allowbreak 6,322 /\allowbreak 7,772\newline (1.63\%) \\
Splenomegaly (579.2) & 111 /\allowbreak 7,008 /\allowbreak 8,212\newline (1.56\%) & 50 /\allowbreak 2,331 /\allowbreak 2,756\newline (2.10\%) & 120 /\allowbreak 2,549 /\allowbreak 23,134\newline (4.50\%) & 30 /\allowbreak 6,307 /\allowbreak 7,862\newline (0.47\%) \\
Cirrhosis Of Liver Without Mention Of Alcohol (571.51) & 110 /\allowbreak 7,083 /\allowbreak 8,138\newline (1.53\%) & 39 /\allowbreak 2,346 /\allowbreak 2,752\newline (1.64\%) & 147 /\allowbreak 2,487 /\allowbreak 23,169\newline (5.58\%) & 137 /\allowbreak 6,224 /\allowbreak 7,838\newline (2.15\%) \\
Diseases Of Spleen (289.5) & 102 /\allowbreak 7,227 /\allowbreak 8,002\newline (1.39\%) & 25 /\allowbreak 2,428 /\allowbreak 2,684\newline (1.02\%) & 84 /\allowbreak 2,567 /\allowbreak 23,152\newline (3.17\%) & 38 /\allowbreak 6,367 /\allowbreak 7,794\newline (0.59\%) \\
\end{longtable}
\noindent\footnotesize Cells give positive / negative / excluded counts, followed by the positive percentage in parentheses. Percentages are calculated as $100\times n/N$, where $n$ is the number of studies labelled as future cases and $N$ is the number with a valid case or control label for that outcome. Progressors had a first recorded diagnosis 1--4 years after CT; controls had no recorded diagnosis and at least four years of follow-up. All other studies were excluded from the outcome-specific denominator. Related and grouped outcomes can overlap and should not be counted as independent diseases; individual Phecode endpoints are not assumed to include all clinically related subtypes. The companion CSV supplies all numerators, denominators and excluded counts. Group assignments follow the study outcome mapping.
\par\endgroup

\subsection{Laboratory input types}
\label{section:supplementary:lab_types}
Table~\ref{supplementary:lab_types} lists the 350 retained numerical input types and the descriptive statistics available in the preprocessing inventory. 
350 types were retained from 14,890 source types, covering 77,502,851 recorded measurements (76.1\% of all laboratory results in the Discovery Site, Internal Test and External Site \#1 combined). Each retained type had at least 90\% numeric measurements. Statistics summarize recorded measurements rather than unique patients and are not specific to a study cohort.

\begingroup
\small
\setlength{\tabcolsep}{3pt}
\renewcommand{\arraystretch}{1.12}
\begin{longtable}{@{}>{\raggedright\arraybackslash}p{\dimexpr0.40\textwidth-10\tabcolsep\relax}>{\raggedright\arraybackslash}p{0.08\textwidth}>{\raggedleft\arraybackslash}p{0.12\textwidth}>{\raggedleft\arraybackslash}p{0.10\textwidth}>{\raggedleft\arraybackslash}p{0.20\textwidth}>{\raggedleft\arraybackslash}p{0.10\textwidth}@{}}
\caption{Retained laboratory input types and available measurement-level summary statistics.}\label{supplementary:lab_types}\\
\toprule
Name (source units) & ID & Measurements & Numeric (\%) & Mean (SD) & IQR \\
\midrule
\endfirsthead
\multicolumn{6}{l}{\tablename~\thetable{} (continued)}\\
\toprule
Name (source units) & ID & Measurements & Numeric (\%) & Mean (SD) & IQR \\
\midrule
\endhead
\midrule
\multicolumn{6}{r}{Continued on next page}\\
\endfoot
\bottomrule
\endlastfoot

\midrule\multicolumn{6}{l}{\textbf{A}}\\*\midrule

Anion Gap (mmol/\allowbreak L) & 2466 & 1,870,639 & 99.4 & 9.41 (3.61) & 4.00 \\

Albumin, Ser/\allowbreak Plas (g/\allowbreak dL) & 1751 & 1,374,127 & 99.9 & 3.49 (10.2) & 1.20 \\

ALT (SGPT), Ser/\allowbreak Plas (U/\allowbreak L) & 1119 & 1,322,529 & 98.5 & 54.8 (158) & 27.0 \\

AST (SGOT), Ser/\allowbreak Plas (U/\allowbreak L) & 1398 & 1,320,653 & 99.2 & 53.2 (239) & 21.0 \\

Alk P'TASE, Total, Ser/\allowbreak Plas (U/\allowbreak L) & 1833 & 1,291,861 & 100.0 & 136 (151) & 66.0 \\

Albumin/\allowbreak Globulin Ratio ((calc)) & 1780 & 55,123 & 100.0 & 1.59 (0.381) & 0.400 \\

AST (SGOT), Ser/\allowbreak Plas (IU/\allowbreak L) & 1397 & 47,624 & 100.0 & 33.7 (72.1) & 13.0 \\
\end{longtable}
\noindent\footnotesize Measurements is the number of source records for the type before nonnumeric values were removed; Numeric is the percentage parseable as numbers. Mean, standard deviation (SD) and interquartile range (IQR; 75th minus 25th percentile) describe the numeric values retained during preprocessing, before model standardisation. Mean, SD and IQR are displayed to three significant figures. Values and units are reproduced from the existing inventory without new harmonisation; measurement frequency does not represent patient prevalence. Types with numeric encodings or measurement-related meanings should not be interpreted as continuous biological analytes. The inventory does not provide patient-level coverage, cohort-specific distributions, medians or reference intervals. Minimum and maximum values and source-measurement proportions are provided in the companion CSV. Laboratory inputs were not used at External Site \#2.
\par\endgroup

\end{document}